\documentclass[review,11pt]{ReportTemplate}
\usepackage{longtable}% for long tables
\usepackage{booktabs}

\usepackage{url}
\usepackage{enumitem}

\usepackage{amsmath}
\usepackage{algorithm}
\usepackage[noend]{algorithmic}
\usepackage{multirow}
\usepackage{array}
\usepackage{graphicx}
\usepackage{subfigure}
\usepackage{multirow}
\usepackage{bm}
\usepackage[american]{babel}
\usepackage{amsmath}
\usepackage{amsfonts,amssymb}
\newtheorem{prop}{Proposition}
\newtheorem{lemma}{Lemma}

\begin{document}

	\def \Z {\mathcal{Z}}
	\def \z {\bm{z}}
	\def \x {\bm{x}}
	\def \f {\bm{f}}
	\def \w {\bm{w}}
	\def \u {\bm{u}}
	\def \bv {\bm{c}}
	\def \y {\bm{y}}
	\def \g {\bm{g}}
	\def \A  {\bm{A}}
	\def \v  {\bm{v}}
	\def \X  {\mathcal{X}}
	\def \Y  {\mathcal{Y}}
	\def \W  {\mathcal{W}}
	\def \D  {\mathcal{D}}
	\def \H  {\mathcal{H}}
	\def \bb {$\begin{small}\bullet\end{small}$}
	\def \tr{\text{tr}}
	\def \j {\bm{j}}
	\def\etal{et al.\;}

\begin{frontmatter}
\title{Multi-Label Learning with 
	Global and Local Label Correlation}

\author{Yue Zhu$^{1}$}
\author{James T. Kwok$^{2}$}
\author{Zhi-Hua Zhou$^{1}$\corref{cor1}}
\address{$^{1}$ National Key Laboratory for Novel Software Technology\\
Nanjing University, Nanjing 210093, China\\
Email: \{zhuy, zhouzh\}@lamda.nju.edu.cn\\
$^{2}$the Department of Computer Science and
Engineering\\ Hong Kong University of Science and Technology, Hong Kong\\
Email: jamesk@cse.ust.hk} \cortext[cor1]{\small Corresponding author.}

\begin{abstract}
	It is well-known that exploiting label correlations is important to multi-label learning. Existing approaches 
	%take advantage of global label correlations by 
	either assume that the label correlations are global and shared by all instances;
	%whereas some other approaches assume 
	or that the label correlations are local and shared only by a data subset.
	In fact, in the real-world applications, both cases may occur that some label correlations
	are globally applicable and some are shared only in a local group of instances. Moreover, it
	is also a usual case that only partial labels are observed, which makes the exploitation of
	the label correlations much more difficult. That is, it is hard to estimate the label
	correlations when many labels are absent. In this paper, we propose a new multi-label
	approach \texttt{GLOCAL} dealing with both the full-label and the missing-label cases,
	exploiting global and local label correlations simultaneously, through learning a latent
	label representation and optimizing label manifolds. The extensive experimental studies
	validate the effectiveness of our approach on both full-label and missing-label data.
\end{abstract}

\begin{keyword}
Global and local label correlation, label manifold, missing labels, multi-label learning.
\end{keyword}
\end{frontmatter}

\section{Introduction}
\label{sec:introduction}
In real-world classification applications, an instance is often associated with more than
one class labels. For example,  a scene image can be annotated with several tags
\cite{boutell2004learning}, a document may belong to multiple topics
\cite{ueda2002parametric},  and a piece of music may be associated with different genres
\cite{turnbull2008semantic}. Thus, multi-label learning 
%aims at handling this kind of data, and 
has attracted a lot of attention in recent years \cite{ zhang2014review}.  

Current studies on multi-label learning try to incorporate label correlations of different orders \cite{zhang2014review}. 
However, existing approaches mostly focus on global
label correlations 
shared by all instances
\cite{furnkranz2008multilabel,ji2008extracting,read2011classifier}.
For example, labels ``fish'' and ``ocean'' are highly correlated, and so are ``stock'' and ``finance''.
On the other hand, certain label correlations are only shared by  a local data subset 
\cite{huang2012}.
For example, ``apple'' is related to ``fruit'' in gourmet magazines, but is related to
``digital devices'' in technology magazines. 
Previous studies focus on exploiting either  global or  local label
correlations. However, considering both of them is obviously more beneficial and desirable.

Another problem with label correlations is that they are usually difficult to specify
manually. As label correlations may vary in different contexts and there is no unified
measure for specifying appropriate correlations, they are usually
estimated from the observed data. 
Some approaches learn the label hierarchies  by hierarchical clustering~\cite{Punera2005Automatically} or Bayesian network structure learning~\cite{zhang2010multi}.  
%thus the label correlations are estimated by the hierarchy. 
However, the hierarchical structure may not exist in some applications.  For example, labels such as
``desert'', ``mountains'', ``sea'', ``sunset'' and ``trees'' do not have any natural hierarchical
correlations,
and label hierarchies may not be useful.  
Others  estimate label correlations by the co-occurrence of labels in training data~\cite{NIPS2011_4239}.  
However, it may cause
overfitting. Moreover, 
co-occurrence is less meaningful
for labels with very few positive instances.

In multi-label learning, some labels may be missing from the training set. For example, 
human labelers may ignore object classes they do not know or of little interest. 
Recently,
multi-label learning with missing labels has become a hot topic.
Xu \etal \cite{xu2013speedup} and Yu \etal \cite{Yu2014}
considered using the low-rank structure  on the instance-label mapping.
A more direct approach to model the
label dependency 
approximates the label matrix as a product of two low-rank matrices
\cite{goldberg2010transduction}.  This leads to simpler recovery of the
missing labels, and produces a latent representation of the label matrix. 

In the missing label cases,
estimation of label correlation becomes even more difficult, 
as the observed label distribution is  different
from the  true one. 
%Hierarchical clustering for example,  where correlated labels are in the same cluster, supposing the true distributions of two class labels are very different and they will be assigned into different clusters if full labels are observed.  %However, the actual observation may be similar due to the missing labels, thus those less-correlated labels may be clustered into the same cluster.  However, if some of the labels are missing, the observed distribution of those two labels may be similar,  namely, two less-correlated labels may be clustered into the same cluster. As a result, the estimation of label correlation will be biased. \footnote{*** rewrite this sentence \textbf{Is it better?}} 
%For example,  missing a label that appears often with other labels will reduce the number of co-occurrence, but the missing rate is assumed to be the same\footnote{*** why? \textbf{Most work on missing labels or partial labels assume that labels are uniformly randomly missing. Can we change this sentence ``but ..., which will be misleading'' into ``if  uniformly random missing labels are considered, it will lead a biased estimation of label correlations.''  }} for all labels, which will be misleading. 
As a result,
the aforementioned methods (based on hierarchical clustering and co-occurrence, for example)
will produce biased estimates of label correlations.

In this paper, we propose a new approach called ``Multi-Label Learning with \textbf{GLO}bal
and lo\textbf{CAL} Correlation'' (\texttt{GLOCAL}),
which simultaneously recovers the missing labels, trains the linear classifiers  and exploits both global and local label
correlations. It learns a latent label representation.
%which is compressed and contains more information for each latent label. 
%The label manifolds are optimized such that 
Classifier outputs are encouraged to be similar on highly positively correlated
labels, and dissimilar on highly negatively correlated labels.  We do not assume 
the presence of external knowledge sources specifying the label correlations. Instead,
these
correlations are learned simultaneously with the latent label representations and instance-label mapping. 

The rest of the paper is organized as follows. In Section \ref{sec:relatedwork}, related
works of multi-label learning with label correlations are introduced. In Section
\ref{sec:approach}, the problem formulation and the \texttt{GLOCAL} approach are proposed. 
Experimental results 
are presented in Section \ref{sec:exp}.
Finally, Section \ref{sec:conclusion} 
concludes
the work.

%For  description simplicity, 
%Let $n$ be the number of training instances, $d$ be the input dimensionality, and $l$ the number of labels.
%the function $\Pi_\Omega(\cdot)$, where 

\noindent
{\bf Notations}
For a matrix $\A$, 
$\A^\top$
denotes its transpose,
$\mathrm{tr(\A)}$
is its trace,
$\|\A\|_F$ is its Frobenius norm, and
$\text{diag}(\A)$ returns a vector containing the diagonal elements of
$\A$.
For two matrices $\bm{A}$ and $\bm{B}$,
$\bm{A}\circ \bm{B}$ denotes the Hadamard 
(element-wise)
product.
For a vector $\bv$, 
$\|\bv\|_2$ is its $\ell_2$-norm, and
$\text{Diag}(\bv)$ returns a diagonal matrix with 
$\bv$ on the diagonal.

\section{Related Work} \label{sec:relatedwork}

Multi-label learning has been widely studied in recent years. Based on the degree of
label correlations used, it can be divided into three categories \cite{zhang2014review}:
(i) first-order; (ii) second-order; and (iii) high-order. 
For the first-order strategy, label correlations are not considered, and the multi-label
problem is transformed into multiple independent binary classification problems. 
For example, 
\texttt{BR} \cite{boutell2004learning} 
trains a classifier for each label
independently. 
For the second-order strategy, pairwise label relations are considered. For example, \texttt{CLR} \cite{furnkranz2008multilabel}  transforms the multi-label learning problem into the pairwise label ranking problem. 
For the high-order strategy, all other labels' influences imposed on each label are taken
into account. For example, \texttt{CC} \cite{read2011classifier} transforms the multi-label
learning problem into a chain of binary classification problems, with the ground-truth
labels encoded into the features.

Most previous studies focus on global label correlations. However, \texttt{MLLOC}
\cite{huang2012} demonstrates that sometimes label correlations may  only be shared by  a local data subset. 
%that is, local label correlations should be taken into account. 
Specifically, it 
enhances the feature representation of each instance 
by embedding a code into the feature space, which encodes the influence of labels of an instance to the local label correlations. This has some limitations. First, when the dimensionality of the feature space is
large, the code is less discriminative and will be dominated by the original features.
Second, 
\texttt{MLLOC} considers only the local 
label correlations, but not the
global ones.
Third, \texttt{MLLOC} cannot learn with  missing labels.

In some real-world applications, labels are partially observed, and
multi-label learning with missing labels has attracted much attention. 
\texttt{MAXIDE} \cite{xu2013speedup} is based on fast low-rank matrix completion, and has 
strong theoretical guarantees. However, it only works in the transductive setting.
Moreover,
a label correlation matrix 
has to be specified
manually.
\texttt{LEML} \cite{Yu2014} also relies on a low-rank structure, and works in an inductive
setting. However, it only implicitly uses global label correlations.
\texttt{ML-LRC}  \cite{xu2014learning} adopts a low-rank structure to capture 
global label correlations, and addresses the missing labels by introducing a supplementary label
matrix.
%which augments the original label matrix via exploiting the global label correlations. 
However, only global label correlations are taken into account.
%whereas local label correlations play important role often \citep{huang2012}. 
Obviously, it would be more desirable to learn both global and local label correlations 
simultaneously.

Manifold regularization \cite{belkin2006manifold} exploits instance similarity by forcing
the predicted values on similar instances to be similar. A similar idea can be adapted to
the label manifold, and so predicted values for correlated labels should be similar.
However, the Laplacian matrix is based on some label similarity or correlation matrix, which
can be hard to specify as discussed in 
Section~\ref{sec:introduction}.
%Instead,  it will be better to learn the laplacian matrix automatically.

%\IEEEPARstart{I}n multi-label learning, we are given a data matrix of $n$ instances $\small X \!=\![\x_1,\dots,\x_n] \!\in\! \mathbb{R}^{d \times n}$ 
%and a label matrix with $l$ class labels $\small Y\!=\![\y_1,\cdots,\y_n]\!\in\! \{-1,0,1\}^{l\times n}$, where $\x_i$ is a feature vector of an instance,  and $\small\y_i=[y_{i1},\dots,y_{il}]^\top$ is the corresponding label vector. 
%$y_{ij}=1$ if $x_i$ belongs to the $j$ th label, and $-1$
%otherwise. We adopt the general setting that both positive and negative labels can be
%missing \cite{goldberg2010transduction,xu2013speedup,Yu2014}, and $y_{ij}=0$ if 
%$x_i$'s
%$j$ th  
%label 
%is missing. Let
%
%$\Omega$ be the set containing indices to the observed labels in $Y$. Our goal is to recover the missing
%labels and predict the unseen instances' labels, by learning from $X$ and $Y$.
%based on $\Omega$ and $X$.

%%%%%%%%%%%%%%%%%%%%%%%%%%%%%%%%%%%%%%%%%%
\section{The Proposed Approach} \label{sec:approach}

In multi-label learning, an instance  can be associated with multiple class labels. Let
$\bm{C}=\{c_1,\dots,c_l\}$ be the class label set of $l$ labels. We denote the feature
vector of an instance by $\x\in \X \subseteq \mathbb{R}^d$, and denote the ground-truth
label vector by $\tilde{\y}\in \Y \subseteq \{-1,1\}^l$, where $[\tilde{\y}]_j= 1$ if
$\x$ is with class label $c_j$, and $-1$ otherwise. As mentioned in Section
\ref{sec:introduction}, instances in the training data may be partially labeled, i.e.,
some labels may be missing. We adopt the general setting that both positive and
negative labels can be missing \cite{goldberg2010transduction,xu2013speedup,Yu2014}.
The observed label vector is denoted $\y$, where $[\y]_j=0$ if class label $c_j$ is not labeled (i.e. it is missing), and $[\y]_j=[\tilde{\y}]_j$ otherwise. Given  the training data $\D=\{(\x_i, \y_i)\}_{i=1}^n$, our goal is to  
learn a mapping function $\Psi: \X\rightarrow\Y$.

%which can predict the label vector of an unseen instance  by $\tilde{\y}_{\text{new}}=\Psi(\x_{\text{new}})$, and recover the label vector of a training instance by $\tilde{\y}_i=\Psi(\x_i), i\in\{1,\cdots,n\}$.

In this paper, we propose the \texttt{GLOCAL} algorithm,
which learns and exploits both global and local label correlations via label manifolds.
To recover the missing labels,
learning of the latent label representation and 
classifier training
are performed simultaneously.

%%%%%%%%%%%%%%%%%%%%%%%%%%%%%%%%%%%%%%%%%%

\subsection{Basic Model}

Let
$\tilde{\bm{Y}}\!=\![\tilde{\y}_1,\dots,\tilde{\y}_n]\!\in\!\{-1,1\}^{l\times n}$ be the
ground-truth label matrix, where each 
$\tilde{\y}_i$ is the label vector for instance $i$.
As discussed in Section~\ref{sec:introduction}, $\tilde{\bm{Y}}$ is low-rank. 
Let its rank be $k<l$. Thus,
%use the product of two rank-$k$ matrices to
$\tilde{\bm{Y}}$ 
can be written as the low-rank decomposition 
$\bm{UV}$, where $\bm{U} \in \mathbb{R}^{l\times k}$  and $\bm{V} \in
\mathbb{R}^{k\times n}$.
Intuitively, 
$\bm{V}$ represents the latent labels that are more compact and more semantically abstract 
than the original labels, while matrix $\bm{U}$ projects the original labels to the latent label space. 

In general, the labels are only partially observed.
Let the observed label matrix
be $\bm{Y}\!=\![\y_1,\dots,\y_n]\!\in\!\{-1,0,1\}^{l\times n}$, and
$\Omega$ be the set containing indices of the observed labels in $\bm{Y}$ (i.e., 
indices of the nonzero elements in $\bm{Y}$). 
We focus on minimizing the
reconstruction error on the observed labels, i.e., $\|\Pi_\Omega(\bm{Y-UV})\|_F^2$, where
$[\Pi_\Omega(\bm{A})]_{ij}=
A_{ij}$  if $\left(i, j\right) \in \Omega$,
and 0 otherwise.
%is the latent representation of $Y$. 
%Intuitively, this latent label representation captures the most information of the original labels along with their correlated ones.
%With the latent representation $V$, 
Moreover, we use
a linear mapping 
$\bm{W} \in \mathbb{R}^{d\times k}$ to map instances to the latent labels.
This $\bm{W}$ 
is learned
by minimizing $\|\bm{V-W}^\top \bm{X}\|_F^2$, where
$\bm{X}=[\x_1,\dots,\x_n]\in \mathbb{R}^{d\times n}$ is the instance matrix.
%Given $\Omega$, in order to focus on the observed labels, we define a linear operator $\mathcal{R}_\Omega: \mathbb{R}^{l\times n}\mapsto \mathbb{R}^{l\times n}$ as
%Combining all above together, minimizing the squared loss between $Y$ and $UV$ on the observed labels and that between $V$ and $W^\top X$, we have
Combining these two, we obtain the following optimization problem:
\begin{equation} \label{eqn:init}
\min_{\bm{U,V,W}}  \|\Pi_\Omega(\bm{Y-UV})\|_F^2 + \lambda\|\bm{V-W}^\top \bm{X}\|_F^2 + \lambda_2\mathcal{R}(\bm{U,V,W}),
\end{equation}
where
$\mathcal{R}(\bm{U,V,W})$ is a regularizer  and $\lambda$, $\lambda_2$ are tradeoff parameters.
While
the square loss 
has been used in
Eqn \eqref{eqn:init}, it can be replaced
by any differentiable loss function.
%The prediction on  $X$ is
%$\mathrm{sign}(\bm{F}_0)$, where
%$\bm{F}_0 = \bm{UW}^\top \bm{X}$.
The prediction on $\bm{\x}$ is $\mathrm{sign}(\f(\x))$, where $\f(\x)= \bm{UW}^\top \x$. Let $\f=[f_1,\cdots, f_l]^\top$, thus
$f_j(\x)$ denotes the predictive value on $j$-th label for $\x$. We concatenate all $\f(\x), \forall \x \in \bm{X}$, denoted by $F_0$, thus $F_0 = [\f(\x_1),\cdots, \f(\x_n)] =  \bm{UW}^\top \bm{X}$.

%We additionally regularize $V$; besides, learning the latent representation (the first term) and the linear mapping (the second term) can be with different importance ($\lambda$). As a result, this optimization is not equivalent to $\min_{U,W}\|\Pi_\Omega(Y-UW^\top X)\|_F^2+ \lambda_2\mathcal{R}(U,W)$. 

%%%%%%%%%%%%%%%%%%%%%%%%%%%%%%%%%%%%%%%%%

\subsection{Global and Local Manifold Regularizers}

Exploiting label correlations is an essential ingredient in multi-label learning. Here, 
we use label correlations to regularize the model. Intuitively, the more positively
correlated two labels are, the closer are the corresponding classifier outputs, and
vice versa. 
%Based on this idea, we design the label manifold regularizers.
Let $\bm{S}_0 =[S_{ij}]\in
\mathbb{R}^{l\times l}$ be the global label correlation matrix. 
The manifold regularizer 
%$\sum_{(i,j)\in [l]\times[l]} 
$\sum_{i,j} S_{ij} \|\bm{f}_{i,:}-\bm{f}_{j,:}\|_2^2 $ should have a small value \cite{melacci2011primallapsvm}. Here, $\bm{f}_{i,:}$, the $i$th row
of $\bm{F}_0$, is the vector of classifier outputs for the $i$th label on   the $n$ samples.
Let $\bm{D}_0$ be the diagonal matrix with diagonal $\bm{S}_0\mathbf{1}$, where $\mathbf{1}$ is the
vector of ones.
The manifold regularizer can be equivalently written as $\mathrm{tr}(\bm{F}_0^\top \bm{L}_0^{} \bm{F}_0^{})$ \cite{luo2009non}, 
where  $\bm{L}_0 = \bm{D}_0-\bm{S}_0$ is the Laplacian matrix of $\bm{S}_0$. 
%if the input is a vector.
%or return a the main diagonals of the input if the input is a matrix.

As discussed in Section~\ref{sec:introduction}, label correlations may vary from one local region to
another.
Assume that the data $\bm{X}$ is partitioned into $g$ groups $\{\bm{X}_1,\dots, \bm{X}_g\}$, where $\bm{X}_m \in
\mathbb{R}^{d\times n_m}$ has 
size $n_m$.
%subset of $X$, and $m \in \{1,\dots,g\}$.
This partitioning can be obtained by domain knowledge (e.g., 
gene pathways \cite{subramanian2005gene} and networks \cite{chuang2007network} 
in bioinformatics applications)
or
clustering.
%Hence, even if the class labels are missing, the partition will not be affected.
Let $\bm{Y}_m$ be the label submatrix 
in $\bm{Y}$ corresponding to $\bm{X}_m$, and $\bm{S}_m \in \mathbb{R}^{l\times l}$ be the local label correlation matrix of group $m$.
%$(i \in [g])$.
Similar to global label correlation, 
to encourage the classifier outputs to be similar on the positively correlated labels and
dissimilar on the negatively correlated ones,
we minimize $\mathrm{tr}(\bm{F}_m^\top \bm{L}_m^{} \bm{F}_m)$,
where $\bm{L}_m$ is the Laplacian matrix of $\bm{S}_m$ and $\bm{F}_m = \bm{UW}^\top \bm{X}_m$ is the classifier output matrix for group $m$.

Combining global and local label correlations
with Eqn. (\ref{eqn:init}), 
we have the 
following
optimization problem:
\begin{equation} \label{opt:uvw}
\min\limits_{\bm{U,V,W}} \|\Pi_\Omega(\bm{Y\!-\!UV})\|_F^2 \!+ \!\lambda\|\bm{V\!-\!W}^\top \!\bm{X}\|_F^2 
+\!\lambda_2\mathcal{R}(\bm{U},\bm{V},\bm{W})
+\lambda_3\mathrm{tr}(\bm{F}_0^\top \bm{L}_0^{} \bm{F}_0^{})
+\sum_{m=1}^g\lambda_4\mathrm{tr}(\bm{F}_m^\top \bm{L}_m^{} \bm{F}_m^{}),
\end{equation}
where 
%$F_0\!=\! UW^\top \!X$, $F_m \!=\! UW^\top\! X_m$ and 
$\lambda, \lambda_2, \lambda_3, \lambda_4$ are tradeoff parameters.

%Since each local group is a data subset, there is a natural connection between the global and local label correlations. 
Intuitively, a large local group contributes more to the global label
correlations. In particular, 
the following Lemma 
shows that
when the cosine similarity is used to compute $\bm{S}_{ij}$, we have $\bm{S}_0 = \sum_{m=1}^g \frac{n_m}{n} \bm{S}_m$.

\begin{lemma}
	\label{lemma}
	Let 
	$[\bm{S}_0]_{ij} =
	\frac{\y_{i,:}^{}\y_{j,:}^\top}{\|\y_{i,:}\|\|\y_{j,:}\|}$ and
	$[\bm{S}_m]_{ij} = \frac{\y_{m,i,:}^{}\y_{m,j,:}^\top}{\|\y_{m,i,:}\|\|\y_{m,j,:}\|}$,
	%=\frac{1}{n}\y_{i,:}^{}\y_{j,:}^\top$,
	where $\y_{i,:}$ is the $i$th row of $\bm{Y}$,
	and
	%=\frac{1}{n_k}\y_{k,i,:}^{}\y_{k,j,:}^\top$,
	$\y_{m,i,:}$ is the $i$th row of $\bm{Y}_m$. Then,
	$\bm{S}_0= \sum_{m=1}^g \frac{n_m}{n}\bm{S}_m$.
\end{lemma}
In general, when the global label correlation matrix is a linear combination of the local label correlation matrices, the following Proposition shows that the
global label Laplacian matrix is also 
a linear combination of the local label Laplacian matrices with 
the same combination coefficients.

\begin{prop} \label{prop}
	If 
	$\bm{S}_0 \!=\! \sum_{m=1}^g \beta_m \bm{S}_m$, then
	$\bm{L}_0 \!=\! \sum_{m=1}^g \beta_m \bm{L}_m$.
\end{prop}

Using
Lemma \ref{lemma}
and 
Proposition \ref{prop}, Eqn. \eqref{opt:uvw} can then be rewritten as follows:
\begin{eqnarray} \label{opt:uvw2}
&\min\limits_{\bm{U,V,W}}\hspace{-.1in}& \|\Pi_\Omega(\bm{Y\!-\!UV})\|_F^2 + \!\lambda\|\bm{V\!-\!W}^\top \!\bm{X}\|_F^2 \nonumber
+\!\lambda_2\mathcal{R}(\bm{U,V,W})\\
&&+\sum_{m=1}^g\!\!\left(\frac{\lambda_3n_m}{n}\mathrm{tr}(\bm{F}_0^\top \bm{L}_m^{} \bm{F}_0^{})+
\lambda_4\mathrm{tr}(\bm{F}_m^\top \bm{L}_m^{} \bm{F}_m^{})\right)\!.
\end{eqnarray}
%where $F_0= UW^\top X$, $F_m = UW^\top X_m$ and $\lambda, \lambda_2, \lambda_3, \lambda_4$ are tradeoff parameters.

The success of label  manifold regularization hinges on a good correlation matrix (or
equivalently, a good Laplacian matrix). In multi-label learning, one rudimentary approach
is to compute the correlation coefficient between two labels by cosine distance
\cite{wang2009image}.
However, 
this can be noisy since some labels may only have very few positive instances in the training data. 
When labels can be missing,
this computation may even become misleading, since the label distribution of observed labels may be much different from that of the ground-truth label distribution due to the missing labels.

%Moreover, correlation coefficient only captures the pairwise label correlation which may be much more complicated in practice.}

In this paper, instead of  specifying any  correlation metric or  label correlation matrix, we learn the 
Laplacian matrices directly. Note that the Laplacian matrices are symmetric positive
definite. Thus, 
for $m \in \{1,\ldots,g\}$,
we decompose  $\bm{L}_m$ as
%\footnote{*** any ref on low-rank laplacian matrix? \textcolor{red}{In \cite{chung1997spectral}, the laplacian matrix can be decomposed as $L=ZZ^\top$, but does not show whether $Z$ is low-rank or not.  \cite{spielman2012} suggests a way of low rank approximation of laplacian matrix by spectral decomposition. There is no reference on low-rank approximation on {$L=ZZ^\top$} where Z is a low-rank matrix.}}
$\bm{Z}_m^{}\bm{Z}_m^\top$, where
$\bm{Z}_m \!\in\! \mathbb{R}^{l\times k}$.
For simplicity, $k$ is set to the dimensionality of the latent representation $\bm{V}$. 
As a result, learning the   Laplacian matrices is transformed to learning 
$\bm{\Z}\equiv \{\bm{Z}_1,\dots, \bm{Z}_g\}$.
Note that optimization w.r.t. $\bm{Z}_m$ may lead to the trivial solution $\bm{Z}_m =\mathbf{0}$.  
To avoid this problem, we add the constraint that the diagonal entries in $\bm{Z}_m^{}\bm{Z}_m^\top$ are 1, for $m \in \{1,\cdots, g\}$. 
This constraint also enables us to obtain
a normalized Laplacian matrix~\cite{chung1997spectral} of $L_m$.

Let $\bm{J}=[ \bm{J}_{ij}]$ be the indicator matrix with $\bm{J}_{ij}= 1$ if $(i, j) \in \Omega$, and 0 otherwise.  $\Pi_\Omega(\bm{Y-UV})$ can be rewritten as the Hadamard product $\bm{J}\circ\bm{(Y-UV})$.
Combining the decomposition of Laplacian matrices and the diagonal constraints of $\bm{Z}_m$, we
obtain the optimization problem as:
\begin{eqnarray}
&\min\limits_{\bm{U,V,W,\Z}} &
\!\!\!\!\!\! \|\bm{J}\!\circ\!(\bm{Y\!-\!UV})\|_F^2
%\|\Pi_\Omega(Y\!-\!UV)\|_F^2 
+ \!\lambda\|\bm{V}\!-\!\bm{W}^\top \!\bm{X}\|_F^2 \nonumber
+\!\lambda_2\mathcal{R}(\bm{U,V,W})\\
&&\!\!\!\!\!\! +\!\sum_{m=1}^g\!\Big(\!\frac{\lambda_3n_m}{n}\!\mathrm{tr}\!\left(\!\bm{F}_0^\top\!\nonumber \bm{Z}_m^{}\!\bm{Z}_m^\top \!\bm{F}_0^{}\!\right) \!+\! \lambda_4\mathrm{tr}(\!\bm{F}_m^\top \!\bm{Z}_m^{}\!\bm{Z}_m^\top\! \bm{F}_m^{}\!)\! \Big) \nonumber\\
&\text{s.t.} &\mathrm{diag}(\bm{Z}_m^{}\bm{Z}_m^\top) = \mathbf{1} , m \in \{1, 2, \dots, g\}.
\label{eqn:opt}
\end{eqnarray}
Moreover, we will use 
$\mathcal{R}(\bm{U,V,W})=\|\bm{U}\|_F^2+\|\bm{V}\|_F^2+\|\bm{W}\|_F^2$.
%where $F_0 = UW^\top X$, $F_m = UW^\top X_m$.
%where $\Z=[Z_1, Z_2, \dots, Z_g]$.
\subsection{Learning by Alternating Minimization}

Problem (\ref{eqn:opt})
can be solved
by alternating minimization
(Algorithm~\ref{alg:algo}).
In each iteration, we 
update one of
the variables in $\{\bm{Z}, \bm{U}, \bm{V}, \bm{W}\}$
with gradient descent, and leave
the others fixed.  Specifically, the MANOPT toolbox \cite{manopt} is utilized to implement gradient
descent with line search on the Euclidean space for the update of $\bm{U}, \bm{V}, \bm{W}$, and on the manifolds for the update of $\bm{Z}$. 
%with line search is used to solve each subproblem.
%since there is no closed-form solution,

\begin{algorithm}[t]
	\small
	\caption{\texttt{GLOCAL}.}
	\label{alg:algo}
	\textbf{Input}: data matrix $\bm{X}$, label matrix $\bm{Y}$, observation indicator matrix $\bm{J}$, and the group partition\\
	\textbf{Output}: $\bm{U, W, \Z}$.
	%\textbf{Process}:
	\begin{algorithmic}[1]
		\STATE initialize $\bm{U}, \bm{V}, \bm{W}, \bm{\Z}$; 
		\STATE \textbf{repeat}
		\STATE \quad \textbf{for} $m=1,\dots,g$
		\STATE \quad\quad update $\bm{Z}_m$ by solving (\ref{uz});$\verb|//|$Fix $\bm{V}, \bm{U}, \bm{W}$, update $\bm{Z}_m$
		\STATE \quad \textbf{end for}
		\STATE \quad update $\bm{V}$ by solving (\ref{uv}); \qquad$\verb|//|$Fix  $\bm{U}, \bm{W}, \bm{\Z}$, update $\bm{V}$
		\STATE \quad update $\bm{U}$ by solving (\ref{uu}); \qquad$\verb|//|$Fix  $\bm{V}, \bm{W} ,  \bm{\Z}$, update $\bm{U}$
		\STATE \quad update $\bm{W}$ by solving (\ref{uw}); \qquad$\verb|//|$Fix  $\bm{U}, \bm{V},  \bm{\Z}$, update $\bm{W}$
		\STATE \textbf{until} convergence or maximum number of iterations;
		\STATE output $\bm{U}, \bm{W}$, and $\bm{\Z}\equiv \{\bm{Z}_1,\dots,\bm{Z}_g\}$.
	\end{algorithmic}
\end{algorithm}

%The detail information of solving the subproblem w.r.t. $Z_m$, $U$, $V$, $W$  is introduced respectively in the following Sections \ref{sec:uz}-\ref{sec:uw}. 

%%%%%%%%%%%%%%%%%%

\subsubsection{Updating $\bm{Z}_m$}\label{sec:uz}

With $\bm{U}, \bm{V}, \bm{W}$ fixed, problem~(\ref{eqn:opt}) reduces to
\begin{eqnarray}
& \min\limits_{\bm{Z}_m} & \frac{\lambda_{3}n_m}{n}\mathrm{tr}\left(\! \bm{F}_0^\top\!  \bm{Z}_m^{}\bm{Z}_m^\top\!  \bm{F}_0  \right) +
\lambda_4 \mathrm{tr}\left(\!  \bm{F}_m^\top\!  \bm{Z}_m^{}\bm{Z}_m^\top\!  \bm{F}_m  \right)\nonumber \\
&\text{s.t.} & \mathrm{diag}(\bm{Z}_m^{}\bm{Z}_m^\top) = \mathbf{1},
\label{uz}
\end{eqnarray}
for each $m\in \{1,\dots,g\}$.
%where $F = UW^\top X$ and $F_i = UW^\top X_i$.
Due to the constraint $\mathrm{diag}(\bm{Z}_m^{}\bm{Z}_m^\top) = \mathbf{1}$,
it has no closed-form solution, and 
we  will solve it with projected gradient descent. 
%and at each iteration, then project new $Z_m$ to the feasible set.
The gradient of the objective w.r.t.
$\bm{Z}_m$
is
\[
\nabla_{\bm{Z}_m} \!=  \!\frac{\lambda_{3}n_m}{n}U \!\bm{W}^\top \! \bm{X} \!\bm{X}^\top \!
\bm{W} \!\bm{U}^\top \! \bm{Z}_m \!+ \!\lambda_4 \bm{UW}^\top \bm{X}_m^{}\bm{X}_m^\top \bm{WU}^\top \bm{Z}_m.
\]
To satisfy the constraint $\mathrm{diag}(\bm{Z}_m^{}\bm{Z}_m^\top) = \mathbf{1}$, we
project
each row of $\bm{Z}_m$ onto the unit norm ball after each update: 
\[ \z_{m,j}\leftarrow \z_{m,j}/\|\z_{m,j}\|, \]
where $\z_{m,j}$ is the $j$th row of $\bm{Z}_m$.

%%%%%%%%%%%%%%%%%%

\subsubsection{Updating $\bm{V}$}

With $\bm{Z}_m$'s and $\bm{U}, \bm{W}$ fixed, 
problem~(\ref{eqn:opt}) reduces to
\begin{equation} \label{uv}
\min_{\bm{V}}\|\bm{J}\circ(\bm{Y}-\bm{UV})\|_F^2 + \lambda \|\bm{V}-\bm{W}^\top \bm{X}\|_F^2 + \lambda_2\|\bm{V}\|_F^2.
\end{equation}
Notice that each column of $\bm{V}$ is independent to each other, and thus
%in Eqn. \eqref{uv}, 
$\bm{V}$ 
can be solved
column-by-column. 
%which will lead to a closed form solution. 
Let $\j_i$ and $\v_i$ be $i$th column of $\bm{J}$ and $\bm{V}$, respectively. The optimization problem for $\v_i$ can be written as:
\[
\min_{\v_i} \|\mathrm{Diag}(\j_i)\y_i-\mathrm{Diag}(\j_i)\bm{U}\v_i\|^2 + \lambda \|\v_i-\bm{W}^\top \x_i\|^2 + \lambda_2\|\v_i\|^2.
\]
Setting the gradient w.r.t. $\v_i$ to 0, we obtain the following closed-form solution of $\v_i$:
\[
\v_i = \big(\bm{U}^\top \mathrm{Diag}(\j_i)\bm{U} + (\lambda+\lambda_2)\bm{\mathrm{I}}\big)^{-1}\big(\lambda \bm{W}^\top \x_i + \bm{U}^\top \mathrm{Diag}(\j_i)\y_i\big).
\]
This involves computing a matrix inverse for each $i$. 
%and there are $n$ columns in all, which will cost much time. 
If this is expensive, we can use gradient descent instead. 
The gradient 
of the objective in (\ref{uv})
w.r.t. $\bm{V}$ is
\[\nabla_{\bm{V}} = \bm{U}^\top \left(\bm{J}\circ\left(\bm{UV}-\bm{Y}\right)\right) + \lambda(\bm{V}-\bm{W}^\top \bm{X}) +
\lambda_{2}\bm{V}.\]

%%%%%%%%%%%%%%%%%%

\subsubsection{Updating $\bm{U}$}

With $\bm{Z}_m$'s and $\bm{V}, \bm{W}$ fixed, 
problem~(\ref{eqn:opt}) reduces to
\begin{equation}
\min\limits_{\bm{U}}  \|\bm{J}\circ(\bm{Y}\!-\!\bm{UV})\|_F^2+ \lambda_2\|\bm{U}\|_F^2  
\!+\!\sum_{m=1}^g \!\Big(\!\frac{\lambda_{3}n_m}{n} \mathrm{tr}(\bm{F}_0^\top\!
\bm{Z}_m^{}\!\bm{Z}_m^\top\! \bm{F}_0) \!+\!\lambda_4  \mathrm{tr}(\bm{F}_m^\top \!\bm{Z}_m^{}\!\bm{Z}_m^\top
\!\bm{F}_m)\!\Big)\!.
\label{uu}
\end{equation}
%where $F = UW^\top X$ and $F_i = UW^\top X_i$.
Again, we use gradient descent, and
the gradient w.r.t. $\bm{U}$ is:
\begin{equation*}
\nabla_{\bm{U}}\!=  \!(\bm{J}\circ(\bm{UV}-\bm{Y}))\bm{V}^\top + \lambda_2\bm{U}
\! +\sum_{m=1}^g \!\! \bm{Z}_i^{}\bm{Z}_i^\top \bm{U}\Big(\frac{\lambda_{3}n_m}{n} \bm{W}^\top
\bm{X}_m^{}\bm{X}_m^\top \bm{W} \!+\! \lambda_4 \bm{W}^\top \bm{XX}^\top \bm{W} \Big).
\end{equation*}
%%%%%%%%%%%%%%%%%%

\subsubsection{Updating $\bm{W}$} \label{sec:uw}

With $\bm{Z}_m$'s and $\bm{U}, \bm{V}$ fixed, 
problem~(\ref{eqn:opt}) reduces to
\begin{equation}
\min\limits_{\bm{W}}\lambda\|\bm{V}-\bm{W}^\top \bm{X}\|_F^2+\lambda_2\|\bm{W}\|_F^2
\!+\!\sum_{m=1}^g\!\left(\!\frac{\lambda_{3}n_m}{n}\mathrm{tr}(\bm{F}_0^\top \!\bm{Z}_m^{}\!\bm{Z}_m^\top \!\bm{F}_0^{}) \!+\!
\lambda_4\mathrm{tr}(\bm{F}_m^\top \!\bm{Z}_m^{}\!\bm{Z}_m^\top \!\bm{F}_m^{})\!\right)\!. \label{uw}
\end{equation}
%where $F = UW^\top X$ and $F_i = UW^\top X_i$.
The gradient w.r.t. $W$ is:
\begin{equation*}
\nabla_{\bm{W}} \hspace{-.1in}=\lambda \bm{X}\left(\bm{X}^\top \bm{W}- \bm{V}^\top\right) + \lambda_2\bm{W}
+  \sum_{m=1}^g \Big(\frac{\lambda_{3}n_m}{n} \bm{XX}^\top +\lambda_4 \bm{X}_m^{}\bm{X}_m^\top \Big)\bm{WU}^\top \bm{Z}_m^{}\bm{Z}_m^\top \bm{U}.
\end{equation*}
\begin{table*}[t]
	\tiny
	\centering
	\caption{Datasets used in the experiments  (``$\#$instance" is the number of instances,
		``$\#$dim"
		is the feature dimensionality, ``$\#$label" is the total size of the class label set, and
		``$\#$label/instance" is the average number of labels possessed by each instance).}
	\label{tbl:data}
	\begin{tabular}{l|cccc|l|cccc} \hline
		& \#instance & \#dim & \#label &  \#label/instance
		&&\#instance & \#dim & \#label &  \#label/instance\\ \hline
		Arts         & 5,000 &  462   &  26   &  1.64 
		&Business      & 5,000 &  438   &  30   &  1.59\\
		Computers     & 5,000 &  681   &  33   & 1.51 
		&Education& 5,000 &  550   &  33   & 1.46 \\
		Entertainment& 5,000 &  640   &  21   & 1.42 
		&Health& 5,000 &  612   &  32   &  1.66\\
		Recreation& 5,000 &  606   &  22   &  1.42
		&Reference& 5,000 &  793   &  33   &  1.17\\
		Science& 5,000 &  743   &  40   &  1.45
		&Social & 5,000 &  1,047   &  39   &  1.28\\
		Society& 5,000 &  636   &  27   &  1.69
		&Enron         & 1,702 &  1,001   &   53   &  3.37\\
		Corel5k   & 5,000 &  499   &   374   &  3.52
		&Image         & 2,000 &  294   &   5   &  1.24\\ 
		\hline
	\end{tabular}
\end{table*}
\section{Experiments} \label{sec:exp}

In this section, extensive experiments are performed on text and image datasets. Performance
on both the full-label and missing-label cases are discussed.  

\subsection{Setup}

%%%%%%%%%%%%%%%%%%%%%%%%%%%%%%%%

\subsubsection{Data sets}

On text, eleven
Yahoo datasets\footnote{\url{http://www.kecl.ntt.co.jp/as/members/ueda/yahoo.tar}}
(Arts, Business, Computers, Education, Entertainment, Health, Recreation,
Reference, Science, Social and Society) and 
the Enron dataset\footnote{\url{http://mulan.sourceforge.net/data sets-mlc.html}} are
used. On images,  the
Corel5k\footnotemark[3] and
Image\footnote{\url{http://cse.seu.edu.cn/people/zhangml/files/Image.rar}}   datasets
are used. 
In the sequel, each dataset is denoted by its first three letters.\footnote{``Society"
	is denoted
	``Soci'',  so as to distinguish it from
	``Social".}
Detailed information of the datasets are shown in Table \ref{tbl:data}. For each
dataset, we randomly select $60\%$ of the instances for training, and the rest for testing.
%In the missing-label experiments, for each class label, we randomly sample $\rho \%$ of the labels as observed and the rest as missing. 

%%%%%%%%%%%%%%%%%%%%%%%%%

\subsubsection{Baselines}

In the \texttt{GLOCAL} algorithm,
we use the \texttt{kmeans} clustering algorithm to partition the data into local groups.
The solution of Eqn. \eqref{eqn:init} is used to warm-start $\bm{U}, \bm{V}$ and $\bm{W}$. 
The $\bm{\Z}_m$'s are
randomly initialized.
\texttt{GLOCAL} 
is compared
with the following state-of-the-art multi-label learning algorithms:
%(all parameters are tuned by cross-validation).
%All are in Matlab  (with
%some C++ code 
%for \texttt{LEML}).

%For group partition,
%The number of examples is n = 5,000, the number of dimensions varies from 438 to 1,047 and the number of labels vary varies 21 to 41. Detailed information of the data sets\footnote{*** put that in a tbl here \textcolor{red}{See tl 1.}} can be found in

%\footnote{*** how did u obtain the groups? better show some expts to show that this will not affect the results significantly. \textcolor{red}{Groups are selected according to the result of kmeans clustering on the feature space. The group number will affect the results to some degree, for example, when it equals to 1, there will be no local information taken into account.}}
%\footnote{*** how to determine the rank of the lap matrix? \textcolor{red}{The same as the dimension of the latent label subspace.}}

%%%%%%%%%%%%%%%%%%%%%%%%%%%%%%%%%%%%%%%%%%

%\subsection{Evaluations}

\begin{enumerate}
	\item \texttt{BR} 
	\cite{boutell2004learning}, which trains a binary linear SVM 
	(using the LIBLINEAR package \cite{REF08a})
	for each label independently;
	%without taking label correlation into account;
	\item \texttt{MLLOC} 
	\cite{huang2012},
	which exploits local label correlations by encoding them into the instance's
	feature representation;
	\item \texttt{LEML} 
	\cite{Yu2014},  which learns a linear instance-to-label mapping with low-rank
	structure, and implicitly takes advantage of global label correlation; 
	\item \texttt{ML-LRC} \cite{xu2014learning}, which learns and exploits low-rank global label correlations for multi-label classification with missing labels.
\end{enumerate}

%On the ability to exploit label correlations, 
Note that \texttt{BR} does not take label correlation into account. \texttt{MLLOC} considers only
local label correlations; \texttt{LEML} implicitly uses global label correlations, whereas \texttt{ML-LRC} models global label correlation directly.
On the ability to handle missing labels, \texttt{BR} and \texttt{MLLOC} can only learn with
full labels. 
%whereas \texttt{LEML} and \texttt{ML-LRC} can learn with missing labels.  
%To evaluate the performance of \texttt{BR} and \texttt{MLLOC} in the missing-label scenario, \texttt{MAXIDE} \cite{xu2013speedup} is  used to first impute the missing labels, and then  apply \texttt{BR} and \texttt{MLLOC}.

For simplicity, we set $\lambda=1$
in \texttt{GLOCAL}.
The other parameters, as well as those of the baseline methods, are selected via 5-fold cross-validation on the training set. 
All the algorithms are implemented in Matlab  (with some C++ code for \texttt{LEML}). 
%For manifold optimization

%involved in \texttt{GLOCAL}, the MANOPT toolbox \cite{manopt} is utilized to implement gradient
%descent with line search. 
%(we adopt the default setting of this tool). 
\subsubsection{Performance Evaluation}

Let $p$ be the 
number of test instances,
$\bm{C}_i^+, \bm{C}_i^-$ be the sets of positive and negative labels associated
with the
$i$th instance; and $\bm{Z}_j^+, \bm{Z}_j^-$ be the sets of positive and
negative instances belonging to the $j$th label. Given 
input $\bm{x}$,
let $\mathrm{rank}_{\bm{f}}(\bm{x}, y)$ be the rank of label $y$ in the predicted label ranking 
(sorted in descending order).
%produced by $f$.
%$i \in \{1, \dots, l\}$) 
For performance evaluation, we use the following popular 
metrics in multi-label learning
\cite{zhang2014review}:
\begin{enumerate}
	\item Ranking loss (Rkl):  This is the fraction that a negative label is ranked
	higher than a positive label. For instance $i$, define
	$\bm{Q}_i=
	\{(j',j'') \; | \;f_{j'}(\x_i)\leq f_{j''}(\x_i),(j', j'')\in \bm{C}_i^+\times
	\bm{C}_i^-\}$. Then,
	\[\text{Rkl} = \frac{1}{p}\sum_{i=1}^p\frac{
		|\bm{Q}_i|
	}{|\bm{C}_i^+||\bm{C}_i^-|}.\]
	\item Average AUC (Auc):  This is the 
	fraction 
	that a positive instance is ranked higher than a negative instance,
	averaged over all labels. Specifically,
	for label $j$, define $\bm{\tilde{Q}}_j= \{(i',i'') \; |\; f_j(\x_{i'})\geq f_j(\x_{i''}),(\x_{i'},\x_{i''})\in
	\bm{Z}_j^+\times \bm{Z}_j^-\}$. Then,
	\[\text{Auc}=\frac{1}{l}\sum_{j=1}^l\frac{
		|\bm{\tilde{Q}}_j|}
	{|\bm{Z}_j^+||\bm{Z}_j^-|}.\]
	\item Coverage (Cvg): This counts how many steps are needed to move down the predicted label
	ranking so as to cover all the positive labels of the instances.
	\[\text{Cvg}=\frac{1}{p}\sum_{i=1}^p \max\{\mathrm{rank}_{\bm{f}}(\x_i, j)\;|\;{j\in
		\bm{C}_i^+}\}-1.\]
	\item Average precision (Ap):
	%on the test\footnote{*** the above 3 measures are also computed on the test set, right?} set, 
	This is the average fraction of positive labels
	ranked higher than a particular positive label.
	For instance $i$, define 
	$\bm{\hat{Q}}_{i,c}=\{j\;|\; \mathrm{rank}_{\bm{f}}(\x_i, j)\leq
	\mathrm{rank}_{\bm{f}}(\x_i,c), j\in \bm{C}_i^+\}$.  Then,
	\[\text{Ap}=\frac{1}{p}\sum_{i=1}^p \frac{1}{|\bm{C}_i^+|}\sum\nolimits_{c\in \bm{C}_i^+}
	\frac{|\bm{\hat{Q}}_{i,c}|}{\mathrm{rank}_{\bm{f}}(\x_i,c)}.\]
\end{enumerate}
For Auc and
Ap, the higher the better; whereas for Rkl and Cvg, the lower the better. To reduce statistical variability,
results are averaged over 10 independent repetitions.
\begin{table*}[!]
	\tiny
	\centering
	\caption{Results for learning with full labels.
		%on  ranking loss(Rkl), average auc(Auc), coverage(Cvg) and average precision(Ap). 
		$\uparrow$ ($\downarrow$) denotes the larger (smaller) the better.  $\bullet$ indicates that \label{tbl:full_r} \texttt{GLOCAL} is significantly better (paired t-tests at 95\% significance level).}
	\begin{tabular}{l|c|ccccc}\hline
		&Measure&\texttt{BR}&\texttt{MLLOC}&\texttt{LEML}& \texttt{ML-LRC}&\texttt{GLOCAL}\\\hline
		\multirow{4}{*}{Arts}
		&Rkl~($\downarrow$)
		&0.201$\pm$0.005$\bullet$
		&0.177$\pm$0.013$\bullet$
		&0.170$\pm$0.005$\bullet$
		&0.157$\pm$0.002$\bullet$
		&\textbf{0.138$\pm$0.002}$~~$\\
		&Auc~($\uparrow$)
		&0.799$\pm$0.006$\bullet$
		&0.823$\pm$0.013$\bullet$
		&0.833$\pm$0.005$\bullet$
		&0.843$\pm$0.001$~~$
		&\textbf{0.846$\pm$0.005}$~~$\\
		&Cvg~($\downarrow$)
		&7.347$\pm$0.196$\bullet$
		&6.762$\pm$0.344$\bullet$
		&6.337$\pm$0.243$\bullet$
		&5.529$\pm$0.037$\bullet$
		&\textbf{5.347$\pm$0.146}$~~$\\
		&Ap~($\uparrow$)
		&0.594$\pm$0.006$\bullet$
		&0.606$\pm$0.006$\bullet$
		&0.590$\pm$0.005$\bullet$
		&0.600$\pm$0.007$\bullet$
		&\textbf{0.619$\pm$0.005}$~~$\\\hline
		\multirow{4}{*}{Business}
		&Rkl~($\downarrow$)
		&0.072$\pm$0.005$\bullet$
		&0.055$\pm$0.009$\bullet$
		&0.056$\pm$0.005$\bullet$
		&0.044$\pm$0.002$~~$
		&\textbf{0.044$\pm$0.002}$~~$\\
		&Auc~($\uparrow$)
		&0.928$\pm$0.005$\bullet$
		&0.944$\pm$0.008$\bullet$
		&0.945$\pm$0.005$\bullet$
		&0.950$\pm$0.005$~~$
		&\textbf{0.955$\pm$0.003}$~~$\\
		&Cvg~($\downarrow$)
		&4.087$\pm$0.268$\bullet$
		&3.265$\pm$0.464$\bullet$
		&3.187$\pm$0.270$\bullet$
		&2.560$\pm$0.059$~~$
		&\textbf{2.559$\pm$0.169}$~~$\\
		&Ap~($\uparrow$)
		&0.861$\pm$0.007$\bullet$
		&0.878$\pm$0.011$\bullet$
		&0.867$\pm$0.007$\bullet$
		& 0.870$\pm$0.005$\bullet$
		&\textbf{0.883$\pm$0.004}$~~$\\\hline
		\multirow{4}{*}{Computers}
		&Rkl~($\downarrow$)
		&0.146$\pm$0.007$\bullet$
		&0.134$\pm$0.014$\bullet$
		&0.138$\pm$0.004$\bullet$
		&0.107$\pm$0.002$~~$
		&\textbf{0.107$\pm$0.002}$~~$\\
		&Auc~($\uparrow$)
		&0.854$\pm$0.007$\bullet$
		&0.866$\pm$0.014$\bullet$
		&0.895$\pm$0.002$~~$
		&0.894$\pm$0.002$~~$
		&\textbf{0.895$\pm$0.002}$~~$\\
		&Cvg~($\downarrow$)&6.654$\pm$0.236$\bullet$
		&6.224$\pm$0.480$\bullet$
		&6.148$\pm$0.183$\bullet$
		&4.893$\pm$0.142$~~$
		&\textbf{4.889$\pm$0.058}$~~$\\
		&Ap~($\uparrow$)&0.680$\pm$0.007$\bullet$
		&0.689$\pm$0.009$\bullet$
		&0.669$\pm$0.007$\bullet$
		& 0.689$\pm$0.005$\bullet$
		&\textbf{0.698$\pm$0.004}$~~$\\\hline
		\multirow{4}{*}{Education}
		&Rkl~($\downarrow$)&0.203$\pm$0.010$\bullet$
		&0.158$\pm$0.021$\bullet$
		&0.145$\pm$0.008$\bullet$
		& 0.099$\pm$0.002$\bullet$
		&\textbf{0.095$\pm$0.002}$~~$\\
		&Auc~($\uparrow$)
		&0.797$\pm$0.102$\bullet$
		&0.842$\pm$0.022$\bullet$
		&0.859$\pm$0.008$\bullet$
		&0.868$\pm$0.006$~~$
		&\textbf{0.878$\pm$0.006}$~~$\\
		&Cvg~($\downarrow$)&8.979$\pm$0.487$\bullet$
		&7.381$\pm$0.765$\bullet$
		&6.711$\pm$0.364$\bullet$
		&4.531$\pm$0.104$~~$
		&\textbf{4.529$\pm$0.206}$~~$\\
		&Ap~($\uparrow$)&0.580$\pm$0.010$\bullet$
		&0.613$\pm$0.004$\bullet$
		&0.596$\pm$0.009$\bullet$
		&0.600$\pm$0.007$\bullet$
		&\textbf{0.628$\pm$0.009}$~~$\\\hline
		\multirow{4}{*}{Entertainment}
		&Rkl~($\downarrow$)&0.185$\pm$0.006$\bullet$
		&0.146$\pm$0.013$\bullet$
		&0.154$\pm$0.005$\bullet$
		&0.130$\pm$0.005$\bullet$
		&\textbf{0.108$\pm$0.004}$~~$\\
		&Auc~($\uparrow$)
		&0.815$\pm$0.006$\bullet$
		&0.854$\pm$0.013$\bullet$
		&0.852$\pm$0.005$\bullet$
		& 0.871$\pm$0.003$~~$
		&\textbf{0.874$\pm$0.005}$~~$\\
		&Cvg~($\downarrow$)&5.006$\pm$0.160$\bullet$
		&4.293$\pm$0.344$\bullet$
		&4.193$\pm$0.139$\bullet$
		&3.505$\pm$0.125$\bullet$
		&\textbf{3.114$\pm$0.110}$~~$\\
		&Ap~($\uparrow$)&0.662$\pm$0.009$\bullet$
		&0.670$\pm$0.005$\bullet$
		&0.647$\pm$0.007$\bullet$
		&0.661$\pm$0.012$\bullet$
		&\textbf{0.681$\pm$0.008}$~~$\\\hline
		\multirow{4}{*}{Health}
		&Rkl~($\downarrow$)&0.113$\pm$0.001$\bullet$
		&0.093$\pm$0.005$\bullet$
		&0.091$\pm$0.003$\bullet$
		&0.071$\pm$0.003$\bullet$
		&\textbf{0.065$\pm$0.002}$~~$\\
		&Auc~($\uparrow$)
		&0.886$\pm$0.003$\bullet$
		&0.907$\pm$0.005$\bullet$
		&0.913$\pm$0.004$\bullet$
		&\textbf{0.929$\pm$0.009}$~~$
		&0.923$\pm$0.007$~~$\\
		&Cvg~($\downarrow$)&6.193$\pm$0.059$\bullet$
		&5.403$\pm$0.157$\bullet$
		&5.063$\pm$0.128$\bullet$
		&\textbf{3.751$\pm$0.128}$~~$
		&3.858$\pm$0.131$~~$\\
		&Ap~($\uparrow$)&0.763$\pm$0.002$\bullet$
		&0.777$\pm$0.004$\bullet$
		&0.750$\pm$0.003$\bullet$
		& 0.755$\pm$0.006$\bullet$
		&\textbf{0.782$\pm$0.001}$~~$\\\hline
		\multirow{4}{*}{Recreation}
		&Rkl~($\downarrow$)&0.197$\pm$0.003$\bullet$
		&0.184$\pm$0.015$\bullet$
		&0.185$\pm$0.001$\bullet$
		& 0.170$\pm$0.004$\bullet$
		&\textbf{0.155$\pm$0.002}$~~$\\
		&Auc~($\uparrow$)
		&0.802$\pm$0.003$\bullet$
		&0.816$\pm$0.015$\bullet$
		&0.822$\pm$0.002$\bullet$
		&0.833$\pm$0.004$\bullet$
		&\textbf{0.840$\pm$0.000}$~~$\\
		&Cvg~($\downarrow$)&5.506$\pm$0.089$\bullet$
		&5.268$\pm$0.333$\bullet$
		&5.110$\pm$0.040$\bullet$
		& 4.515$\pm$0.045$\bullet$
		&\textbf{4.431$\pm$0.048}$~~$\\
		&Ap~($\uparrow$)&0.609$\pm$0.005$\bullet$
		&0.620$\pm$0.004$\bullet$
		&0.595$\pm$0.004$\bullet$
		&0.604$\pm$0.003$\bullet$
		&\textbf{0.625$\pm$0.004}$~~$\\\hline
		\multirow{4}{*}{Reference}
		&Rkl~($\downarrow$)&0.155$\pm$0.005$\bullet$
		&0.138$\pm$0.008$\bullet$
		&0.137$\pm$0.004$\bullet$
		& 0.092$\pm$0.003$\bullet$
		&\textbf{0.086$\pm$0.003}$~~$\\
		&Auc~($\uparrow$)
		&0.845$\pm$0.005$\bullet$
		&0.862$\pm$0.008$\bullet$
		&0.872$\pm$0.004$\bullet$
		& \textbf{0.900$\pm$0.006}$~~$
		&0.894$\pm$0.004$~~$\\
		&Cvg~($\downarrow$)&6.171$\pm$0.219$\bullet$
		&5.514$\pm$0.309$\bullet$
		&5.277$\pm$0.171$\bullet$
		& 3.438$\pm$0.133$~~$
		&\textbf{3.387$\pm$0.118}$~~$\\
		&Ap~($\uparrow$)&0.685$\pm$0.005$\bullet$
		&\textbf{0.688$\pm$0.003}$~~$
		&0.667$\pm$0.003$\bullet$
		&0.667$\pm$0.007$\bullet$
		&\textbf{0.688$\pm$0.007}$~~$\\\hline
		\multirow{4}{*}{Science}
		&Rkl~($\downarrow$)&0.197$\pm$0.009$\bullet$
		&0.166$\pm$0.017$\bullet$
		&0.170$\pm$0.005$\bullet$
		& 0.131$\pm$0.002$\bullet$
		&\textbf{0.118$\pm$0.003}$~~$\\
		&Auc~($\uparrow$)
		&0.802$\pm$0.010$\bullet$
		&0.834$\pm$0.018$~~$
		&0.834$\pm$0.005$\bullet$
		&\textbf{0.860$\pm$0.003}$~~$
		&0.853$\pm$0.010$~~$\\
		&Cvg~($\downarrow$)&10.189$\pm$0.435$\bullet$
		&8.867$\pm$0.751$\bullet$
		&8.885$\pm$0.197$\bullet$
		& 6.704$\pm$0.122$\bullet$
		&\textbf{6.434$\pm$0.137}$~~$\\
		&Ap~($\uparrow$)&0.568$\pm$0.012$\bullet$
		&\textbf{0.581$\pm$0.009}$~~$
		&0.551$\pm$0.008$\bullet$
		& 0.561$\pm$0.009$\bullet$
		&0.580$\pm$0.009$~~$\\\hline
		\multirow{4}{*}{Social}
		&Rkl~($\downarrow$)
		&0.112$\pm$0.001$\bullet$
		&0.094$\pm$0.013$\bullet$
		&0.106$\pm$0.006$\bullet$
		&0.075$\pm$0.005$~~$
		&\textbf{0.075$\pm$0.005}$~~$\\
		&Auc~($\uparrow$)
		&0.888$\pm$0.002$\bullet$
		&0.906$\pm$0.013$\bullet$
		&0.894$\pm$0.006$\bullet$
		&\textbf{0.917$\pm$0.005}$~~$
		&0.915$\pm$0.005$~~$\\
		&Cvg~($\downarrow$)
		&6.036$\pm$0.125$\bullet$
		&5.147$\pm$0.401$\bullet$
		&5.521$\pm$0.301$\bullet$
		&4.651$\pm$0.102$~~$
		&\textbf{4.537$\pm$0.258}$~~$\\
		&Ap~($\uparrow$)
		&0.724$\pm$0.005$\bullet$
		&\textbf{0.764$\pm$0.008}$~~$
		&0.731$\pm$0.005$\bullet$
		&0.719$\pm$0.003$\bullet$
		&0.758$\pm$0.008$~~$\\\hline
		\multirow{4}{*}{Society}
		&Rkl~($\downarrow$)&0.204$\pm$0.004$\bullet$
		&0.182$\pm$0.006$\bullet$
		&0.182$\pm$0.007$\bullet$
		& 0.142$\pm$0.002$\bullet$
		&\textbf{0.136$\pm$0.005}$~~$\\
		&Auc~($\uparrow$)
		&0.796$\pm$0.005$\bullet$
		&0.818$\pm$0.006$\bullet$
		&0.822$\pm$0.008$\bullet$
		&0.840$\pm$0.006$~~$
		&\textbf{0.844$\pm$0.006}$~~$\\
		&Cvg~($\downarrow$)&8.048$\pm$0.108$\bullet$
		&7.392$\pm$0.216$\bullet$
		&7.438$\pm$0.162$\bullet$
		& 5.973$\pm$0.108$~~$
		&\textbf{5.852$\pm$0.194}$~~$\\
		&Ap~($\uparrow$)&0.610$\pm$0.007$\bullet$
		&0.623$\pm$0.004$\bullet$
		&0.599$\pm$0.006$\bullet$
		& 0.605$\pm$0.006$\bullet$
		&\textbf{0.633$\pm$0.009}$~~$\\\hline
		\multirow{4}{*}{Enron}
		&Rkl~($\downarrow$)
		&0.194$\pm$0.006$\bullet$
		&0.169$\pm$0.012$\bullet$
		&0.159$\pm$0.005$\bullet$
		&0.133$\pm$0.004$\bullet$
		&\textbf{0.125$\pm$0.004}$~~$\\
		&Auc~($\uparrow$)
		&0.806$\pm$0.006$\bullet$
		&0.831$\pm$0.009$\bullet$
		&0.851$\pm$0.006$\bullet$
		&0.869$\pm$0.004$\bullet$
		&\textbf{0.877$\pm$0.005}$~~$\\
		&Cvg~($\downarrow$)
		&23.618$\pm$0.450$\bullet$
		&21.724$\pm$0.950$\bullet$
		&18.531$\pm$0.707$\bullet$
		&\textbf{16.654$\pm$0.198}$~~$
		&16.737$\pm$0.622$~~$\\
		&Ap~($\uparrow$)
		&0.575$\pm$0.006$\bullet$
		&0.586$\pm$0.009$\bullet$
		&0.600$\pm$0.004$\bullet$
		&0.591$\pm$0.004$\bullet$
		& \textbf{0.647$\pm$0.006}$~~$\\\hline
		\multirow{4}{*}{Corel5k}
		&Rkl~($\downarrow$)
		&0.271$\pm$0.006$\bullet$
		&0.230$\pm$0.012$\bullet$
		&0.246$\pm$0.004$\bullet$
		&\textbf{0.170$\pm$0.002}~~~
		&0.173$\pm$0.005~~~\\
		&Auc~($\uparrow$)
		&0.699$\pm$0.006$\bullet$
		&0.757$\pm$0.012$\bullet$
		&0.754$\pm$0.005$\bullet$
		&0.825$\pm$0.005$~~$
		&\textbf{0.827$\pm$0.005}~~~\\
		&Cvg~($\downarrow$)
		&261.99$\pm$3.15$\bullet$
		&201.80$\pm$6.71$\bullet$
		&184.58$\pm$1.72$\bullet$
		&137.31$\pm$2.49~~~
		&\textbf{136.91$\pm$3.21}~~~\\
		&Ap~($\uparrow$)
		&0.153$\pm$0.001$\bullet$
		&0.182$\pm$0.005$\bullet$
		&0.188$\pm$0.004$\bullet$
		&0.198$\pm$0.003~~~
		&\textbf{0.200$\pm$0.004}~~~\\\hline
		\multirow{4}{*}{Image}
		&Rkl~($\downarrow$)&0.181$\pm$0.011~~~
		&0.180$\pm$0.008~~~
		&0.181$\pm$0.012~~~
		&0.180$\pm$0.009~~~
		&\textbf{0.179$\pm$0.004}~~~\\
		&Auc~($\uparrow$)
		&0.812$\pm$0.011~~~
		&0.810$\pm$0.012~~~
		&0.786$\pm$0.005$\bullet$
		&0.748$\pm$0.010$\bullet$
		&\textbf{0.819$\pm$0.009}~~~\\
		&Cvg~($\downarrow$)&1.004$\pm$0.050~~~
		&\textbf{0.975$\pm$0.060}~~~
		&1.000$\pm$0.027~~~
		&1.000$\pm$0.019~~~
		&\textbf{0.975$\pm$0.054}~~~\\
		&Ap~($\uparrow$)&0.788$\pm$0.008~~~
		&0.794$\pm$0.010~~~
		&0.790$\pm$0.008~~~
		&0.790$\pm$0.010~~~
		&\textbf{0.795$\pm$0.007}~~~\\\hline
	\end{tabular}
\end{table*}

%\subsection{Results}
\subsection{Learning with Full Labels} 

In this experiment, all elements in the training label matrix are observed.
Performance 
%(ranking loss, average AUC, coverage and average precision) 
on the test data 
%on ranking loss, average auc, coverage and average precision respectively and results 
is shown in Table~\ref{tbl:full_r}.
%both the average result and standard deviation are reported.
%where mean and standard diviation over 10 runs are reported and 
As expected, \texttt{BR} is the worst , since it treats each label independently without considering label correlations.
\texttt{MLLOC} only considers local label correlations and \texttt{LEML} only makes use of
the low-rank structure. Though \texttt{ML-LRC} takes advantage of both the low-rank
structure and label correlations, only global label correlations are considered.
%whereas label correlations may differ locally.
As a result, 
\texttt{GLOCAL} is the best
overall, 
as
it models both global and local label correlations.

\begin{figure}[!]
	\begin{center}
		\subfigure[Group 1. \label{fig:img1}]{\includegraphics[width=0.9\textwidth,height=0.2\textwidth]{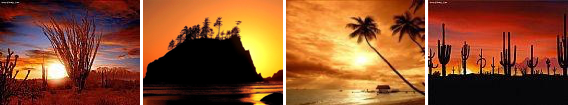}}
		
		\subfigure[Group 2. \label{fig:img2}]{\includegraphics[width=0.9\textwidth,height=0.2\textwidth]{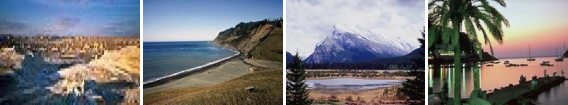}}
		
		\subfigure[Local (group 1). \label{fig:local1}]{\includegraphics[width=0.3\textwidth,height=0.3\textwidth]{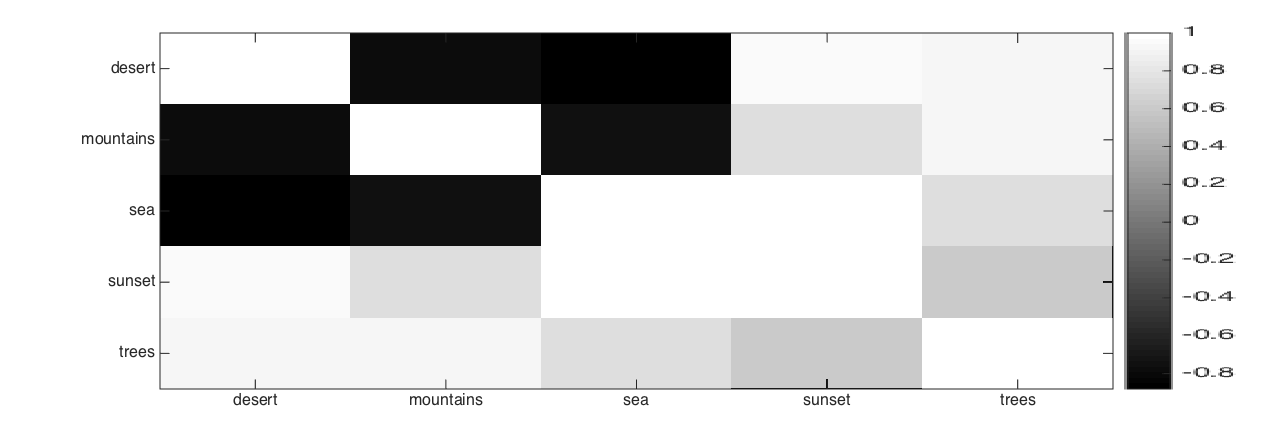}}
		\subfigure[Local (group 2). \label{fig:local2}]{\includegraphics[width=0.3\textwidth,height=0.3\textwidth]{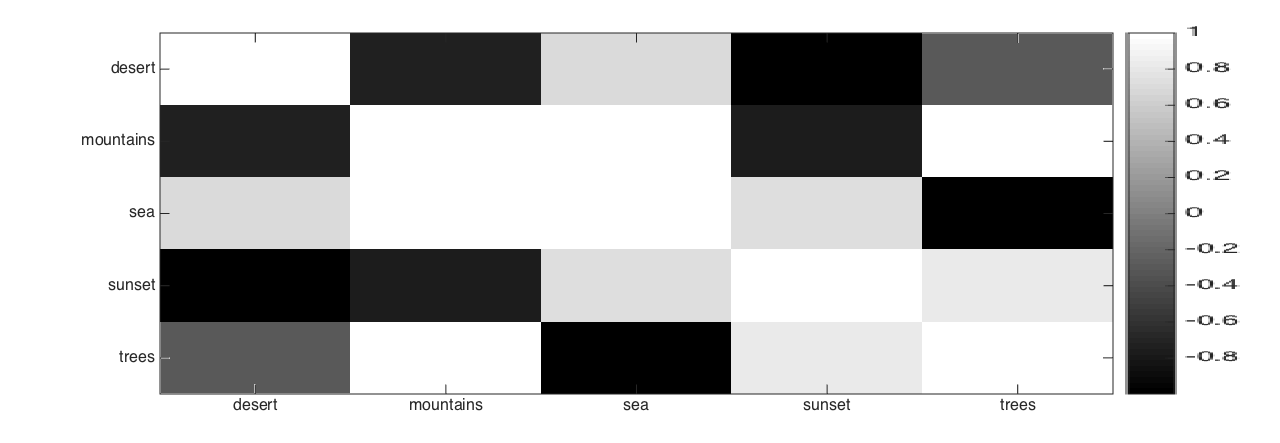}}
		\subfigure[Global. \label{fig:global}]{ \includegraphics[width=0.3\textwidth,height=0.3\textwidth]{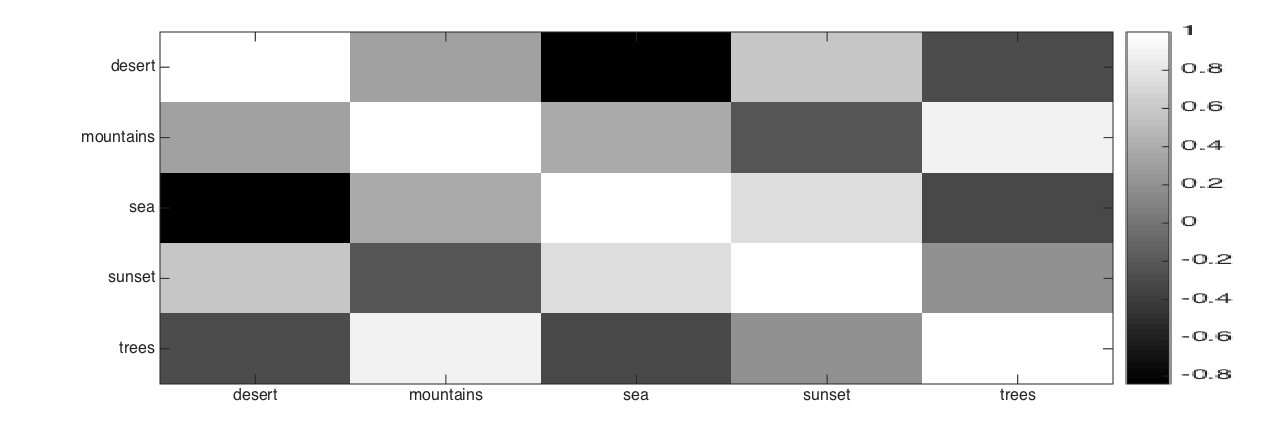}}
		\caption{Example images from two local groups in the \emph{Image} data set, and the
			corresponding $5\times 5$ label correlation matrices.  The labels are (top-to-down,
			left-to-right) ``desert'', ``mountains", ``sea", ``sunset" and ``trees". 
			%The correlation ranges from -1 to 1.
			%corresponding to different gray level. Specifically, -1 corresponds to black, which suggests two labels are strongly negatively correlated; 1 corresponds to white, which suggests two labels are strongly positively correlated.  In (e) for example,  as we can observed that, ``sea'' and ``sunset'', ``mountain'' and ``trees'' are positively correlated, whereas ``desert'' and ``sea'', ``desert'' and ``trees'' are negatively correlated, which is consistent with our common sense.  Compare (c), (d) and (e), we can find that, in  certain local groups, local label correlation  does vary from group to group, and can be very different from the global one. 
		}
		\label{fig:sample}
	\end{center}
\end{figure}

To show the example correlations learned by \texttt{GLOCAL}, we use two local groups
extracted from the Image dataset.
Figure~\ref{fig:sample} shows that local label correlation does vary from group to group,
and is different from global correlation. 
For group 1,
``sunset" is highly correlated with ``desert" and ``sea"
%(suggesting high co-occurrence)
(Figure~\ref{fig:local1}).
This can also be seen from 
the images in Figure~\ref{fig:img1}.
%where sunset is appearing in many scenes including sea and desert.
Moreover,
``trees" sometimes co-occurs with ``deserts" (first and last images in Figure~\ref{fig:img1}).
%(e.g. the first and the last picture in Figure \ref{fig:img1}), which leads to a high weight of tree-desert in the label correlation matrix (Figure \ref{fig:local1}). 
However, in group 2 (Figure~\ref{fig:local2}), 
%the frequency of presence of sunset is much lower than group a but
``mountain" and ``sea" 
often occur together and ``trees" occurs less often with ``desert" (Figure~\ref{fig:img2}). 
Figure \ref{fig:global} shows the learned global label correlation: ``sea'' and ``sunset'',
``mountain'' and ``trees'' are positively correlated, whereas ``desert'' and ``sea'',
``desert'' and ``trees'' are negatively correlated. All these correlations are consistent with 
intuition.

To further validate  the effectiveness of global
and local label correlations, we study
two degenerate versions of \texttt{GLOCAL}: (i) \texttt{GLObal},
which uses only global label correlations; and (ii)
\texttt{loCAL}, which uses only local label correlations. Note that the local groups obtained by clustering are not of
equal sizes. For some datasets, the largest cluster contains more than $40\%$ of instances,
while
some small ones contain fewer than
$5\%$ each. 
%and Lemma~\ref{lemma}, 
Global correlation is then dominated by the local correlation matrix of the largest
cluster (Proposition~\ref{prop}), making
the performance difference on the whole test set obscure.
Hence, we focus on the performance of the
small clusters. 
As can be seen from Table~\ref{tbl:small_cluster}, 
using only global or local correlation may be good enough on some data sets (such as Health).
On the other hand,
considering both types of correlation as in
\texttt{GLOCAL} 
achieves comparable or even better performance.

\begin{table*}[t]
	\tiny
	\centering
	\caption{Results for learning with full labels on the small clusters (each containing fewer than $5\%$ of the samples).  $\uparrow$ ($\downarrow$) denotes the larger (smaller) the better.  $\bullet$ indicates that \texttt{GLOCAL} is significantly better (paired t-tests at 95\% significance level).
		%based on top-3 precision (Tp3), ranking loss (Rkl), coverage (Cvg) and average precision (Ap). 
	}
	\label{tbl:small_cluster} 
	\begin{tabular}{l|c|ccc|l|c|ccc}\hline
		&&\texttt{GLObal}&\texttt{loCAL}&\texttt{GLOCAL}&
		&&\texttt{GLObal}&\texttt{loCAL}&\texttt{GLOCAL}\\\hline
		
		\multirow{4}{*}{Art}
		&Rkl~($\downarrow$)
		&0.137$\pm$0.003$\bullet$
		&0.137$\pm$0.002$\bullet$
		&\textbf{0.130$\pm$0.005}$~~$
		&\multirow{4}{*}{Bus}
		&Rkl~($\downarrow$)
		&0.040$\pm$0.002$~~$
		&0.040$\pm$0.002$~~$
		&\textbf{0.040$\pm$0.003}$~~$\\
		&Auc~($\uparrow$)
		&0.863$\pm$0.003$\bullet$
		&0.863$\pm$0.002$\bullet$
		&\textbf{0.870$\pm$0.005}$~~$
		&
		&Auc~($\uparrow$)
		&0.958$\pm$0.003$~~$
		&0.958$\pm$0.003$~~$
		&\textbf{0.958$\pm$0.003}$~~$\\
		&Cvg~($\downarrow$)
		&5.286$\pm$0.046$\bullet$
		&5.286$\pm$0.046$\bullet$&\textbf{5.197$\pm$0.065}$~~$
		&
		&Cvg~($\downarrow$)
		&2.529$\pm$0.035$~~$
		&2.528$\pm$0.040$~~$
		&\textbf{2.528$\pm$0.040}$~~$\\
		&Ap~($\uparrow$)
		&0.602$\pm$0.013$\bullet$
		&0.602$\pm$0.010$\bullet$
		&\textbf{0.631$\pm$0.011}$~~$
		&
		&Ap~($\uparrow$)
		&0.882$\pm$0.002$\bullet$
		&0.882$\pm$0.002$\bullet$
		&\textbf{0.886$\pm$0.003}$~~$\\\hline
		\multirow{4}{*}{Com}
		&Rkl~($\downarrow$)
		&0.095$\pm$0.002$\bullet$
		&0.095$\pm$0.002$\bullet$
		&\textbf{0.092$\pm$0.002}$~~$
		&\multirow{4}{*}{Edu}
		&Rkl~($\downarrow$)
		&0.101$\pm$0.002$\bullet$
		&0.101$\pm$0.002$\bullet$
		&\textbf{0.097$\pm$0.002}$~~$\\
		&Auc~($\uparrow$)
		&0.905$\pm$0.002$\bullet$
		&0.905$\pm$0.002$\bullet$
		&\textbf{0.908$\pm$0.001}$~~$				
		&
		&Auc~($\uparrow$)
		&0.899$\pm$0.002$\bullet$
		&0.899$\pm$0.002$\bullet$
		&\textbf{0.903$\pm$0.002}$~~$\\
		&Cvg~($\downarrow$)
		&4.482$\pm$0.032$\bullet$
		&4.486$\pm$0.040$\bullet$
		&\textbf{4.364$\pm$0.055}$~~$
		&
		&Cvg~($\downarrow$)
		&4.803$\pm$0.033$\bullet$
		&4.805$\pm$0.036$\bullet$
		&\textbf{4.672$\pm$0.051}$~~$\\
		&Ap~($\uparrow$)
		&0.677$\pm$0.003$~~$
		&0.676$\pm$0.003$~~$
		&\textbf{0.678$\pm$0.005}$~~$
		&
		&Ap~($\uparrow$)
		&0.605$\pm$0.003$\bullet$
		&0.605$\pm$0.003$\bullet~$
		&\textbf{0.624$\pm$0.005}$~~$\\\hline
		\multirow{4}{*}{Ent}
		&Rkl~($\downarrow$)
		&0.091$\pm$0.002$\bullet$
		&0.091$\pm$0.002$\bullet$
		&\textbf{0.086$\pm$0.003}$~~$
		&\multirow{4}{*}{Hea}
		&Rkl~($\downarrow$)
		&0.054$\pm$0.002$~~$
		&0.054$\pm$0.003$~~$
		&\textbf{0.053$\pm$0.004}$~~$\\	
		&Auc~($\uparrow$)
		&0.909$\pm$0.002$\bullet$
		&0.909$\pm$0.002$\bullet$
		&\textbf{0.914$\pm$0.002}$~~$
		&
		&Auc~($\uparrow$)
		&0.945$\pm$0.003$~~$
		&0.946$\pm$0.003$~~$
		&\textbf{0.947$\pm$0.003}$~~$\\
		&Cvg~($\downarrow$)
		&2.817$\pm$0.027$\bullet$
		&2.797$\pm$0.035$\bullet$
		&\textbf{2.709$\pm$0.059}$~~$
		&
		&Cvg~($\downarrow$)
		&3.508$\pm$0.036$~~$
		&3.506$\pm$0.049$~~$
		&\textbf{3.504$\pm$0.041}$~~$\\
		&Ap~($\uparrow$)
		&0.748$\pm$0.003$\bullet$
		&0.749$\pm$0.004$\bullet$
		&\textbf{0.759$\pm$0.006}$~~$
		&
		&Ap~($\uparrow$)
		&0.810$\pm$0.004$~~$
		&0.810$\pm$0.004$~~$
		&\textbf{0.812$\pm$0.006}$~~$\\\hline		
		\multirow{4}{*}{Rec}
		&Rkl~($\downarrow$)
		&0.124$\pm$0.002$\bullet$
		&0.124$\pm$0.002$\bullet$
		&\textbf{0.118$\pm$0.002}$~~$
		&\multirow{4}{*}{Ref}
		&Rkl~($\downarrow$)
		&0.060$\pm$0.002$\bullet$
		&0.061$\pm$0.003$\bullet$
		&\textbf{0.054$\pm$0.004}$~~$\\
		&Auc~($\uparrow$)
		&0.871$\pm$0.003$~~$
		&0.870$\pm$0.003$~~$
		&\textbf{0.872$\pm$0.004}$~~$
		&&Auc~($\uparrow$)
		&0.940$\pm$0.003$\bullet$
		&0.939$\pm$0.004$\bullet$
		&\textbf{0.946$\pm$0.004}$~~$\\
		&Cvg~($\downarrow$)
		&3.704$\pm$0.033$~~$
		&3.700$\pm$0.037$~~$
		&\textbf{3.700$\pm$0.042}$~~$
		&&Cvg~($\downarrow$)
		&2.552$\pm$0.043$\bullet$
		&2.559$\pm$0.057$\bullet$
		&\textbf{2.325$\pm$0.060}$~~$\\
		&Ap~($\uparrow$)
		&0.670$\pm$0.004$~~$
		&0.670$\pm$0.004$~~$
		&\textbf{0.672$\pm$0.005}$~~$
		&&Ap~($\uparrow$)
		&0.739$\pm$0.004$\bullet$
		&0.739$\pm$0.004$\bullet$
		&\textbf{0.783$\pm$0.005}$~~$\\\hline
		\multirow{4}{*}{Sci}
		&Rkl~($\downarrow$)
		&0.107$\pm$0.004$~~$
		&0.108$\pm$0.004$~~$
		&\textbf{0.107$\pm$0.004}$~~$
		&\multirow{4}{*}{Soc}
		&Rkl~($\downarrow$)
		&0.063$\pm$0.002$\bullet$
		&0.063$\pm$0.002$\bullet$
		&\textbf{0.060$\pm$0.002}$~~$\\			
		&Auc~($\uparrow$)
		&0.893$\pm$0.004$~~$
		&0.892$\pm$0.004$~~$
		&\textbf{0.893$\pm$0.005}$~~$
		&&Auc~($\uparrow$)
		&0.930$\pm$0.002$\bullet$
		&0.930$\pm$0.002$\bullet$
		&\textbf{0.934$\pm$0.002}$~~$\\		
		&Cvg~($\downarrow$)
		&5.937$\pm$0.041$\bullet$
		&5.941$\pm$0.049$\bullet$
		&\textbf{5.845$\pm$0.054}$~~$
		&&Cvg~($\downarrow$)
		&3.558$\pm$0.033$~~$
		&3.559$\pm$0.038$~~$
		&\textbf{3.552$\pm$0.049}$~~$\\			
		&Ap~($\uparrow$)
		&0.608$\pm$0.003$~~$
		&0.608$\pm$0.003$~~$
		&\textbf{0.610$\pm$0.003}$~~$
		&&Ap~($\uparrow$)
		&0.797$\pm$0.002$~~$
		&0.797$\pm$0.003$~~$
		&\textbf{0.798$\pm$0.003}$~~$\\\hline		
		\multirow{4}{*}{Soci}
		&Rkl~($\downarrow$)
		&0.126$\pm$0.003$\bullet$
		&0.126$\pm$0.005$\bullet$
		&\textbf{0.113$\pm$0.005}$~~$
		&\multirow{4}{*}{Enr}
		&Rkl~($\downarrow$)
		&0.117$\pm$0.002$\bullet$
		&0.119$\pm$0.003$\bullet$
		&\textbf{0.105$\pm$0.005}$~~$\\
		&Auc~($\uparrow$)
		&0.874$\pm$0.003$\bullet$
		&0.874$\pm$0.004$\bullet$
		&\textbf{0.887$\pm$0.005}$~~$
		&&Auc~($\uparrow$)
		&0.883$\pm$0.004$\bullet$
		&0.881$\pm$0.004$\bullet$
		&\textbf{0.895$\pm$0.004}$~~$\\				
		&Cvg~($\downarrow$)
		&5.554$\pm$0.047$\bullet$
		&5.553$\pm$0.053$\bullet$
		&\textbf{5.208$\pm$0.059}$~~$
		&&Cvg~($\downarrow$)
		&19.440$\pm$0.833$\bullet$
		&19.372$\pm$0.915$\bullet$
		&\textbf{17.511$\pm$1.231}$~~$\\	
		&Ap~($\uparrow$)
		&0.670$\pm$0.004$\bullet$
		&0.670$\pm$0.005$\bullet$
		&\textbf{0.711$\pm$0.005}$~~$
		&&Ap~($\uparrow$)
		&0.685$\pm$0.005$\bullet$
		&0.673$\pm$0.005$\bullet$
		&\textbf{0.706$\pm$0.007}$~~$\\\hline
		\multirow{4}{*}{Cor}
		&Rkl~($\downarrow$)
		&0.163$\pm$0.002$\bullet$
		&0.163$\pm$0.002$\bullet$
		&\textbf{0.160$\pm$0.002}$~~$
		&\multirow{4}{*}{Ima}
		&Rkl~($\downarrow$)
		&0.197$\pm$0.003$\bullet$
		&0.199$\pm$0.004$\bullet$
		&\textbf{0.190$\pm$0.004}$~~$\\
		&Auc~($\uparrow$)
		&0.837$\pm$0.002$\bullet$
		&0.837$\pm$0.002$\bullet$
		&\textbf{0.840$\pm$0.002}$~~$
		&&Auc~($\uparrow$)
		&0.803$\pm$0.003$\bullet$
		&0.801$\pm$0.003$\bullet$
		&\textbf{0.810$\pm$0.003}$~~$\\
		&Cvg~($\downarrow$)
		&130.84$\pm$1.01$\bullet$
		&131.13$\pm$1.21$\bullet$
		&\textbf{128.40$\pm$1.30}$~~$
		&&Cvg~($\downarrow$)
		&1.064$\pm$0.015$\bullet$
		&1.066$\pm$0.021$\bullet$
		&\textbf{1.027$\pm$0.027}$~~$\\		
		&Ap~($\uparrow$)
		&0.212$\pm$0.003$~~$
		&0.212$\pm$0.003$~~$
		&\textbf{0.214$\pm$0.005}$~~$
		&&Ap~($\uparrow$)
		&0.764$\pm$0.003$\bullet$
		&0.763$\pm$0.004$\bullet$
		&\textbf{0.771$\pm$0.005}$~~$\\\hline	
	\end{tabular}
\end{table*}

\subsection{Learning with Missing Labels}

In this experiment, 
%we validate the performance with missing labels.
we randomly sample $\rho \%$ 
%(where $\rho=10, 30, 50, 70, 90$) 
of the elements in the label matrix
%positive and negative 
as observed, and the rest as missing. 
Note that \texttt{BR} and \texttt{MLLOC} can only handle datasets with full labels. 
%missing-label 
Hence, we first use \texttt{MAXIDE} \cite{xu2013speedup}, a matrix
completion algorithm for transductive multi-label learning, to fill in the missing
labels before they can be applied. 
We use \texttt{MBR} for \texttt{MAXIDE+BR}, and \texttt{MMLLOC}  for
\texttt{MAXIDE+MLLOC}.

Tables~\ref{tbl:mis_r} 
and \ref{tbl:mis_p} 
show the results on the training and test data, respectively.\footnote{To fit the
	tables on one page, we do not report the standard deviation.
	\texttt{MBR}, which performs worst, is also not shown.}

As can be seen,	
performance increases
with more observed entries in general.
%which makes sence that more observation provides more infomation. 
Overall, \texttt{GLOCAL}
performs best 
at different $\rho$'s, as it simultaneously considers both global and local
label correlations with label manifold regularization.
%missing label recovery via latent label decomposition, and  the instance mapping to the latent labels.
In contrast, \texttt{MBR} and \texttt{MMLLOC} 
%two stage approaches which 
handle label recovery and learning 
separately. Moreover,  \texttt{MMLLOC} takes only local label correlation, and  \texttt{MBR}
does not consider label correlations. As a result, they perform much worse 
than \texttt{GLOCAL}. Though \texttt{LEML} and \texttt{ML-LRC} perform
learning with missing label recovery together, they consider only global  correlation, and are thus often worse than  \texttt{GLOCAL}.

\subsection{Convergence}

\begin{figure}[h]
	\scriptsize
	\centering
	\subfigure[Arts.]{\includegraphics[width=0.24\textwidth,height=0.2\textwidth]{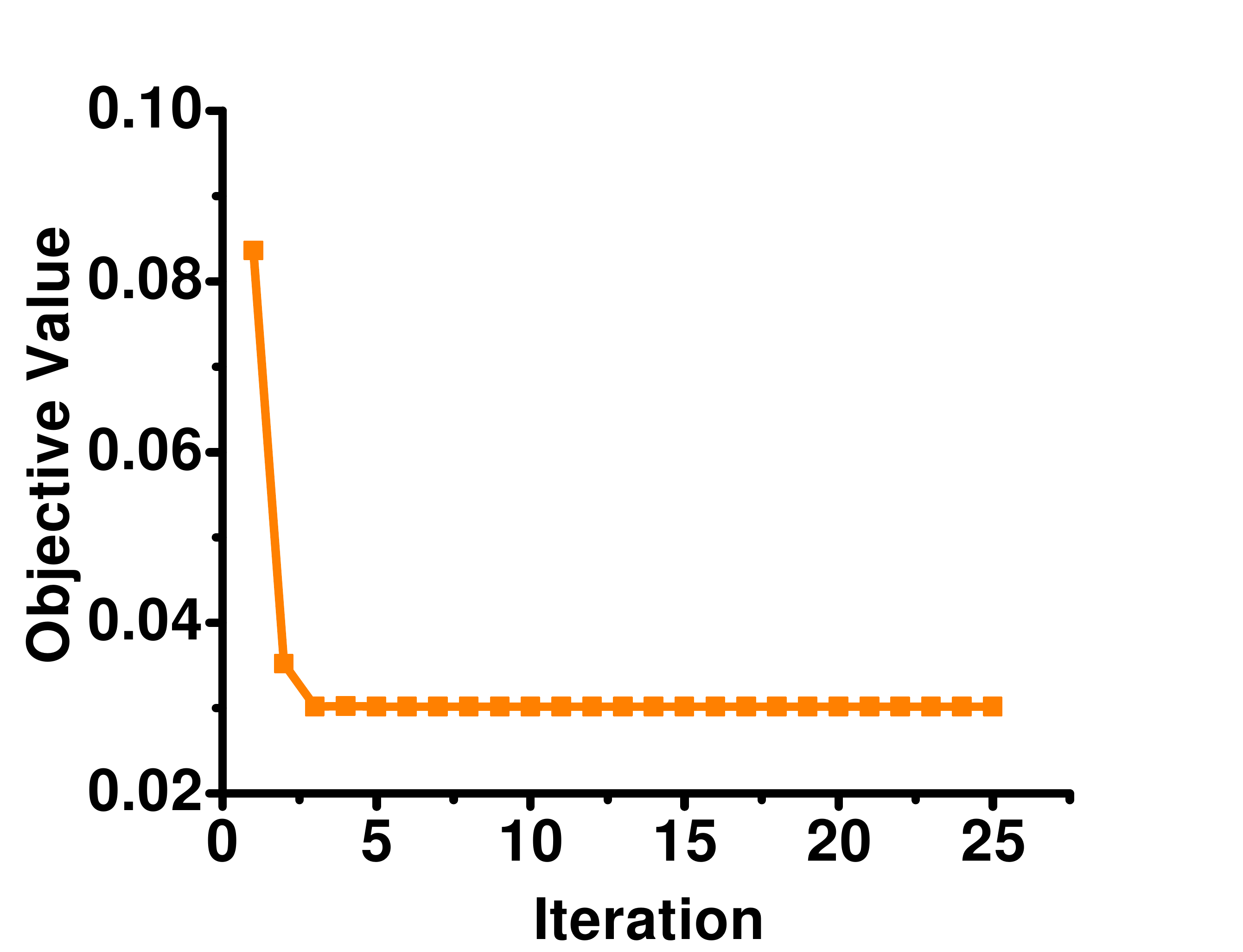}}
	\subfigure[Business.]{\includegraphics[width=0.24\textwidth,height=0.2\textwidth]{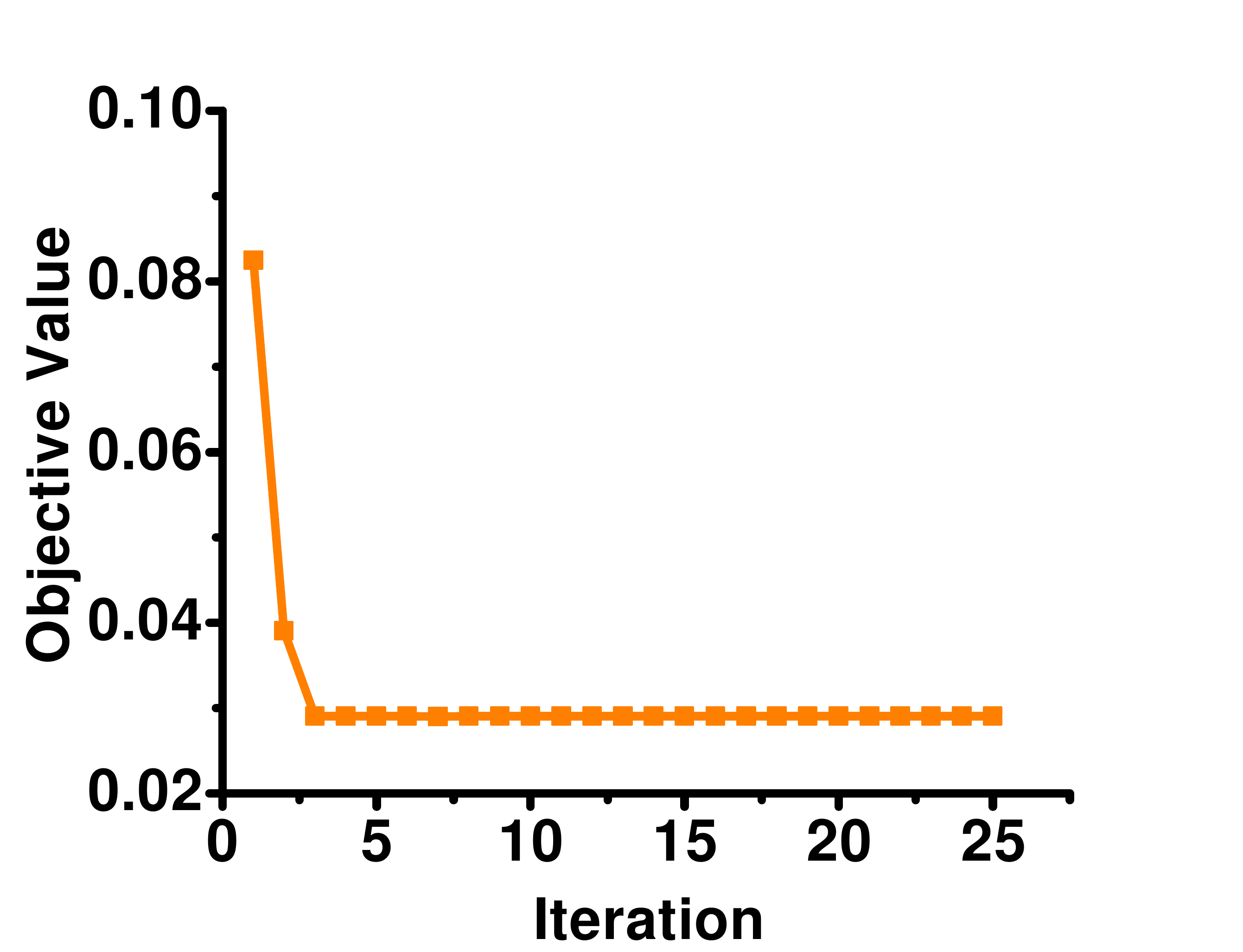}}
	\subfigure[Enron.]{\includegraphics[width=0.24\textwidth,height=0.2\textwidth]{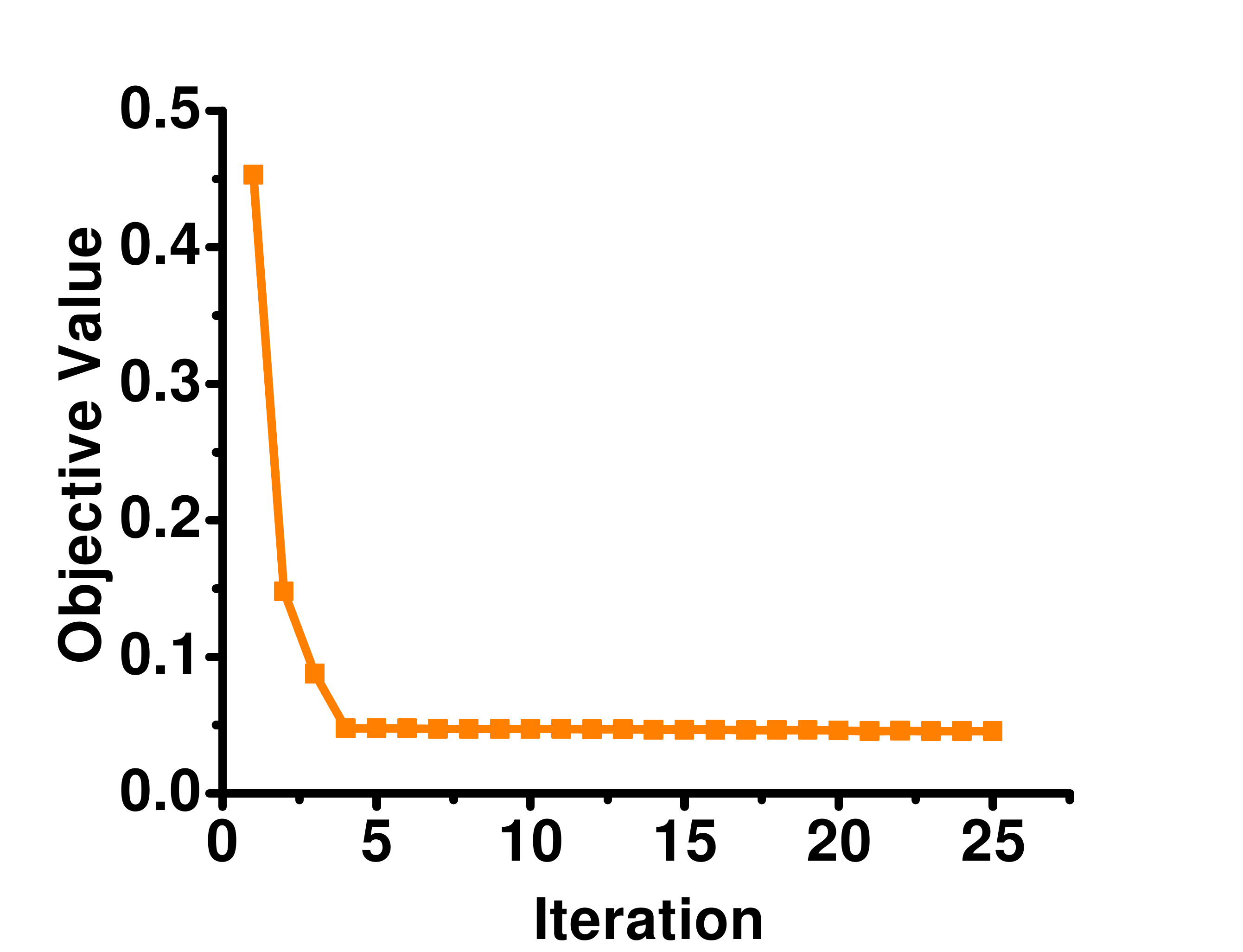}}
	\subfigure[Image.]{\includegraphics[width=0.24\textwidth,height=0.2\textwidth]{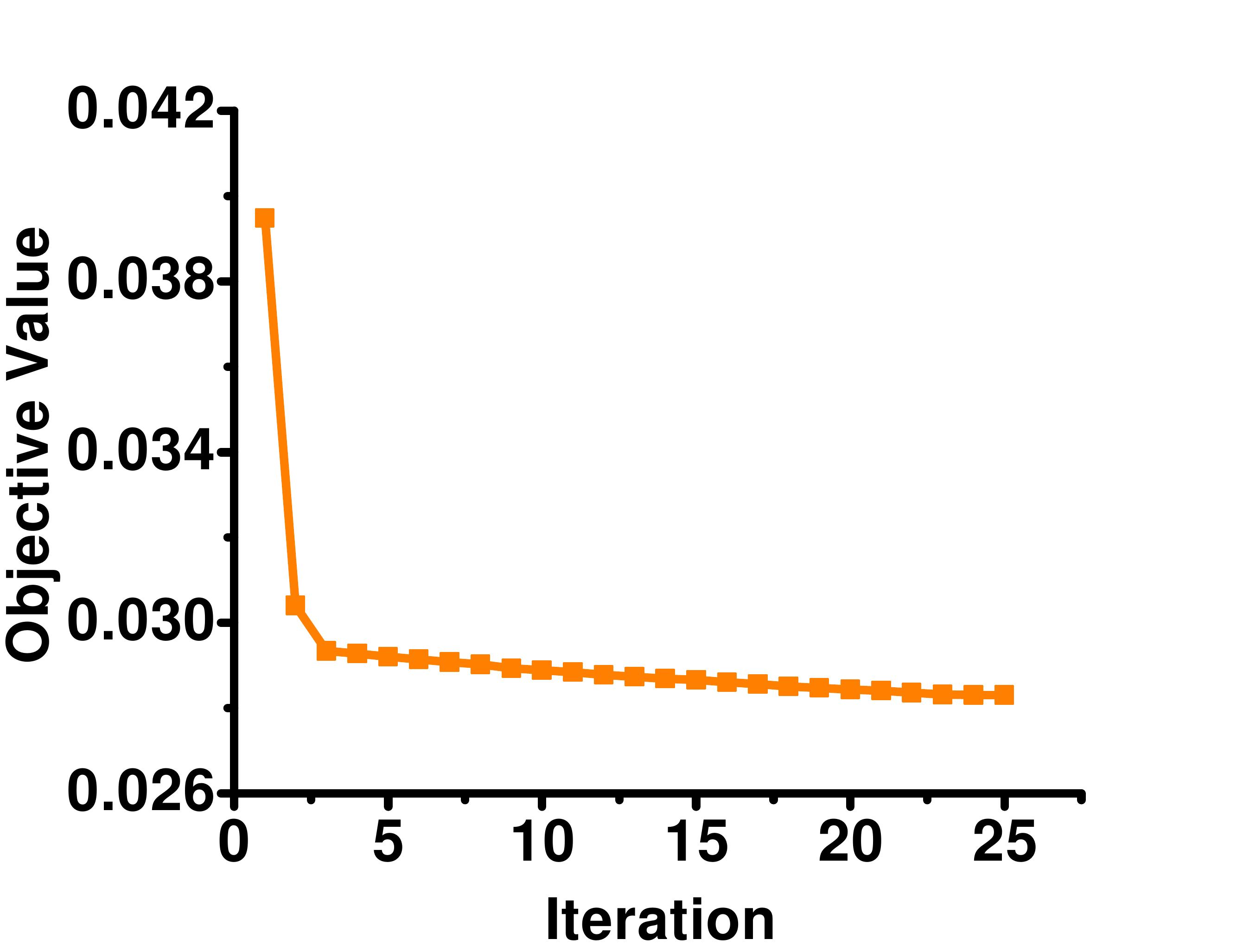}}
	\vspace{-.1in}
	\caption{Convergence of \texttt{GLOCAL} on the Arts, Business, Enron and Image datasets.\label{fig:convergence}}
\end{figure}

In this section, we empirically study the convergence of \texttt{GLOCAL}.
Figure \ref{fig:convergence} 
%shows the relative difference
%of the objective value in adjacent iterations 
shows the objective value w.r.t. the number of iterations for the full-label case. Because of the
lack of space, results are only shown
on the Arts, Business, Enron and Image datasets.
%As can be seen, the relative difference of the objective value drops very quickly from 1 to nearly 0 in the first 10 iterations.
As can be seen, the objective converges quickly in a few iterations.
%\texttt{GLOCAL} converges at last. 
A similar phenomenon can be observed on the other datasets. 

\begin{table*}[!]
	\tiny
	\centering
	\caption{Recovery results for missing label data on  ranking loss(Rkl), average auc(Auc), coverage(Cvg) and average precision(Ap).. $\uparrow$ ($\downarrow$) denotes the larger (smaller) the better.  $\bullet$ indicates that the \texttt{GLOCAL} is significantly better (paired t-tests at 95\% significance level). }
	%based on top-3 precision (Tp3), ranking loss (Rkl), coverage (Cvg) and average precision (Ap). 
	\label{tbl:mis_r} 
	\begin{tabular}{l|c|c|cccc|l|c|c|cccc}\hline
		&Measure&$\rho$&\texttt{MAXIDE}&\texttt{LEML}&\texttt{ML-LRC}&\texttt{GLOCAL}
		&&Measure&$\rho$&\texttt{MAXIDE}&\texttt{LEML}&\texttt{ML-LRC}&\texttt{GLOCAL}\\\hline
		\multirow{8}{*}{Art}
		&\multirow{2}{*}{Rkl~($\downarrow$)}
		&30
		&0.131$\bullet$
		&0.133$\bullet$
		&0.137$\bullet$
		&\textbf{0.103}$~~$
		&\multirow{8}{*}{Bus}
		&\multirow{2}{*}{Rkl~($\downarrow$)}
		&30
		&0.044$\bullet$
		&0.046$\bullet$
		&0.046$\bullet$
		&\textbf{0.029}$~~$
		\\
		&&70
		&0.083$\bullet$
		&0.090$\bullet$
		&0.083$\bullet$
		&\textbf{0.074}$~~$
		&&&70
		&0.026$\bullet$
		&0.027$\bullet$
		&0.024$\bullet$
		&\textbf{0.021}$~~$
		\\\cline{2-7}\cline{9-14}	
		
		&\multirow{2}{*}{Auc~($\uparrow$)}
		&30
		&0.871$\bullet$
		&0.848$\bullet$
		&0.879$\bullet$
		&\textbf{0.897}$~~$
		&&\multirow{2}{*}{Auc~($\uparrow$)}
		&30
		&0.956$\bullet$
		&0.954$\bullet$
		&0.954$\bullet$
		&\textbf{0.971}$~~$
		\\
		&&70
		&0.918$\bullet$
		&0.912$\bullet$
		&0.910$\bullet$
		&\textbf{0.928}$~~$
		&&&70
		&0.974$\bullet$
		&0.973$\bullet$
		&0.974$\bullet$
		&\textbf{0.979}$~~$
		\\\cline{2-7}\cline{9-14}	
		&\multirow{2}{*}{Cvg~($\downarrow$)}
		&30
		&5.195$\bullet$
		&5.231$\bullet$
		&5.161$\bullet$
		&\textbf{4.189}$~~$
		&&\multirow{2}{*}{Cvg~($\downarrow$)}
		&30
		&2.550$\bullet$
		&2.622$\bullet$
		&2.622$\bullet$
		&\textbf{1.830}$~~$
		\\
		&&70
		&3.616$\bullet$
		&3.733$\bullet$
		&3.778$\bullet$
		&\textbf{3.234}$~~$
		&&&70
		&1.742$\bullet$
		&1.783$\bullet$
		&1.746$\bullet$
		&\textbf{1.477}$~~$
		\\\cline{2-7}\cline{9-14}	
		&\multirow{2}{*}{Ap~($\uparrow$)}
		&30
		&0.645$\bullet$
		&0.634$\bullet$
		&0.640$\bullet$
		&\textbf{0.652}$~~$
		&&\multirow{2}{*}{Ap~($\uparrow$)}
		&30
		&0.876$\bullet$
		&0.878$\bullet$
		&0.876$\bullet$
		&\textbf{0.893}$~~$
		\\
		&&70
		&0.720$~~$
		&0.720$~~$
		&0.709$\bullet$
		&0.720$~~$
		&&&70
		&0.905$\bullet$
		&0.901$\bullet$
		&0.903$\bullet$
		&\textbf{0.908}$~~$
		\\\hline
		\multirow{8}{*}{Com}
		&\multirow{2}{*}{Rkl~($\downarrow$)}
		&30
		&0.101$\bullet$
		&0.098$\bullet$
		&0.097$\bullet$
		&\textbf{0.073}$~~$
		&\multirow{8}{*}{Edu}
		&\multirow{2}{*}{Rkl~($\downarrow$)}
		&30
		&0.097$\bullet$
		&0.093$\bullet$
		&0.089$\bullet$
		&\textbf{0.069}$~~$
		\\
		&&70
		&0.059$\bullet$
		&0.063$\bullet$
		&0.061$\bullet$
		&\textbf{0.052}$~~$
		&&&70
		&0.061$\bullet$
		&0.061$\bullet$
		&0.061$\bullet$
		&\textbf{0.058}$~~$
		\\\cline{2-7}\cline{9-14}	
		&\multirow{2}{*}{Auc~($\uparrow$)}
		&30
		&0.905$\bullet$
		&0.908$\bullet$
		&0.909$\bullet$
		&\textbf{0.933}$~~$
		&&\multirow{2}{*}{Auc~($\uparrow$)}
		&30
		&0.902$\bullet$
		&0.907$\bullet$
		&0.911$\bullet$
		&\textbf{0.932}$~~$
		\\
		&&70
		&0.947$\bullet$
		&0.943$\bullet$
		&0.945$\bullet$
		&\textbf{0.955}$~~$
		&&&70
		&0.938$\bullet$
		&0.938$\bullet$
		&0.940$~~$
		&\textbf{0.942}$~~$
		\\\cline{2-7}\cline{9-14}	
		&\multirow{2}{*}{Cvg~($\downarrow$)}
		&30
		&4.627$\bullet$
		&4.586$\bullet$
		&4.565$\bullet$
		&\textbf{3.511}$~~$
		&&\multirow{2}{*}{Cvg~($\downarrow$)}
		&30
		&4.672$\bullet$
		&4.372$\bullet$
		&3.914$\bullet$
		&\textbf{3.171}$~~$
		\\
		&&70
		&2.912$\bullet$
		&3.100$\bullet$
		&3.095$\bullet$
		&\textbf{2.586}$~~$
		&&&70
		&3.113$\bullet$
		&3.106$\bullet$
		&3.000$~~$
		&\textbf{2.815}$~~$
		\\\cline{2-7}\cline{9-14}	
		&\multirow{2}{*}{Ap~($\uparrow$)}
		&30
		&0.709$\bullet$
		&0.700$\bullet$
		&0.705$\bullet$
		&\textbf{0.726}$~~$
		&&\multirow{2}{*}{Ap~($\uparrow$)}
		&30
		&0.653$~~$
		&0.648$\bullet$
		&0.653$~~$
		&\textbf{0.655}$~~$
		\\
		&&70
		&0.787$~~$
		&0.787$~~$
		&0.787$~~$
		&0.787$~~$
		&&&70
		&\textbf{0.711}$~~$
		&0.702$\bullet$
		&0.710$~~$
		&\textbf{0.711}$~~$
		\\\hline
		\multirow{8}{*}{Ent}
		&\multirow{2}{*}{Rkl~($\downarrow$)}
		&30
		&0.104$\bullet$
		&0.103$\bullet$
		&0.106$\bullet$
		
		&\textbf{0.085}$~~$
		&\multirow{8}{*}{Hea}
		&\multirow{2}{*}{Rkl~($\downarrow$)}
		&30
		&0.060$\bullet$
		&0.057$\bullet$
		&0.054$\bullet$
		&\textbf{0.041}$~~$
		\\
		&&70
		&0.063$~~$
		&0.063$~~$
		&0.063$~~$
		&\textbf{0.062}$~~$
		&&&70
		&0.037$\bullet$
		&0.036$\bullet$
		&0.032$~~$
		&\textbf{0.030}$~~$
		\\\cline{2-7}\cline{9-14}	
		&\multirow{2}{*}{Auc~($\uparrow$)}
		&30
		&0.898$\bullet$
		&0.899$\bullet$
		&0.899$\bullet$
		&\textbf{0.916}$~~$
		&&\multirow{2}{*}{Auc~($\uparrow$)}
		&30
		&0.941$\bullet$
		&0.943$\bullet$
		&0.947$\bullet$
		&\textbf{0.960}$~~$
		\\
		&&70
		&\textbf{0.940}$~~$
		&0.938$~~$
		&\textbf{0.940}$~~$
		&\textbf{0.940}$~~$
		&&&70
		&0.964$\bullet$
		&0.964$\bullet$
		&0.968$~~$
		&\textbf{0.971}$~~$
		\\\cline{2-7}\cline{9-14}	
		&\multirow{2}{*}{Cvg~($\downarrow$)}
		&30
		&3.058$\bullet$
		&2.994$\bullet$
		&3.022$\bullet$
		&\textbf{2.512}$~~$
		&&\multirow{2}{*}{Cvg~($\downarrow$)}
		&30
		&3.577$\bullet$
		&3.462$\bullet$
		&3.465$\bullet$
		&\textbf{2.567}$~~$
		\\
		&&70
		&1.987$~~$
		&2.051$~~$
		&2.080$~~$
		&\textbf{1.957}$~~$
		&&&70
		&2.524$\bullet$
		&2.465$\bullet$
		&2.450$\bullet$
		&\textbf{2.152}$~~$
		\\\cline{2-7}\cline{9-14}	
		&\multirow{2}{*}{Ap~($\uparrow$)}
		&30
		&0.704$~~$
		&0.698$\bullet$
		&0.698$\bullet$
		&\textbf{0.704}$~~$
		&&\multirow{2}{*}{Ap~($\uparrow$)}
		&30
		&0.796$\bullet$
		&0.794$\bullet$
		&0.798$~~$
		&\textbf{0.801}$~~$
		\\
		&&70
		&0.763$\bullet$
		&0.765$~~$
		&0.765$~~$
		&\textbf{0.768}$~~$
		&&&70
		&\textbf{0.848}$~~$
		&0.842$\bullet$
		&\textbf{0.848}$~~$
		&\textbf{0.848}$~~$
		\\\hline
		\multirow{8}{*}{Rec}
		&\multirow{2}{*}{Rkl~($\downarrow$)}
		&30
		&0.130$\bullet$
		&0.133$\bullet$
		&0.135$\bullet$
		&\textbf{0.110}$~~$
		&\multirow{8}{*}{Ref}
		&\multirow{2}{*}{Rkl~($\downarrow$)}
		&30
		&0.083$\bullet$
		&0.083$\bullet$
		&0.083$\bullet$
		&\textbf{0.063}$~~$
		\\
		&&70
		&0.078$\bullet$
		&0.080$\bullet$
		&0.080$\bullet$
		&\textbf{0.068}$~~$
		&&&70
		&0.048$~~$
		&0.049$~~$
		&0.049$~~$
		&\textbf{0.048}$~~$
		\\\cline{2-7}\cline{9-14}	
		&\multirow{2}{*}{Auc~($\uparrow$)}
		&30
		&0.873$\bullet$
		&0.870$\bullet$
		&0.869$\bullet$
		&\textbf{0.895}$~~$
		&&\multirow{2}{*}{Auc~($\uparrow$)}
		&30
		&0.919$\bullet$
		&0.919$\bullet$
		&0.918$\bullet$
		&\textbf{0.939}$~~$
		\\
		&&70
		&0.925$\bullet$
		&0.923$\bullet$
		&0.920$\bullet$
		&\textbf{0.934}$~~$
		&&&70
		&\textbf{0.955}$~~$
		&0.953$~~$
		&0.953$~~$
		&\textbf{0.955}$~~$
		\\\cline{2-7}\cline{9-14}	
		&\multirow{2}{*}{Cvg~($\downarrow$)}
		&30
		&3.899$\bullet$
		&3.919$\bullet$
		&4.048$\bullet$
		&\textbf{3.291}$~~$
		&&\multirow{2}{*}{Cvg~($\downarrow$)}
		&30
		&3.436$\bullet$
		&3.392$\bullet$
		&3.372$\bullet$
		&\textbf{2.520}$~~$
		\\
		&&70
		&2.560$\bullet$
		&2.607$\bullet$
		&2.620$\bullet$
		&\textbf{2.262}$~~$
		&&&70
		&2.039$\bullet$
		&2.103$\bullet$
		&2.195$\bullet$
		&\textbf{1.972}$~~$
		\\\cline{2-7}\cline{9-14}	
		&\multirow{2}{*}{Ap~($\uparrow$)}
		&30
		&0.680$\bullet$
		&0.663$\bullet$
		&0.660$\bullet$
		&\textbf{0.681}$~~$
		&&\multirow{2}{*}{Ap~($\uparrow$)}
		&30
		&\textbf{0.681}$~~$
		&0.664$\bullet$
		&0.674$~~$
		&0.679$~~$
		\\
		&&70
		&0.767$\bullet$
		&0.763$\bullet$
		&0.760$\bullet$
		&\textbf{0.770}$~~$
		&&&70
		&0.745$~~$
		&\textbf{0.746}$~~$
		&\textbf{0.746}$~~$
		&\textbf{0.746}$~~$\\\hline
		\multirow{8}{*}{Sci}
		&\multirow{2}{*}{Rkl~($\downarrow$)}
		&30
		&0.110$\bullet$
		&0.111$\bullet$
		&0.110$\bullet$
		&\textbf{0.086}$~~$
		&\multirow{8}{*}{Soc}
		&\multirow{2}{*}{Rkl~($\downarrow$)}
		&30
		&0.069$\bullet$
		&0.069$\bullet$
		&0.063$\bullet$
		&\textbf{0.042}$~~$
		\\
		&&70
		&0.063$~~$
		&0.071$\bullet$
		&0.070$\bullet$
		&\textbf{0.063}$~~$
		&&&70
		&0.041$\bullet$
		&0.040$\bullet$
		&0.040$\bullet$
		&\textbf{0.026}$~~$
		\\\cline{2-7}\cline{9-14}	
		&\multirow{2}{*}{Auc~($\uparrow$)}
		&30
		&0.889$\bullet$
		&0.889$\bullet$
		&0.889$\bullet$
		&\textbf{0.913}$~~$
		&&\multirow{2}{*}{Auc~($\uparrow$)}
		&30
		&0.930$\bullet$
		&0.930$\bullet$
		&0.936$\bullet$
		&\textbf{0.957}$~~$
		\\
		&&70
		&\textbf{0.935}$~~$
		&0.928$\bullet$
		&0.923$\bullet$
		&\textbf{0.935}$~~$
		&&&70
		&0.964$\bullet$
		&0.959$\bullet$
		&0.966$\bullet$
		&\textbf{0.973}$~~$
		\\\cline{2-7}\cline{9-14}	
		&\multirow{2}{*}{Cvg~($\downarrow$)}
		&30
		&6.193$\bullet$
		&6.141$\bullet$
		&6.271$\bullet$
		&\textbf{4.845}$~~$
		&&\multirow{2}{*}{Cvg~($\downarrow$)}
		&30
		&3.865$\bullet$
		&3.920$\bullet$
		&3.304$\bullet$
		&\textbf{2.443}$~~$
		\\
		&&70
		&3.771$~~$
		&3.914$\bullet$
		&3.878$\bullet$
		&\textbf{3.751}$~~$
		&&&70
		&2.103$\bullet$
		&2.386$\bullet$
		&2.373$\bullet$
		&\textbf{1.663}$~~$
		\\\cline{2-7}\cline{9-14}	
		&\multirow{2}{*}{Ap~($\uparrow$)}
		&30
		&0.615$~~$
		&0.613$~~$
		&0.614$~~$
		&\textbf{0.615}$~~$
		&&\multirow{2}{*}{Ap~($\uparrow$)}
		&30
		&0.780$\bullet$
		&0.780$\bullet$
		&0.784$\bullet$
		&\textbf{0.802}$~~$
		\\
		&&70
		&0.689$\bullet$
		&0.647$\bullet$
		&0.650$\bullet$
		&\textbf{0.691}$~~$
		&&&70
		&0.854$\bullet$
		&\textbf{0.865}$~~$
		&\textbf{0.865}$~~$
		&\textbf{0.865}$~~$\\\hline
		\multirow{8}{*}{Soci}
		&\multirow{2}{*}{Rkl~($\downarrow$)}
		&30
		&0.129$\bullet$
		&0.128$\bullet$
		&0.123$\bullet$
		&\textbf{0.102}$~~$
		&\multirow{8}{*}{Enr}
		&\multirow{2}{*}{Rkl~($\downarrow$)}
		&30
		&0.091$\bullet$
		&0.115$\bullet$
		&0.085$\bullet$
		&\textbf{0.075}$~~$
		\\
		&&70
		&0.074$~~$
		&0.081$\bullet$
		&\textbf{0.073}$~~$
		&\textbf{0.073}$~~$
		&&&70
		&0.042$~~$
		&0.060$\bullet$
		&\textbf{0.040}$~~$
		&\textbf{0.040}$~~$
		\\\cline{2-7}\cline{9-14}	
		&\multirow{2}{*}{Auc~($\uparrow$)}
		&30
		&0.871$\bullet$
		&0.872$\bullet$
		&0.877$\bullet$
		&\textbf{0.898}$~~$
		&&\multirow{2}{*}{Auc~($\uparrow$)}
		&30
		&0.910$\bullet$
		&0.887$\bullet$
		&0.918$\bullet$
		&\textbf{0.926}$~~$
		\\
		&&70
		&0.926$~~$
		&0.919$\bullet$
		&0.928$~~$
		&\textbf{0.929}$~~$
		&&&70
		&0.960$~~$
		&0.942$\bullet$
		&\textbf{0.962}$~~$
		&\textbf{0.962}$~~$
		\\\cline{2-7}\cline{9-14}	
		&\multirow{2}{*}{Cvg~($\downarrow$)}
		&30
		&5.557$\bullet$
		&5.459$\bullet$
		&5.167$\bullet$
		&\textbf{4.496}$~~$
		&&\multirow{2}{*}{Cvg~($\downarrow$)}
		&30
		&14.24$\bullet$
		&16.65$\bullet$
		&13.45$\bullet$
		&\textbf{12.05}$~~$
		\\
		&&70
		&3.641$\bullet$
		&3.824$\bullet$
		&3.608$\bullet$
		&\textbf{3.442}$~~$
		&&&70
		&7.961$\bullet$
		&10.33$\bullet$
		&\textbf{7.480}$~~$
		&7.510$~~$
		\\\cline{2-7}\cline{9-14}	
		&\multirow{2}{*}{Ap~($\uparrow$)}
		&30
		&0.646$~~$
		&0.629$\bullet$
		&0.650$~~$
		&\textbf{0.652}$~~$
		&&\multirow{2}{*}{Ap~($\uparrow$)}
		&30
		&\textbf{0.739}$~~$
		&0.711$\bullet$
		&\textbf{0.739}$~~$
		&\textbf{0.739}$~~$
		\\
		&&70
		&\textbf{0.719}$~~$
		&0.717$~~$
		&\textbf{0.719}$~~$
		&\textbf{0.719}$~~$
		&&&70
		&0.854$~~$
		&0.842$\bullet$
		&\textbf{0.855}$~~$
		&\textbf{0.855}$~~$
		\\\hline
		\multirow{8}{*}{Cor}
		&\multirow{2}{*}{Rkl~($\downarrow$)}
		&30
		&0.226$\bullet$
		&0.214$\bullet$
		&0.206$\bullet$
		&\textbf{0.185}$~~$
		&\multirow{8}{*}{Ima}
		&\multirow{2}{*}{Rkl~($\downarrow$)}
		&30
		&0.302$\bullet$
		&0.184$\bullet$
		&0.175$~~$
		&\textbf{0.173}$~~$
		\\
		&&70
		&0.138$\bullet$
		&0.131$\bullet$
		&\textbf{0.123}$~~$
		&0.125$~~$
		&&&70
		&0.251$\bullet$
		&\textbf{0.148}$~~$
		&\textbf{0.148}$~~$
		&\textbf{0.148}$~~$
		\\\cline{2-7}\cline{9-14}	
		&\multirow{2}{*}{Auc~($\uparrow$)}
		&30
		&0.773$\bullet$
		&0.786$\bullet$
		&0.794$\bullet$
		&\textbf{0.814}$~~$
		&&\multirow{2}{*}{Auc~($\uparrow$)}
		&30
		&0.820$\bullet$
		&\textbf{0.828}$~~$
		&0.826$~~$
		&\textbf{0.828}$~~$
		\\
		&&70
		&0.874$~~$
		&0.874$~~$
		&0.874$~~$
		&0.874$~~$
		&&&70
		&0.834$\bullet$
		&\textbf{0.857}$~~$
		&0.855$~~$
		&0.855$~~$
		\\\cline{2-7}\cline{9-14}	
		&\multirow{2}{*}{Cvg~($\downarrow$)}
		&30
		&204.90$\bullet$
		&182.76$\bullet$
		&178.60$\bullet$
		&\textbf{153.82}$~~$
		&&\multirow{2}{*}{Cvg~($\downarrow$)}
		&30
		&1.493$\bullet$
		&1.104$\bullet$
		&0.967$~~$
		&\textbf{0.950}$~~$
		\\
		&&70
		&103.63$~~$
		&102.42$~~$
		&\textbf{102.30}$~~$
		&\textbf{102.30}$~~$
		&&&70
		&0.790$\bullet$
		&\textbf{0.760}$~~$
		&0.770$~~$
		&\textbf{0.760}$~~$
		\\\cline{2-7}\cline{9-14}	
		&\multirow{2}{*}{Ap~($\uparrow$)}
		&30
		&\textbf{0.275}$~~$
		&0.259$\bullet$
		&\textbf{0.275}$~~$
		&\textbf{0.275}$~~$
		&&\multirow{2}{*}{Ap~($\uparrow$)}
		&30
		&0.739$\bullet$
		&0.776$\bullet$
		&0.775$\bullet$
		&\textbf{0.785}$~~$
		\\
		&&70
		&0.279$~~$
		&0.279$~~$
		&0.279$~~$
		&0.279$~~$
		&&&70
		&0.768$\bullet$
		&\textbf{0.841}$~~$
		&0.834$~~$
		&\textbf{0.841}$~~$
		\\\hline
	\end{tabular}
\end{table*}

\begin{table*}[!]
	\tiny
	\centering
	\caption{Prediction results for missing label data on  ranking loss(Rkl), average auc(Auc), coverage(Cvg) and average precision(Ap).. $\uparrow$ ($\downarrow$) denotes the larger (smaller) the better.  $\bullet$ indicates that the \texttt{GLOCAL} is significantly better (paired t-tests at 95\% significance level). }
	%based on top-3 precision (Tp3), ranking loss (Rkl), coverage (Cvg) and average precision (Ap). 
	\label{tbl:mis_p} 
	\begin{tabular}{l|c|c|cccc|l|c|c|cccc}\hline
		&Measure&$\rho$&\texttt{MMLLOC}&\texttt{LEML}&\texttt{ML-LRC}&\texttt{GLOCAL}
		&&Measure&$\rho$&\texttt{MMLLOC}&\texttt{LEML}&\texttt{ML-LRC}&\texttt{GLOCAL}\\\hline
		\multirow{8}{*}{Art}
		&\multirow{2}{*}{Rkl~($\downarrow$)}
		&30
		&0.225$\bullet$
		&0.204$\bullet$
		&0.184$\bullet$
		&\textbf{0.144}$~~$
		&\multirow{8}{*}{Bus}
		&\multirow{2}{*}{Rkl~($\downarrow$)}
		&30
		&0.083$\bullet$
		&0.063$\bullet$
		&0.061$\bullet$
		&\textbf{0.054}$~~$
		\\
		&&70
		&0.193$\bullet$
		&0.181$\bullet$
		&0.159$\bullet$
		&\textbf{0.139}$~~$
		&&&70
		&0.064$\bullet$
		&0.058$\bullet$
		&0.046$~~$
		&\textbf{0.046}$~~$
		\\\cline{2-7}\cline{9-14}	
		&\multirow{2}{*}{Auc~($\uparrow$)}
		&30
		&0.781$\bullet$
		&0.801$\bullet$
		&0.828$~~$
		&\textbf{0.831}$~~$
		&&\multirow{2}{*}{Auc~($\uparrow$)}
		&30
		&0.917$\bullet$
		&0.928$\bullet$
		&0.937$~~$
		&\textbf{0.937}$~~$
		\\
		&&70
		&0.819$\bullet$
		&0.825$\bullet$
		&0.838$~~$
		&\textbf{0.840}$~~$
		&&&70
		&0.935$\bullet$
		&0.942$\bullet$
		&0.950$~~$
		&\textbf{0.952}$~~$
		\\\cline{2-7}\cline{9-14}	
		&\multirow{2}{*}{Cvg~($\downarrow$)}
		&30
		&9.033$\bullet$
		&7.369$\bullet$
		&6.281$\bullet$
		&\textbf{5.867}$~~$
		&&\multirow{2}{*}{Cvg~($\downarrow$)}
		&30
		&4.643$\bullet$
		&3.954$\bullet$
		&3.279$\bullet$
		&\textbf{2.863}$~~$
		\\
		&&70
		&7.262$\bullet$
		&6.431$\bullet$
		&5.432$~~$
		&\textbf{5.352}$~~$
		&&&70
		&3.670$\bullet$
		&3.303$\bullet$
		&2.580$~~$
		&\textbf{2.579}$~~$
		\\\cline{2-7}\cline{9-14}	
		&\multirow{2}{*}{Ap~($\uparrow$)}
		&30
		&0.529$\bullet$
		&0.503$\bullet$
		&0.517$\bullet$
		&\textbf{0.572}$~~$
		&&\multirow{2}{*}{Ap~($\uparrow$)}
		&30
		&0.843$\bullet$
		&0.866$\bullet$
		&0.858$\bullet$
		&\textbf{0.879}$~~$
		\\
		&&70
		&0.583$\bullet$
		&0.589$\bullet$
		&0.588$\bullet$
		&\textbf{0.607}$~~$
		&&&70
		&0.861$\bullet$
		&0.870$\bullet$
		&0.870$\bullet$
		&\textbf{0.881}$~~$\\\hline
		\multirow{8}{*}{Com}
		&\multirow{2}{*}{Rkl~($\downarrow$)}
		&30
		&0.201$\bullet$
		&0.179$\bullet$
		&\textbf{0.152}
		&0.154$~~$
		&\multirow{8}{*}{Edu}
		&\multirow{2}{*}{Rkl~($\downarrow$)}
		&30
		&0.187$\bullet$
		&0.176$\bullet$
		&0.144$\bullet$
		&\textbf{0.137}$~~$
		\\
		&&70
		&0.150$\bullet$
		&0.141$\bullet$
		&0.115$~~$
		&\textbf{0.113}$~~$
		&&&70
		&0.165$\bullet$
		&0.151$\bullet$
		&0.113$~~$
		&\textbf{0.111}$~~$
		\\\cline{2-7}\cline{9-14}	
		&\multirow{2}{*}{Auc~($\uparrow$)}
		&30
		&0.849$\bullet$
		&0.880$~~$
		&0.873$\bullet$
		&\textbf{0.883}$~~$
		&&\multirow{2}{*}{Auc~($\uparrow$)}
		&30
		&0.815$\bullet$
		&0.817$\bullet$
		&0.845$~~$
		&\textbf{0.846}$~~$
		\\
		&&70
		&0.868$\bullet$
		&0.894$~~$
		&0.895$~~$
		&\textbf{0.896}$~~$
		&&&70
		&0.844$\bullet$
		&0.842$\bullet$
		&0.860$~~$
		&\textbf{0.860}$~~$
		\\\cline{2-7}\cline{9-14}	
		&\multirow{2}{*}{Cvg~($\downarrow$)}
		&30
		&8.808$\bullet$
		&7.392$\bullet$
		&6.052$\bullet$
		&\textbf{5.798}$~~$
		&&\multirow{2}{*}{Cvg~($\downarrow$)}
		&30
		&11.089$\bullet$
		&9.672$\bullet$
		&6.350$~~$
		&\textbf{6.338}$~~$
		\\
		&&70
		&6.871$\bullet$
		&6.306$\bullet$
		&5.000$~~$
		&\textbf{4.976}$~~$
		&&&70
		&8.096$\bullet$
		&7.595$\bullet$
		&5.075$~~$
		&\textbf{5.070}$~~$
		\\\cline{2-7}\cline{9-14}	
		&\multirow{2}{*}{Ap~($\uparrow$)}
		&30
		&0.631$\bullet$
		&0.646$\bullet$
		&0.636$\bullet$
		&\textbf{0.669}$~~$
		&&\multirow{2}{*}{Ap~($\uparrow$)}
		&30
		&0.538$\bullet$
		&0.537$\bullet$
		&0.543$\bullet$
		&\textbf{0.592}$~~$
		\\
		&&70
		&0.674$\bullet$
		&0.665$\bullet$
		&0.667$\bullet$
		&\textbf{0.691}$~~$
		&&&70
		&0.586$\bullet$
		&0.591$\bullet$
		&0.600$\bullet$
		&\textbf{0.622}$~~$
		\\\hline
		\multirow{8}{*}{Ent}
		&\multirow{2}{*}{Rkl~($\downarrow$)}
		&30
		&0.229$\bullet$
		&0.175$\bullet$
		&0.152$\bullet$
		&\textbf{0.122}$~~$
		&\multirow{8}{*}{Hea}
		&\multirow{2}{*}{Rkl~($\downarrow$)}
		&30
		&0.137$\bullet$
		&0.095$\bullet$
		&0.085$~~$
		&\textbf{0.085}$~~$
		\\
		&&70
		&0.164$\bullet$
		&0.159$\bullet$
		
		&0.129$\bullet$
		&\textbf{0.109}$~~$
		&&&70
		&0.109$\bullet$
		&0.074$\bullet$
		&0.071$\bullet$
		&\textbf{0.065}$~~$
		\\\cline{2-7}\cline{9-14}	
		&\multirow{2}{*}{Auc~($\uparrow$)}
		&30
		&0.832$\bullet$
		&0.826$\bullet$
		&0.849$\bullet$
		&\textbf{0.859}$~~$
		&&\multirow{2}{*}{Auc~($\uparrow$)}
		&30
		&0.894$\bullet$
		&0.896$\bullet$
		&\textbf{0.907}$~~$
		&0.906$~~$
		\\
		&&70
		&0.842$~~$
		&0.850$\bullet$
		&0.870$~~$
		&\textbf{0.871}$~~$
		&&&70
		&0.901$\bullet$
		&\textbf{0.920}$~~$
		&\textbf{0.920}$~~$
		&\textbf{0.920}$~~$
		\\\cline{2-7}\cline{9-14}	
		&\multirow{2}{*}{Cvg~($\downarrow$)}
		&30
		&6.029$\bullet$
		&5.755$\bullet$
		&4.170$~~$
		&\textbf{4.153}$~~$
		&&\multirow{2}{*}{Cvg~($\downarrow$)}
		&30
		&7.104$\bullet$
		&6.248$\bullet$
		&4.924$~~$
		&\textbf{4.814}$~~$
		\\
		&&70
		&4.857$\bullet$
		&4.643$\bullet$
		&3.483$\bullet$
		&\textbf{3.117}$~~$
		&&&70
		&5.866$\bullet$
		&5.167$\bullet$
		&\textbf{3.960}$~~$
		&3.963$~~$
		\\\cline{2-7}\cline{9-14}	
		&\multirow{2}{*}{Ap~($\uparrow$)}
		&30
		&0.601$\bullet$
		&0.601$\bullet$
		&0.601$\bullet$
		&\textbf{0.645}$~~$
		&&\multirow{2}{*}{Ap~($\uparrow$)}
		&30
		&0.727$\bullet$
		&0.715$\bullet$
		&0.720$\bullet$
		&\textbf{0.752}$~~$
		\\
		&&70
		&0.635$\bullet$
		&0.645$\bullet$
		&0.643$\bullet$
		&\textbf{0.670}$~~$
		&&&70
		&0.762$\bullet$
		&0.770$\bullet$
		&0.766$\bullet$
		&\textbf{0.775}$~~$
		\\\hline
		\multirow{8}{*}{Rec}
		&\multirow{2}{*}{Rkl~($\downarrow$)}
		&30
		&0.266$\bullet$
		&0.245$\bullet$
		&0.202$\bullet$
		&\textbf{0.165}$~~$
		&\multirow{8}{*}{Ref}
		&\multirow{2}{*}{Rkl~($\downarrow$)}
		&30
		&0.199$\bullet$
		&0.187$\bullet$
		&0.137$\bullet$
		&\textbf{0.098}$~~$
		\\
		&&70
		&0.204$\bullet$
		&0.196$\bullet$
		&0.167$\bullet$
		&\textbf{0.156}$~~$
		&&&70
		&0.155$\bullet$
		&0.145$\bullet$
		&0.098$\bullet$
		&\textbf{0.086}$~~$
		\\\cline{2-7}\cline{9-14}	
		&\multirow{2}{*}{Auc~($\uparrow$)}
		&30
		&0.785$\bullet$
		&0.828$\bullet$
		&0.802$\bullet$
		&\textbf{0.839}$~~$
		&&\multirow{2}{*}{Auc~($\uparrow$)}
		&30
		&0.851$\bullet$
		&0.847$\bullet$
		&0.868$\bullet$
		&\textbf{0.886}$~~$
		\\
		&&70
		&0.800$\bullet$
		&0.837$\bullet$
		&0.836$\bullet$
		&\textbf{0.845}$~~$
		&&&70
		&0.861$\bullet$
		&0.869$\bullet$
		&0.895$~~$
		&\textbf{0.898}$~~$
		\\\cline{2-7}\cline{9-14}	
		&\multirow{2}{*}{Cvg~($\downarrow$)}
		&30
		&7.084$\bullet$
		&6.842$\bullet$
		&5.397$\bullet$
		&\textbf{4.545}$~~$
		&&\multirow{2}{*}{Cvg~($\downarrow$)}
		&30
		&7.549$\bullet$
		&6.463$\bullet$
		&5.052$\bullet$
		&\textbf{3.367}$~~$
		\\
		&&70
		&5.952$\bullet$
		&5.685$\bullet$
		&4.490$~~$
		&\textbf{4.430}$~~$
		&&&70
		&6.419$\bullet$
		&6.130$\bullet$
		&3.694$\bullet$
		&\textbf{3.348}$~~$
		\\\cline{2-7}\cline{9-14}	
		&\multirow{2}{*}{Ap~($\uparrow$)}
		&30
		&0.547$\bullet$
		&0.540$\bullet$
		&0.540$\bullet$
		&\textbf{0.573}$~~$
		&&\multirow{2}{*}{Ap~($\uparrow$)}
		&30
		&0.631$~~$
		&0.609$\bullet$
		&0.611$\bullet$
		&\textbf{0.638}$~~$
		\\
		&&70
		&0.597$\bullet$
		&0.567$\bullet$
		&0.600$\bullet$
		&\textbf{0.614}$~~$
		&&&70
		&\textbf{0.675}$~~$
		&0.653$\bullet$
		&0.653$\bullet$
		&0.672$~~$
		\\\hline
		\multirow{8}{*}{Sci}
		&\multirow{2}{*}{Rkl~($\downarrow$)}
		&30
		&0.257$\bullet$
		&0.203$\bullet$
		&0.169$\bullet$
		&\textbf{0.144}$~~$
		&\multirow{8}{*}{Soc}
		&\multirow{2}{*}{Rkl~($\downarrow$)}
		&30
		&0.149$\bullet$
		&0.089$\bullet$
		&0.095$\bullet$
		&\textbf{0.075}$~~$
		\\
		&&70
		&0.189$\bullet$
		&0.174$\bullet$
		&0.134$~~$
		&\textbf{0.129}$~~$
		&&&70
		&0.108$\bullet$
		&0.079$\bullet$
		&0.076$\bullet$
		&\textbf{0.073}$~~$
		\\\cline{2-7}\cline{9-14}	
		&\multirow{2}{*}{Auc~($\uparrow$)}
		&30
		&0.827$\bullet$
		&0.827$\bullet$
		&0.830$\bullet$
		&\textbf{0.837}$~~$
		&&\multirow{2}{*}{Auc~($\uparrow$)}
		&30
		&0.906$\bullet$
		&0.906$\bullet$
		&0.905$\bullet$
		&\textbf{0.913}$~~$
		\\
		&&70
		&0.840$\bullet$
		&0.849$~~$
		&\textbf{0.850}$~~$
		&\textbf{0.850}$~~$
		&&&70
		&0.910$\bullet$
		&0.900$\bullet$
		&\textbf{0.914}$~~$
		&\textbf{0.914}$~~$
		\\\cline{2-7}\cline{9-14}	
		&\multirow{2}{*}{Cvg~($\downarrow$)}
		&30
		&12.805$\bullet$
		&10.587$\bullet$
		&8.794$\bullet$
		&\textbf{6.809}$~~$
		&&\multirow{2}{*}{Cvg~($\downarrow$)}
		&30
		&7.652$\bullet$
		&7.567$\bullet$
		&6.308$~~$
		&\textbf{6.088}$~~$
		\\
		&&70
		&9.960$\bullet$
		&9.501$\bullet$
		&6.900$\bullet$
		&\textbf{6.416}$~~$
		&&&70
		&5.886$\bullet$
		&5.386$\bullet$
		&5.103$~~$
		&\textbf{4.929}$~~$
		\\\cline{2-7}\cline{9-14}	
		&\multirow{2}{*}{Ap~($\uparrow$)}
		&30
		&0.503$\bullet$
		&0.479$\bullet$
		&0.485$\bullet$
		&\textbf{0.531}$~~$
		&&\multirow{2}{*}{Ap~($\uparrow$)}
		&30
		&0.712$\bullet$
		&0.682$\bullet$
		&0.700$\bullet$
		&\textbf{0.738}$~~$
		\\
		&&70
		&0.569$~~$
		&0.551$\bullet$
		&0.570$\bullet$
		&\textbf{0.574}$~~$
		&&&70
		&0.748$\bullet$
		&0.719$\bullet$
		&0.728$\bullet$
		&\textbf{0.761}$~~$
		\\\hline
		\multirow{8}{*}{Soci}
		&\multirow{2}{*}{Rkl~($\downarrow$)}
		&30
		&0.252$\bullet$
		&0.202$\bullet$
		&0.175$\bullet$
		&\textbf{0.139}$~~$
		&\multirow{8}{*}{Enr}
		&\multirow{2}{*}{Rkl~($\downarrow$)}
		&30
		&0.179$\bullet$
		&0.172$\bullet$
		&0.173$\bullet$
		&\textbf{0.149}$~~$
		\\
		&&70
		&0.208$\bullet$
		&0.194$\bullet$
		&0.141$\bullet$
		&\textbf{0.136}$~~$
		&&&70
		&0.170$\bullet$
		&0.162$\bullet$
		&0.152$\bullet$
		&\textbf{0.129}$~~$
		\\\cline{2-7}\cline{9-14}	
		&\multirow{2}{*}{Auc~($\uparrow$)}
		&30
		&0.804$\bullet$
		&0.808$\bullet$
		&\textbf{0.826}$~~$
		&\textbf{0.826}$~~$
		&&\multirow{2}{*}{Auc~($\uparrow$)}
		&30
		&0.820$\bullet$
		&0.830$\bullet$
		&9,843$\bullet$
		&\textbf{0.853}$~~$
		\\
		&&70
		&0.816$\bullet$
		&0.816$\bullet$
		&\textbf{0.840}$~~$
		&\textbf{0.840}$~~$
		&&&70
		&0.829$\bullet$
		&0.839$\bullet$
		&0.849$\bullet$
		&\textbf{0.872}$~~$
		\\\cline{2-7}\cline{9-14}	
		&\multirow{2}{*}{Cvg~($\downarrow$)}
		&30
		&9.550$\bullet$
		&8.637$\bullet$
		&6.944$\bullet$
		&\textbf{5.816}$~~$
		&&\multirow{2}{*}{Cvg~($\downarrow$)}
		&30
		&22.72$\bullet$
		&21.41$\bullet$
		&20.42$\bullet$
		&\textbf{19.01}$~~$
		\\
		&&70
		&8.227$\bullet$
		&7.638$\bullet$
		&\textbf{5.750}$~~$
		&\textbf{5.750}$~~$
		&&&70
		&21.90$\bullet$
		&19.53$\bullet$
		&18.17$\bullet$
		&\textbf{17.16}$~~$
		\\\cline{2-7}\cline{9-14}	
		&\multirow{2}{*}{Ap~($\uparrow$)}
		&30
		&0.569$\bullet$
		&0.563$\bullet$
		&0.565$\bullet$
		&\textbf{0.601}$~~$
		&&\multirow{2}{*}{Ap~($\uparrow$)}
		&30
		&0.580$\bullet$
		&0.582$\bullet$
		&0.580$\bullet$
		&\textbf{0.589}$~~$
		\\
		&&70
		&0.606$\bullet$
		&0.589$\bullet$
		&0.590$\bullet$
		&\textbf{0.625}$~~$
		&&&70
		&0.585$\bullet$
		&0.601$\bullet$
		&0.607$\bullet$
		&\textbf{0.635}$~~$
		\\\hline
		\multirow{8}{*}{Cor}
		&\multirow{2}{*}{Rkl~($\downarrow$)}
		&30
		&0.332$\bullet$
		&0.308$\bullet$
		&0.331$\bullet$
		&\textbf{0.285}$~~$
		&\multirow{8}{*}{Ima}
		&\multirow{2}{*}{Rkl~($\downarrow$)}
		&30
		&0.224$\bullet$
		&0.204$\bullet$
		&0.220$\bullet$
		&\textbf{0.200}$~~$
		\\
		&&70
		&0.248$\bullet$
		&0.250$\bullet$
		&0.199$~~$
		&\textbf{0.194}$~~$
		&&&70
		&0.195$\bullet$
		&0.188$~~$
		&0.197$\bullet$
		&\textbf{0.187}$~~$
		\\\cline{2-7}\cline{9-14}	
		&\multirow{2}{*}{Auc~($\uparrow$)}
		&30
		&0.673$\bullet$
		&0.693$\bullet$
		&0.670$\bullet$
		&\textbf{0.714}$~~$
		&&\multirow{2}{*}{Auc~($\uparrow$)}
		&30
		&0.796$\bullet$
		&0.795$\bullet$
		&0.800$~~$
		&\textbf{0.801}$~~$
		\\
		&&70
		&0.747$\bullet$
		&0.749$\bullet$
		&0.801$~~$
		&\textbf{0.805}$~~$
		&&&70
		&0.812$~~$
		&0.811$~~$
		&0.810$~~$
		&\textbf{0.813}$~~$
		\\\cline{2-7}\cline{9-14}	
		&\multirow{2}{*}{Cvg~($\downarrow$)}
		&30
		&275.41$\bullet$
		&233.83$\bullet$
		&240.17$\bullet$
		&\textbf{211.84}$~~$
		&&\multirow{2}{*}{Cvg~($\downarrow$)}
		&30
		&1.160$\bullet$
		&1.103$\bullet$
		&1.131$\bullet$
		&\textbf{1.070}$~~$
		\\
		&&70
		&212.84$\bullet$
		&190.83$\bullet$
		&160.59$\bullet$
		&\textbf{151.23}$~~$
		&&&70
		&1.066$\bullet$
		&1.030$~~$
		&1.040$\bullet$
		&\textbf{1.025}$~~$
		\\\cline{2-7}\cline{9-14}	
		&\multirow{2}{*}{Ap~($\uparrow$)}
		&30
		&0.158$\bullet$
		&0.166$\bullet$
		&0.165$\bullet$
		&\textbf{0.174}$~~$
		&&\multirow{2}{*}{Ap~($\uparrow$)}
		&30
		&0.745$\bullet$
		&0.752$\bullet$
		&0.744$\bullet$
		&\textbf{0.760}$~~$
		\\
		&&70
		&0.176$\bullet$
		&0.185$\bullet$
		&0.188$~~$
		&\textbf{0.192}$~~$
		&&&70
		&0.768$\bullet$
		&0.772$~~$
		&0.770$\bullet$
		&\textbf{0.777}$~~$
		
		\\\hline
	\end{tabular}
\end{table*}

\begin{table*}[!]
	\tiny
	\centering 
	\caption{CPU timing results for learning with missing labels ($\rho=70$).  F is the time to
		fill in the missing labels. C is the time for clustering, I is the time  for
		initialization, and R is the time of the main learning procedure. A is the total time
		(sum of F, I, C and R).  Note that some algorithms may not need
		F, C or I. \label{tbl:miss_time}}
	\begin{tabular}{l|c|cc|c|cccc|c|cc|c|cc|c|ccc}\hline
		& \multicolumn{3}{c|}{MBR} & \multicolumn{5}{c|}{MMLLOC} & \multicolumn{3}{c|}{LEML} & \multicolumn{3}{c|}{ML-LRC}  & \multicolumn{4}{c}{GLOCAL}\\\cline{2-19}
		&A &F &   R   &A &F & C  & I &   R  &A & I &   R  &A & I &   R &A & C  & I &   R  \\\hline 
		Arts&109
		&8
		&101
		&107
		&8
		&1
		&0
		&98
		&34
		&0
		&34
		&87
		&0
		&87
		&47
		&1
		&20
		&26
		\\
		Business&38
		&6
		&32
		&104
		&6
		&1
		&0
		&97
		&35
		&0
		&35
		&82
		&0
		&82
		&49
		&1
		&24
		&24\\
		Computers
		&78
		&11
		&67
		&121
		&11
		&1
		&0
		&109
		&46
		&0
		&46
		&94
		&0
		&94
		&53
		&1
		&31
		&21\\
		Education
		&60
		&8
		&52
		&115
		&8
		&1
		&0
		&106
		&45
		&0
		&45
		&64
		&0
		&64
		&45
		&1
		&29
		&15
		\\
		Entertainment
		&66
		&6
		&60
		&91
		&6
		&1
		&0
		&84
		&42
		&0
		&42
		&73
		&0
		&73
		&53
		&2
		&22
		&29
		\\
		Health
		&64
		&11
		&53
		&116
		&11
		&1
		&0
		&104
		&41
		&0
		&41
		&75
		&0
		&75
		&67
		&1
		&32
		&34\\
		Recreation
		&63
		&4
		&59
		&97
		&5
		&1
		&0
		&91
		&46
		&0
		&46
		&55
		&0
		&55
		&51
		&2
		&22
		&27\\
		Reference
		&75
		&14
		&61
		&131
		&15
		&9
		&0
		&107
		&38
		&0
		&38
		&91
		&0
		&91
		&78
		&8
		&32
		&38\\
		Science
		&101
		&15
		&86
		&133
		&15
		&1
		&0
		&117
		&53
		&0
		&53
		&103
		&0
		&103
		&77
		&2
		&32
		&43\\
		Social
		&163
		&36
		&127
		&149
		&33
		&8
		&0
		&108
		&37
		&0
		&37
		&147
		&0
		&147
		&90
		&7
		&35
		&48\\
		Society
		&83
		&8
		&75
		&106
		&8
		&1
		&0
		&97
		&32
		&0
		&32
		&117
		&0
		&117
		&44
		&2
		&18
		&24\\
		Enron&47
		&10
		&37
		&59
		&10
		&1
		&0
		&48
		&38
		&0
		&38
		&78
		&0
		&78
		&69
		&1
		&25
		&43\\
		Corel5k
		&458
		&272
		&186
		&1529
		&268
		&1
		&0
		&1260
		&307
		&0
		&307
		&709
		&0
		&709
		&413
		&1
		&78
		&344
		\\
		Image&5
		&1
		&4
		&25
		&2
		&1
		&0
		&22
		&28
		&0
		&28
		&14
		&0
		&14
		&15
		&1
		&5
		&9\\\hline				
	\end{tabular}
\end{table*}

%\subsection{Time Comparison}

Table~\ref{tbl:miss_time}
shows the  timing results on learning with missing labels (with
$\rho=70$).
%does not explicitly use label correlations and does not need clustering. Thus, 
\texttt{GLOCAL} and \texttt{LEML} 
train a classifier for all the labels 
jointly, and also can take advantage of the low-rank structure of either the model or label
matrix during training.
Thus, they are the fastest.  
However, \texttt{GLOCAL}  
has to be warm-started by Eqn.~\eqref{eqn:init},
%which will cost much time and for \texttt{GLOCAL}, there will be an additional 
and requires an additional clustering step to obtain local groups of the instances.  Hence, it is
slower than \texttt{LEML}.  However, as have been observed in previous sections,
\texttt{GLOCAL} outperforms
\texttt{LEML} 
in terms of label recovery.
\texttt{ML-LRC}  uses a low-rank label correlation matrix. However, it does not reduce the size
of the label matrix or model involved in each iteration, and so is slower than \texttt{GLOCAL}.
\texttt{MBR} and  \texttt{MMLLOC} require training a classifier for each label, and also an
additional step
to recover the missing labels. Thus, they are often the slowest, especially when the number of class labels is large.
Similar results can be observed with $\rho=30$, which are not reported here.

\subsection{Sensitivity to Parameters}

In this experiment, we study the influence of parameters, including the number of
clusters $g$, regularization parameters $\lambda_3$ and $\lambda_4$ (corresponding to
the manifold regularizer for global and local label correlations, respectively),
regularization parameter $\lambda_2$ for the Frobenius norm regularizer, and
dimensionality $k$ of the latent representation. 
We vary one  parameter, while keeping the others fixed
%Specifically, the range of parameters are listed in Table \ref{tbl:param}, and The best  parameter setting for Enron are $\{g=16, \lambda_3=1E-3, \lambda_4=1E-3, k=15, \lambda_2=0.5\}$; in the following sections, we vary one of the parameters, whereas keep the others the same value as those in 
at their best setting.
\subsubsection{Varying the Number of Clusters $g$} 

\begin{figure}[h]
	\centering
	\begin{minipage}[t]{1\linewidth} 
		\centering
		\includegraphics[width=0.24\textwidth,height=0.2\textwidth]{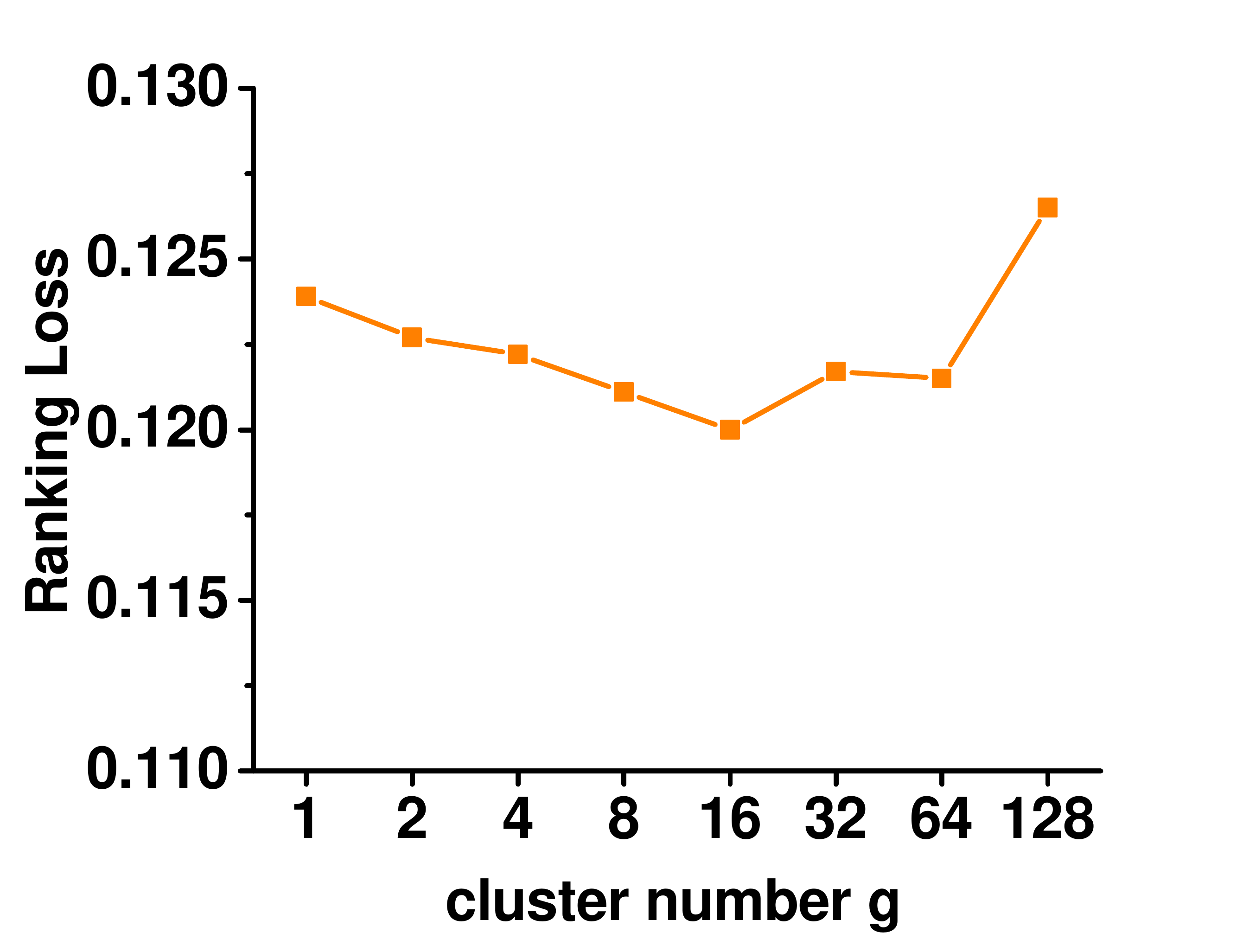}~~
		\includegraphics[width=0.24\textwidth,height=0.2\textwidth]{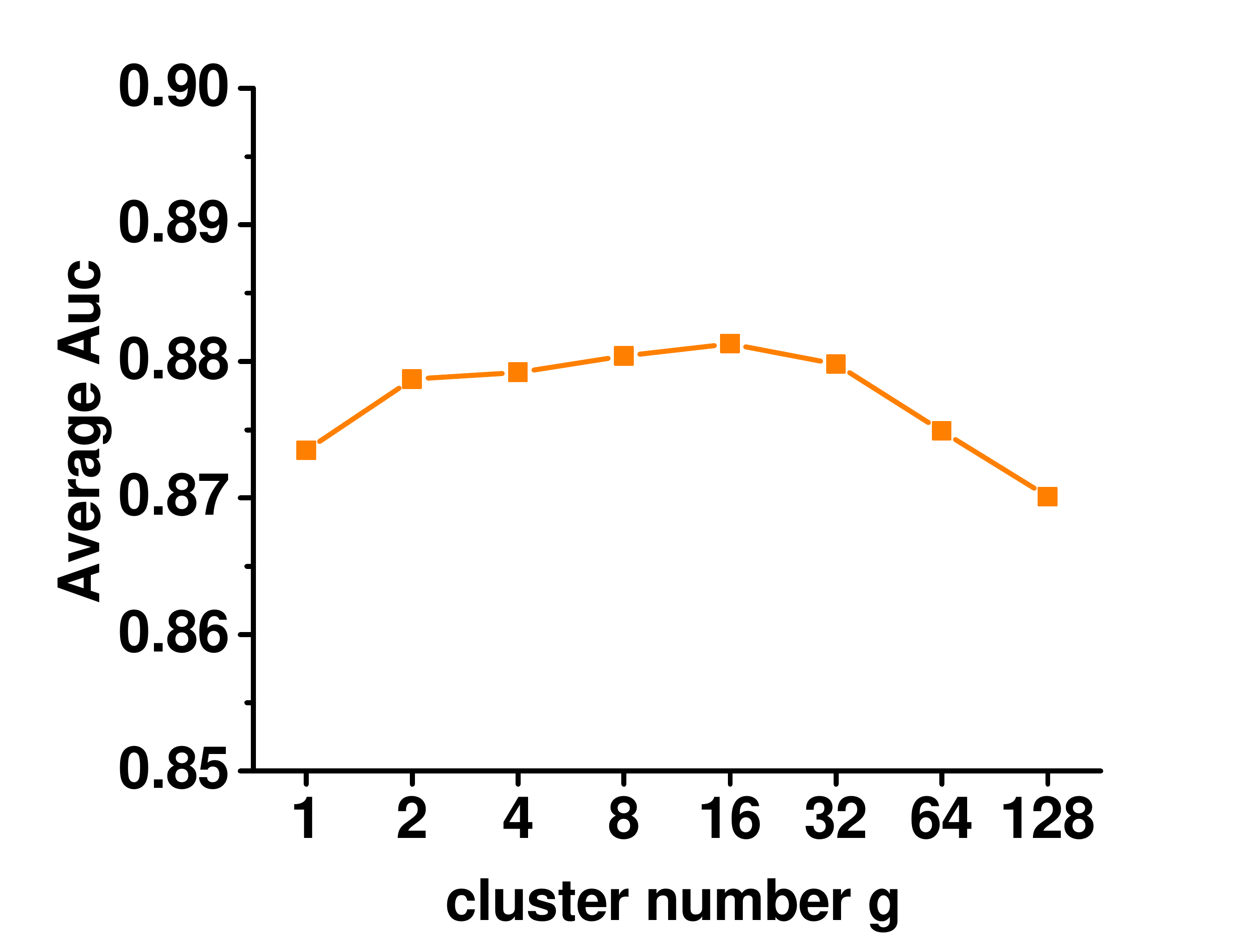}~~
		\includegraphics[width=0.24\textwidth,height=0.2\textwidth]{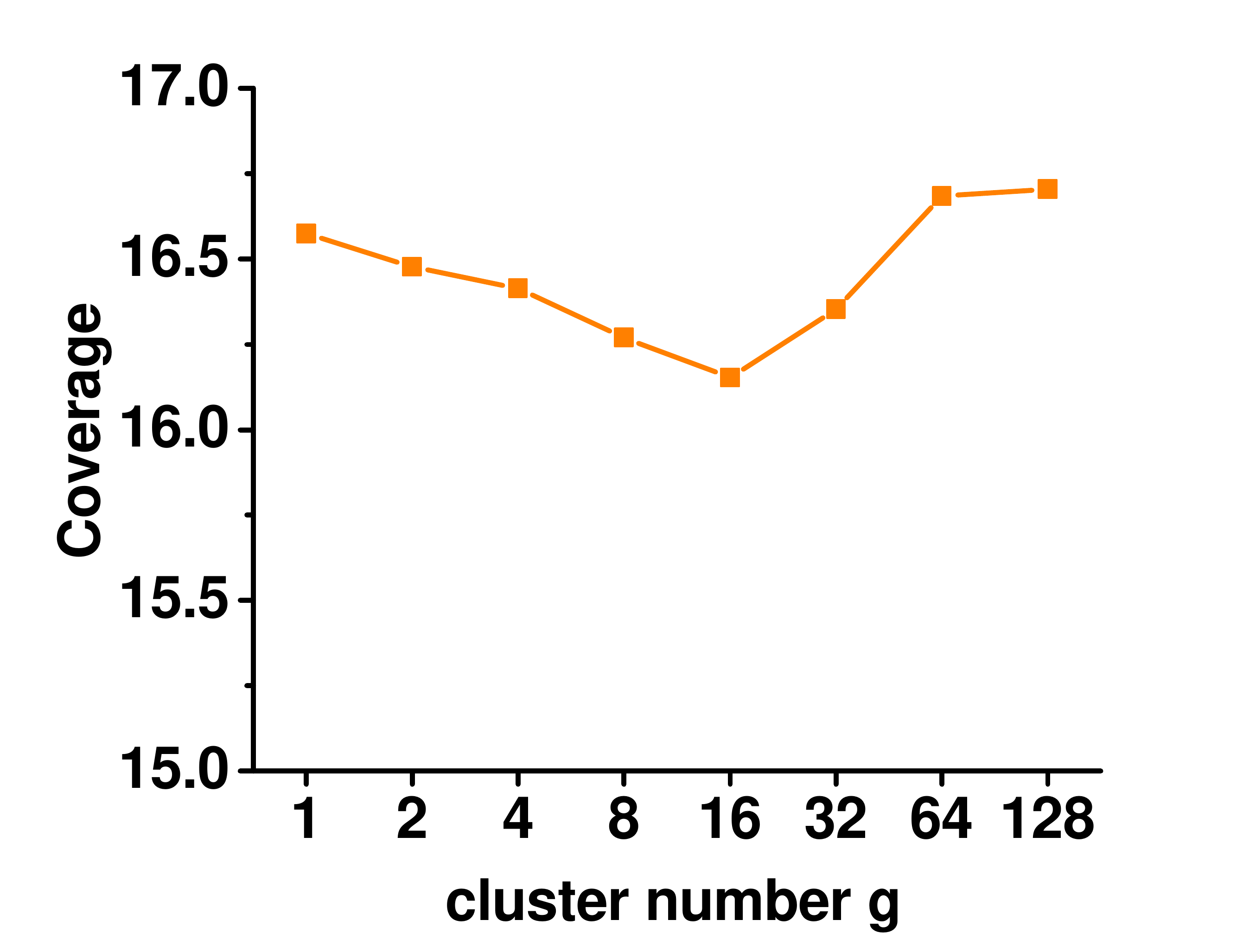}~~
		\includegraphics[width=0.24\textwidth,height=0.2\textwidth]{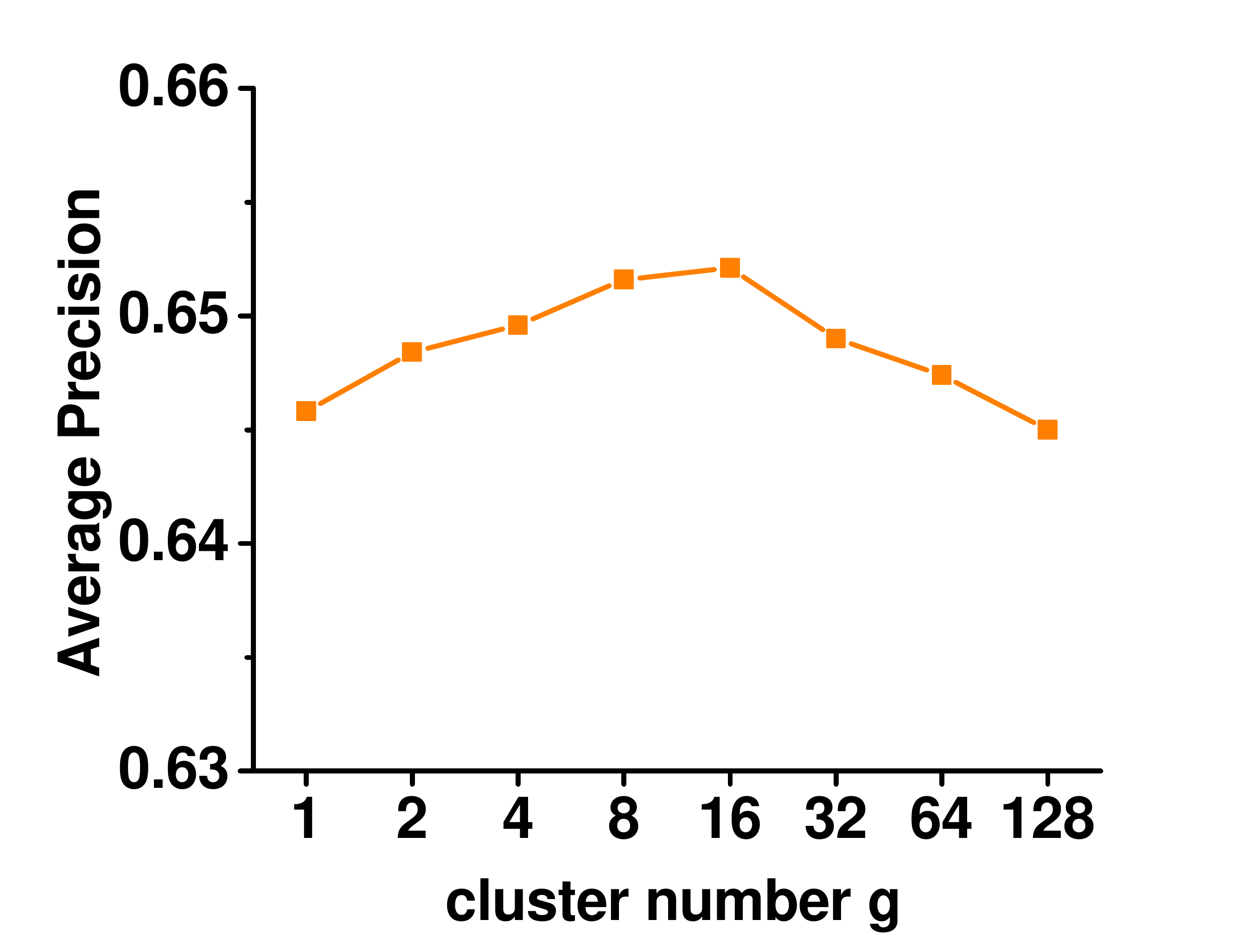}~~
	\end{minipage}
	\caption{Varying the number of clusters $g$ on the Enron dataset.
		\label{fig:clust} 
	}
\end{figure}

Figure \ref{fig:clust} shows the influence on the Enron dataset.
When there is only one cluster, no local label correlation is considered. 
With more clusters, performance improves as more local label correlations are taken into account. 
When
too many
clusters are used,
very few instances are  placed in each cluster, and
the local label correlations cannot be reliably estimated. Thus, the performance starts to deteriorate.

\subsubsection{Influence of Label Manifold Regularizers ($\lambda_3$ and $\lambda_4$)}

%Parameters $\lambda_3$ and $\lambda_4$ control the importance of the global and local label correlations, respectively. 

\begin{figure}[h]
	\centering
	\begin{minipage}[t]{1\linewidth} 
		\centering
		\includegraphics[width=0.24\textwidth,height=0.2\textwidth]{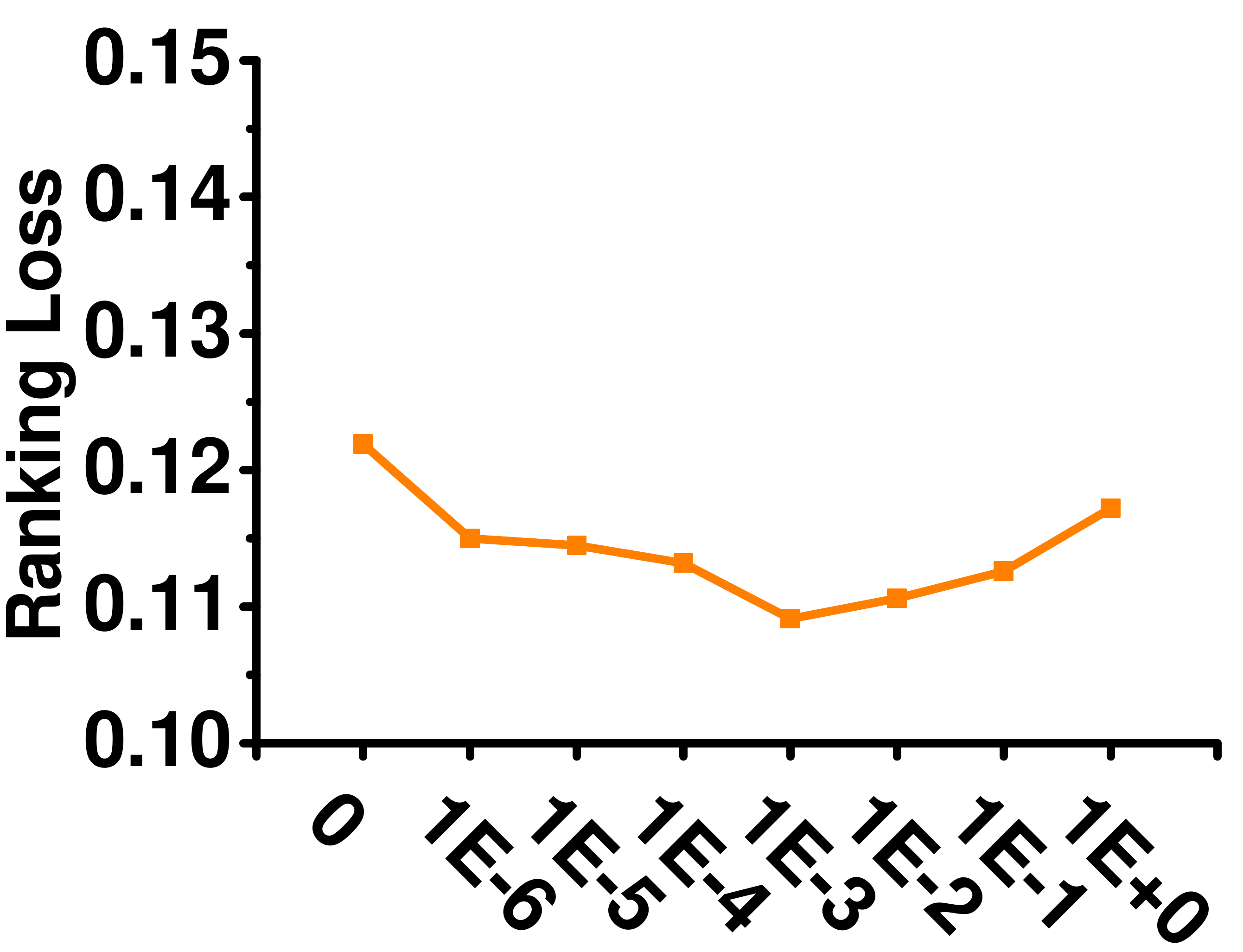}~
		\includegraphics[width=0.24\textwidth,height=0.2\textwidth]{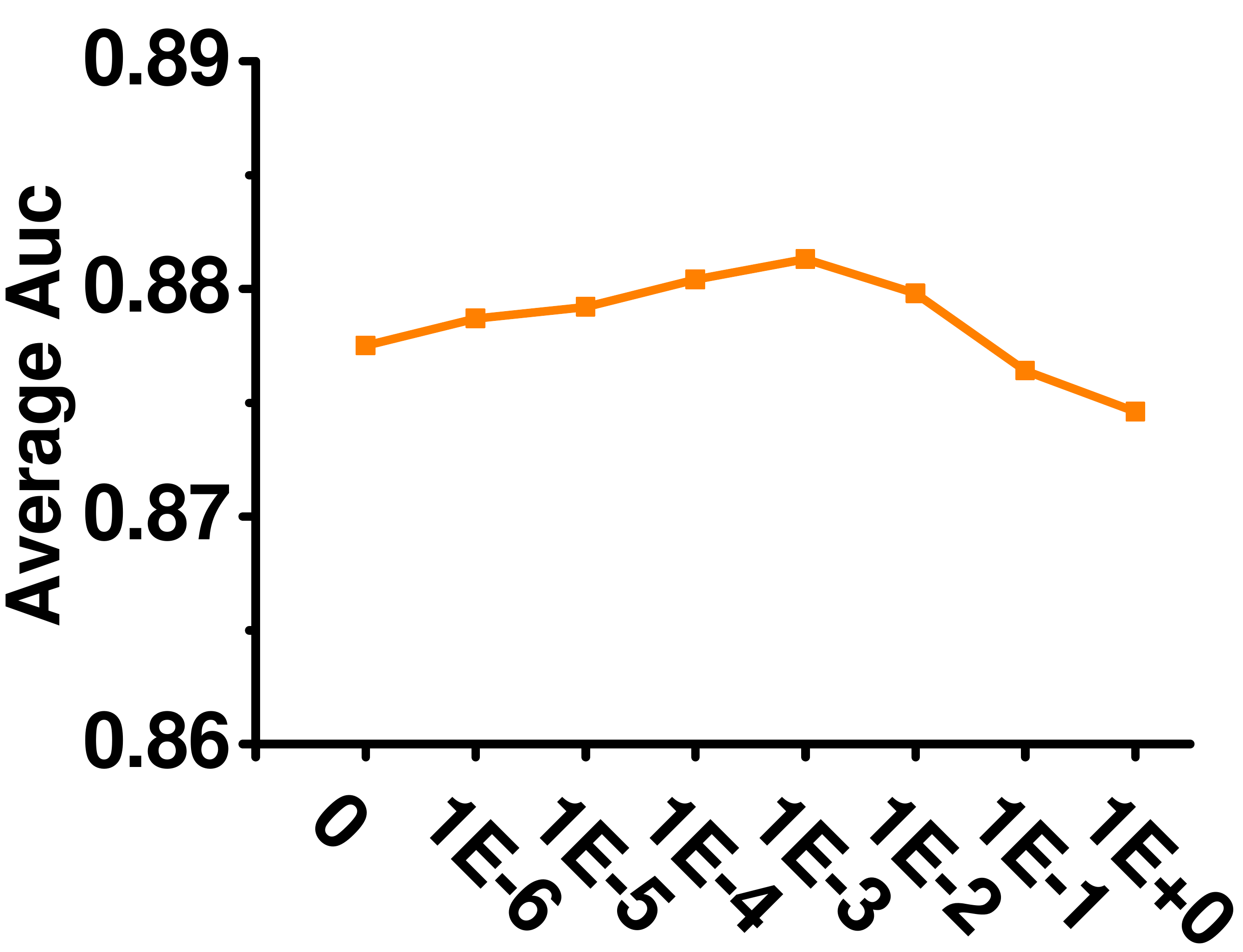}~
		\includegraphics[width=0.24\textwidth,height=0.2\textwidth]{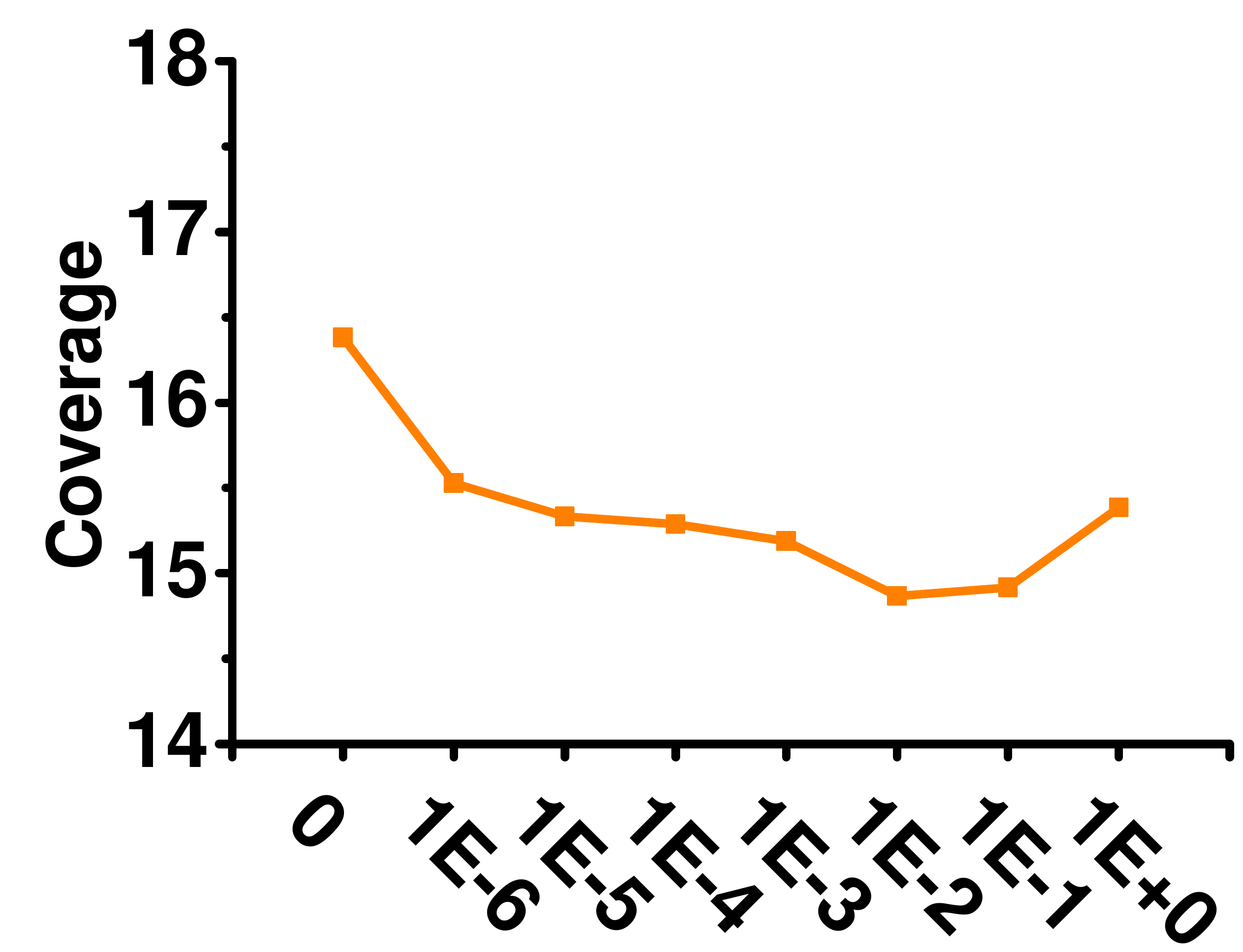}~
		\includegraphics[width=0.24\textwidth,height=0.2\textwidth]{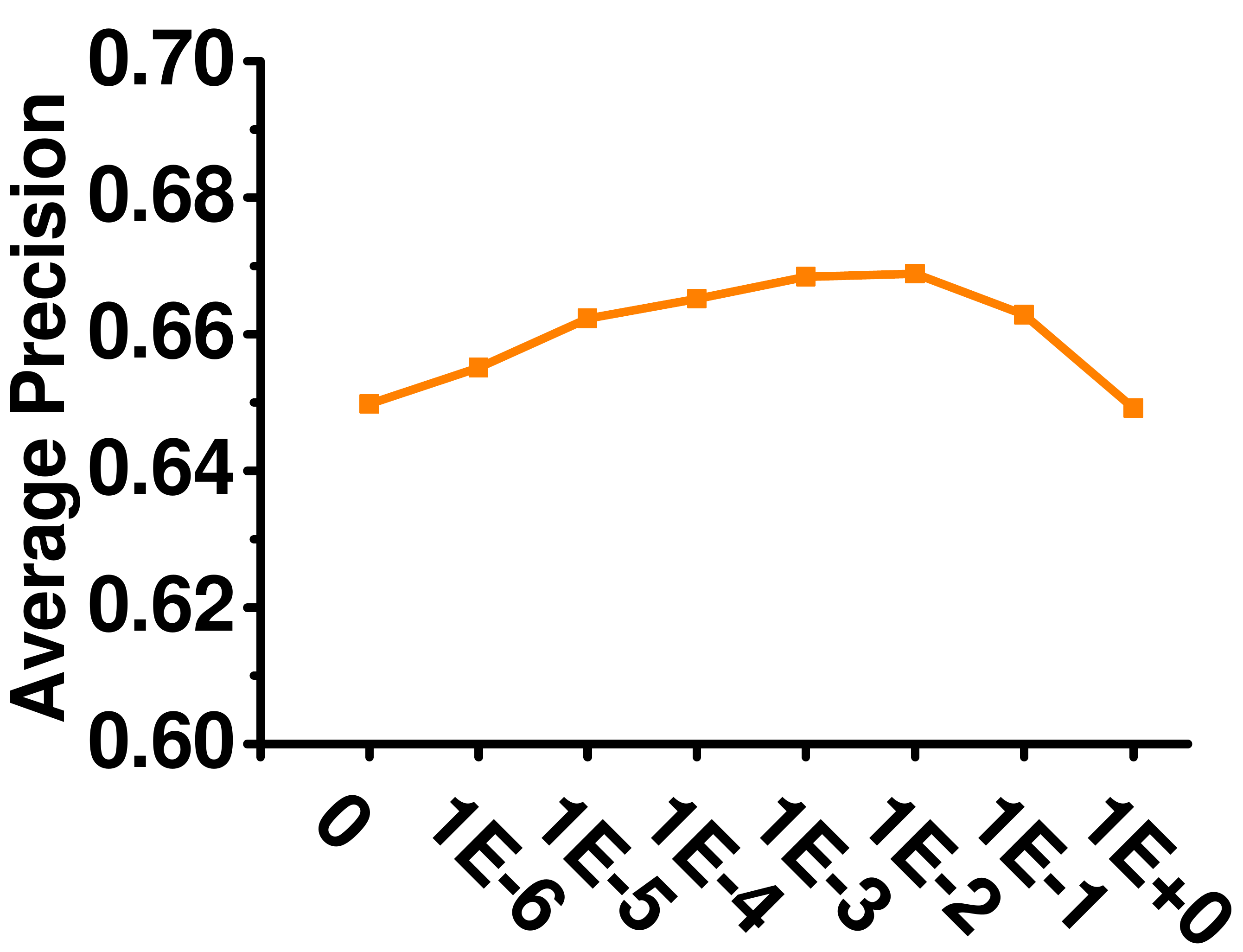}~~
	\end{minipage}
	\caption{Varying the global label manifold regularization parameter $\lambda_3$ on the
		Enron dataset. \label{fig:lambda3}}
	\centering	
	\begin{minipage}[t]{1\linewidth} 
		\includegraphics[width=0.24\textwidth,height=0.2\textwidth]{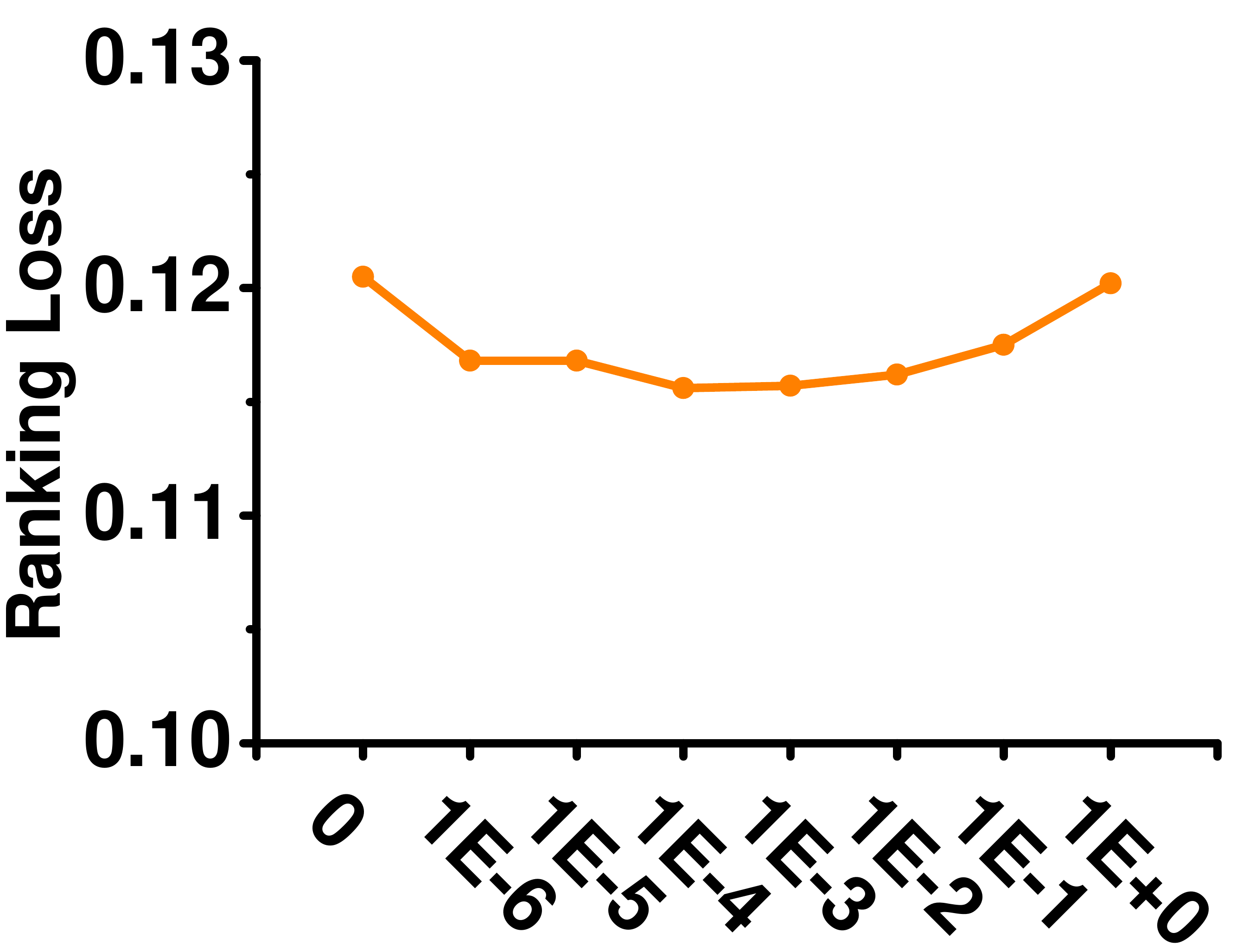}~
		\includegraphics[width=0.24\textwidth,height=0.2\textwidth]{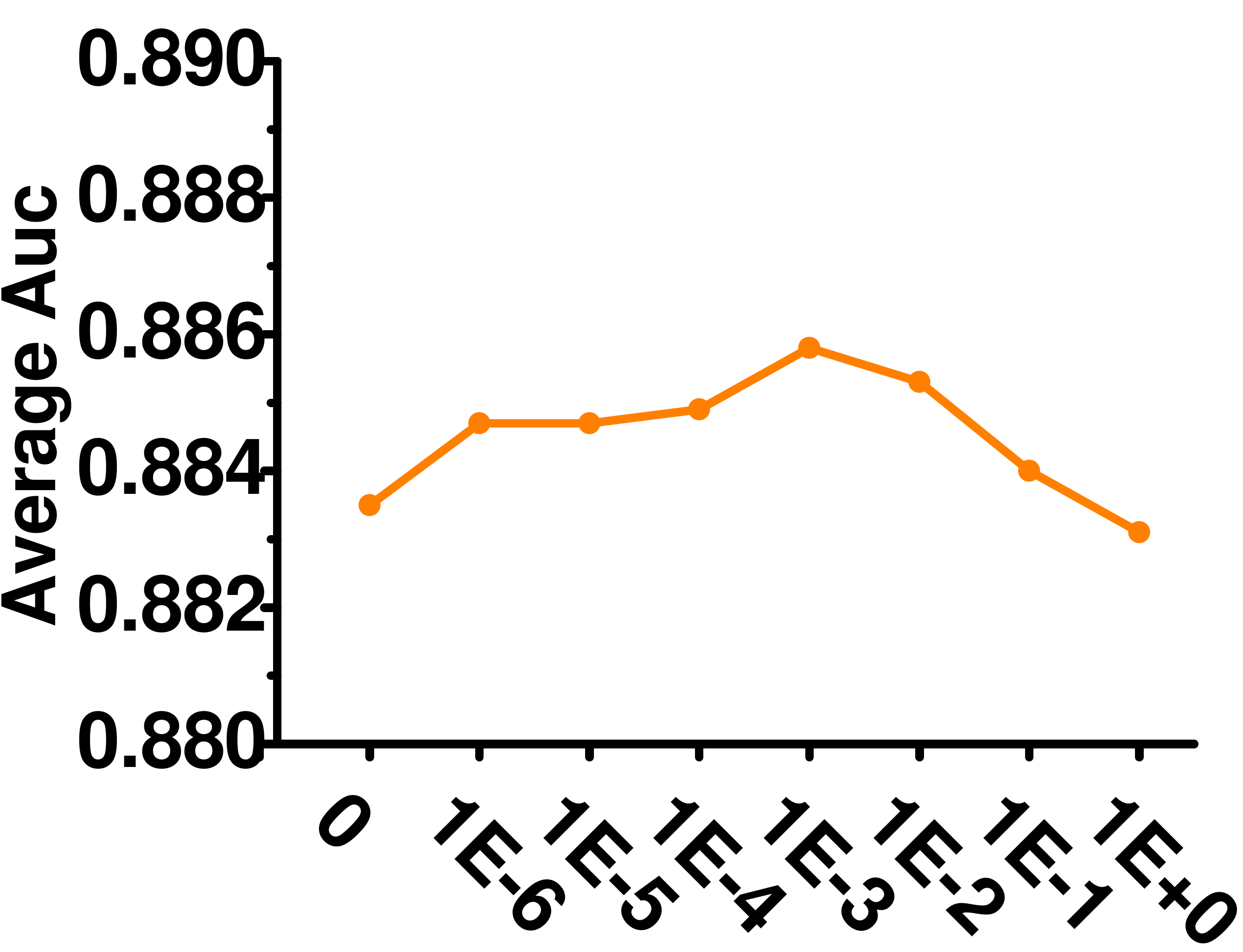}
		~	\includegraphics[width=0.24\textwidth,height=0.2\textwidth]{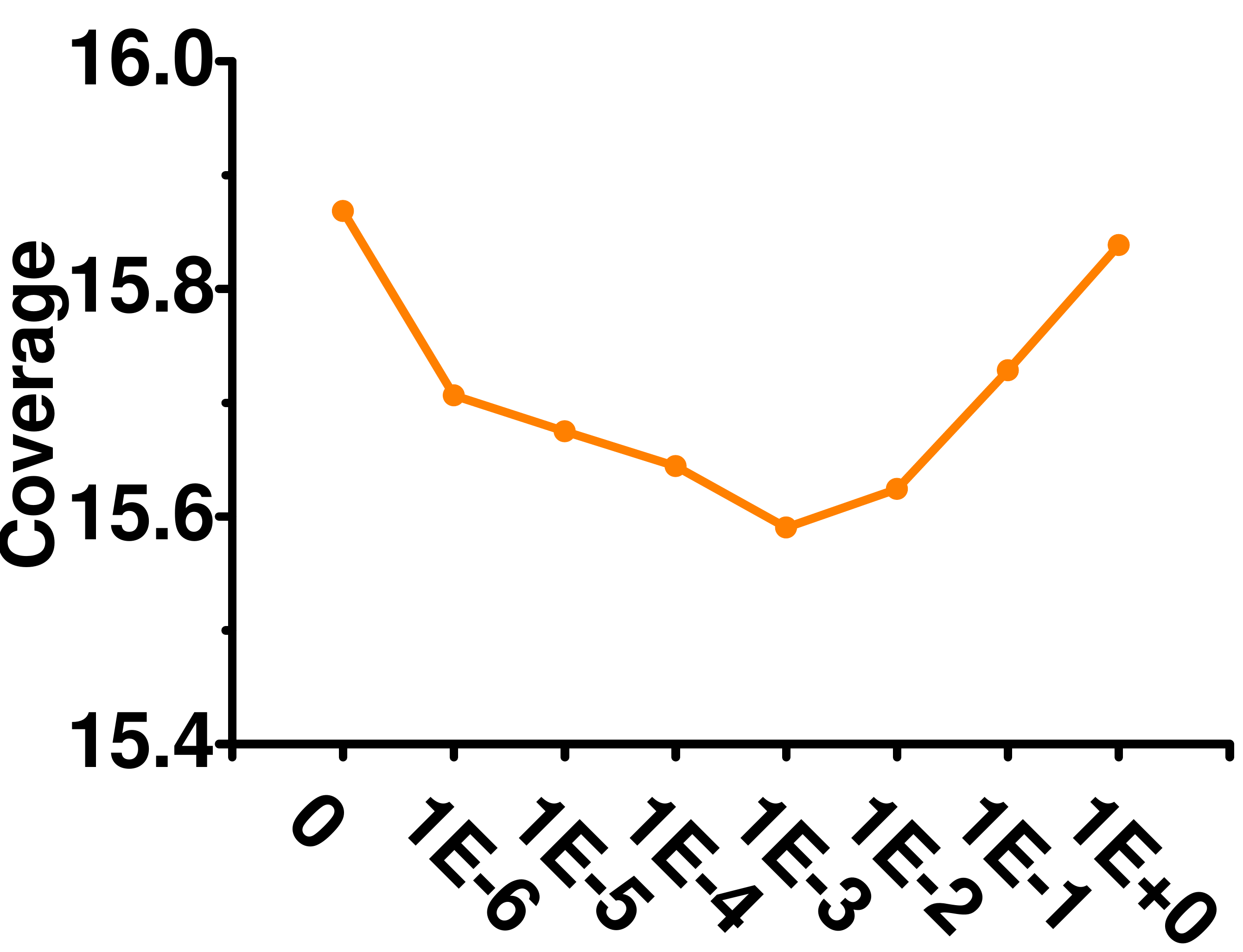}~
		\includegraphics[width=0.24\textwidth,height=0.2\textwidth]{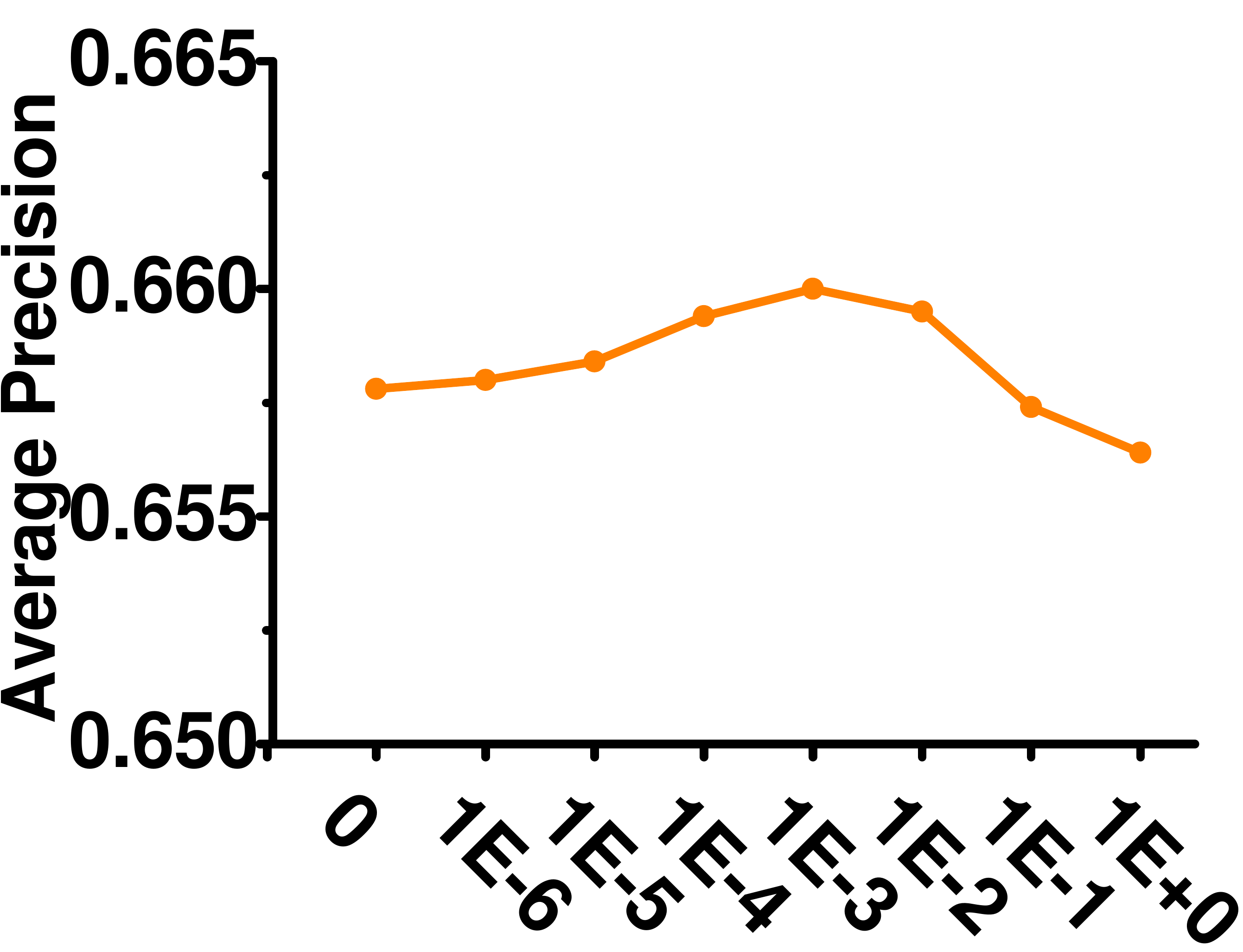}~~
	\end{minipage}
	\caption{Varying the local label manifold regularization parameter $\lambda_4$ on the Enron
		dataset. \label{fig:lambda4}}
\end{figure}

A larger  $\lambda_3$  means higher importance of global label
correlation, whereas a larger $\lambda_4$ means higher importance of local label correlation. 
%It suggests to choose similar $\lambda_3$ and $\lambda_4$.
Figures \ref{fig:lambda3} and 
\ref{fig:lambda4} 
show their effects 
%of varying $\lambda_3$  and $\lambda_4$ 
on the Enron dataset.
When $\lambda_3=0$, only local label correlations are considered, and the performance
is poor. 
With increasing $\lambda_3$,
performance improves.
%due to that more influence of global label correlations takes a part. 
However,
when $\lambda_3$ is 
very large, performance deteriorates as 
the global  label correlations
dominate.
A similar phenomenon can be observed for
$\lambda_4$. 

%Though Figures \ref{fig:lambda3} and \ref{fig:lambda4} are example plots on Enron dataset, similar results can be concluded on other datasets.

\subsubsection{Varying the Latent Representation Dimensionality $k$} 

\begin{figure}[h]
	\centering
	\begin{minipage}[t]{1\linewidth} 
		\centering
		\includegraphics[width=0.24\textwidth,height=0.2\textwidth]{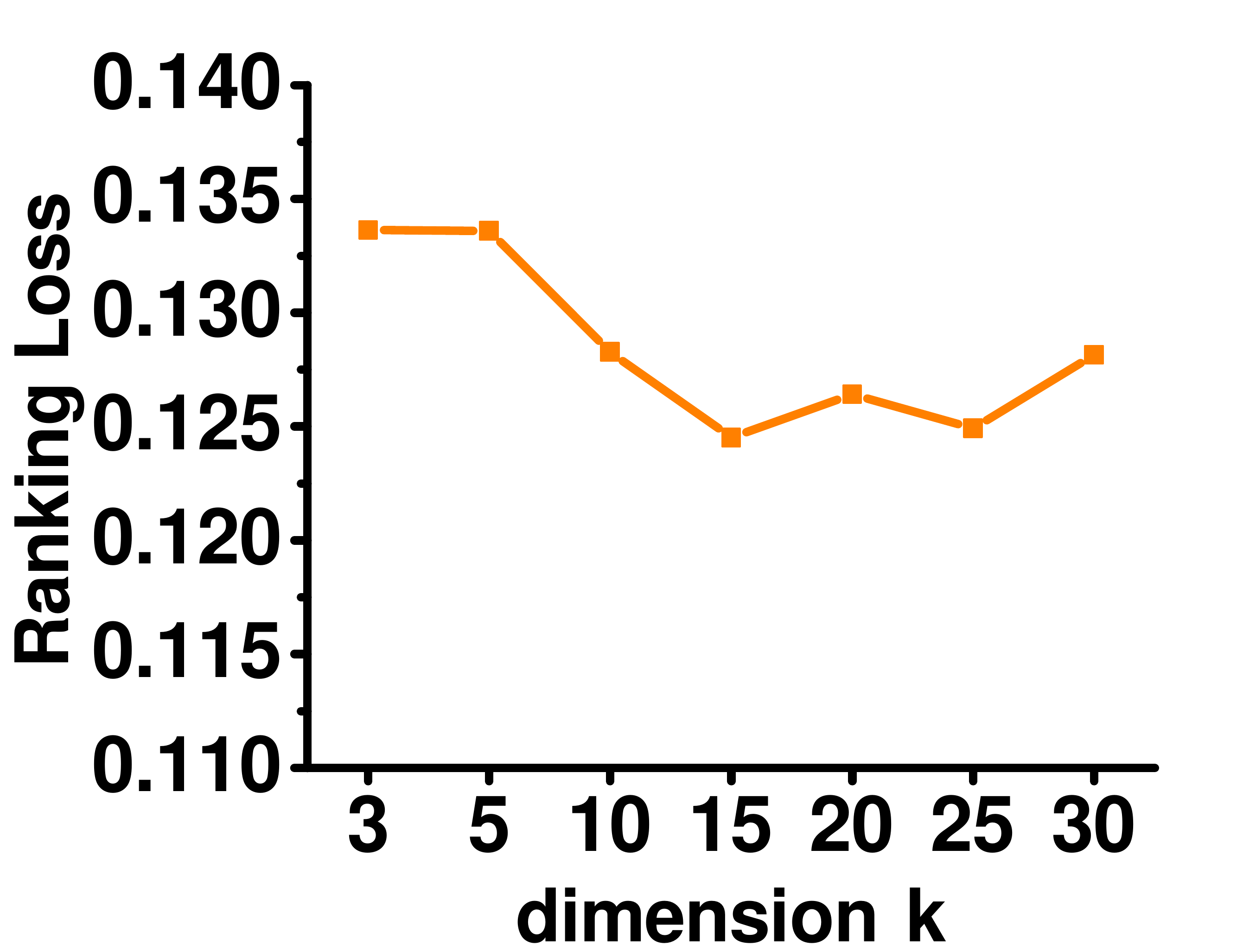}~
		\includegraphics[width=0.24\textwidth,height=0.2\textwidth]{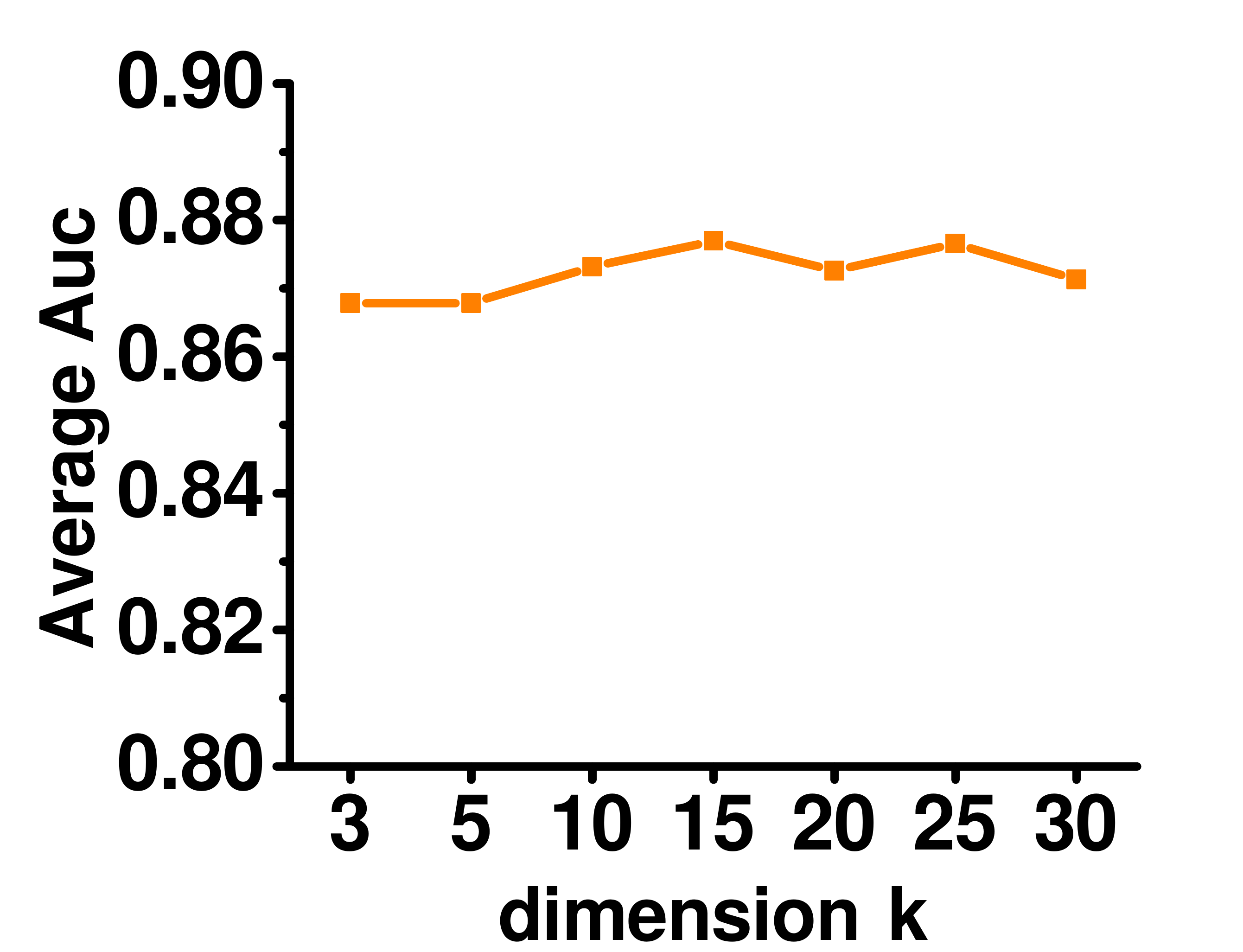}~		
		\includegraphics[width=0.24\textwidth,height=0.2\textwidth]{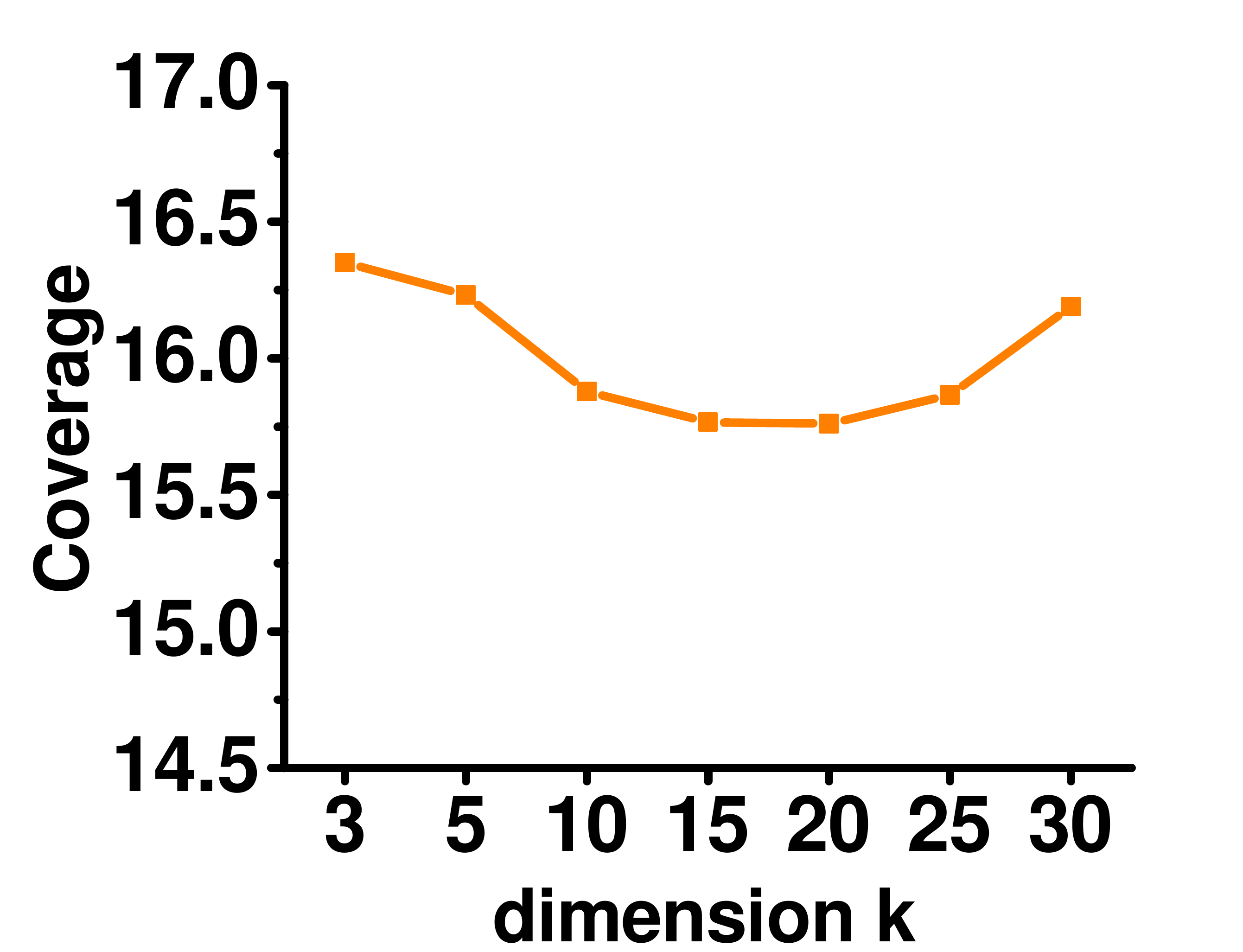}~
		\includegraphics[width=0.24\textwidth,height=0.2\textwidth]{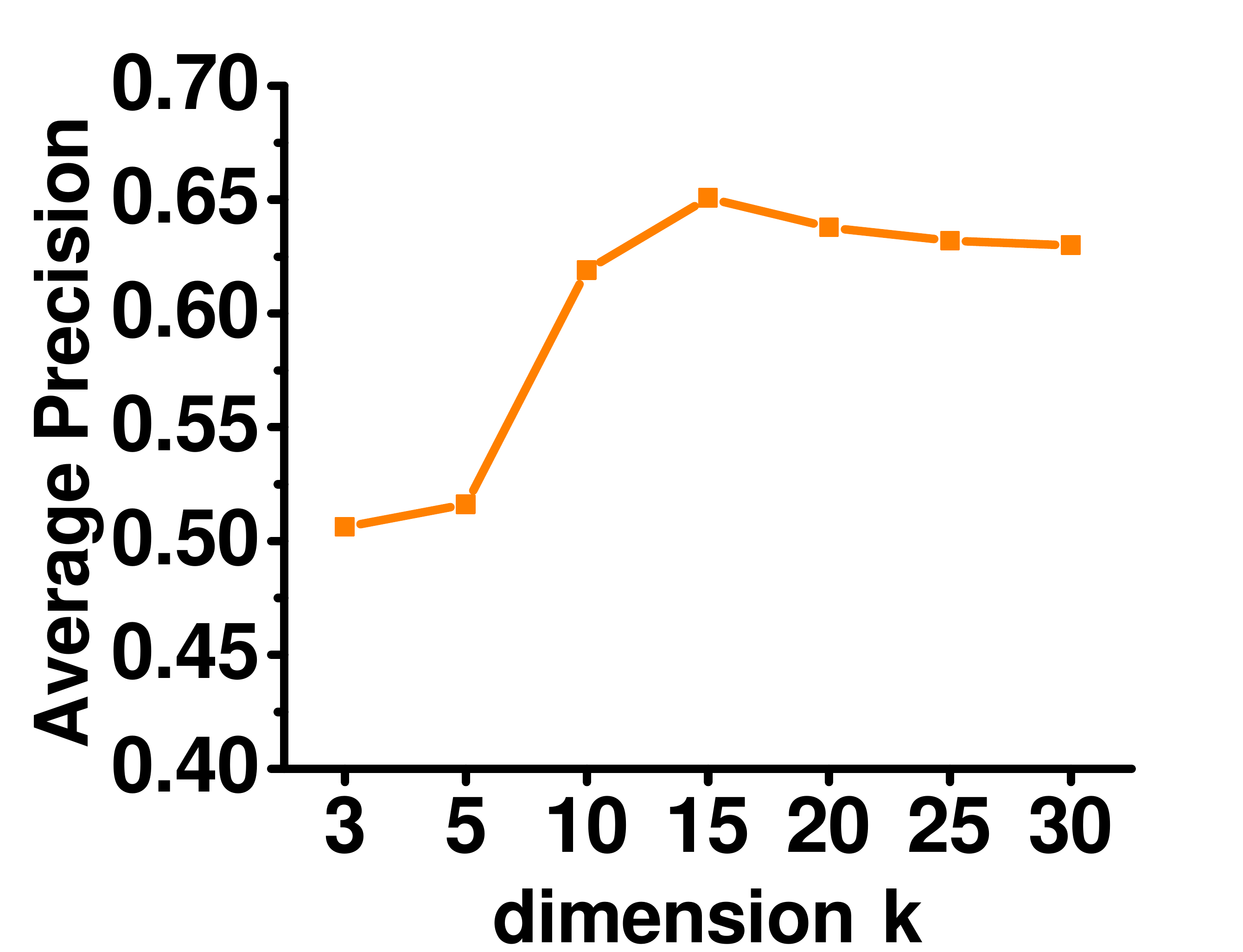}~
	\end{minipage}
	\caption{Varying the latent representation dimensionality 
		on the Enron dataset.
		\label{fig:k} 
	}
\end{figure}

Figure \ref{fig:k} shows the effect of varying $k$ on the Enron dataset. As
can be seen, 
when $k$ is too small, the latent representation cannot capture enough information.
With increasing $k$, 
performance improves.
When  $k$ is too
large, 
the low-rank structure is not fully utilized, and performance starts to get worse.

\subsubsection{Influence of $\lambda_2$} 
\begin{figure}[h]
	\centering
	\begin{minipage}[t]{1\linewidth} 
		\centering
		\includegraphics[width=0.24\textwidth,height=0.2\textwidth]{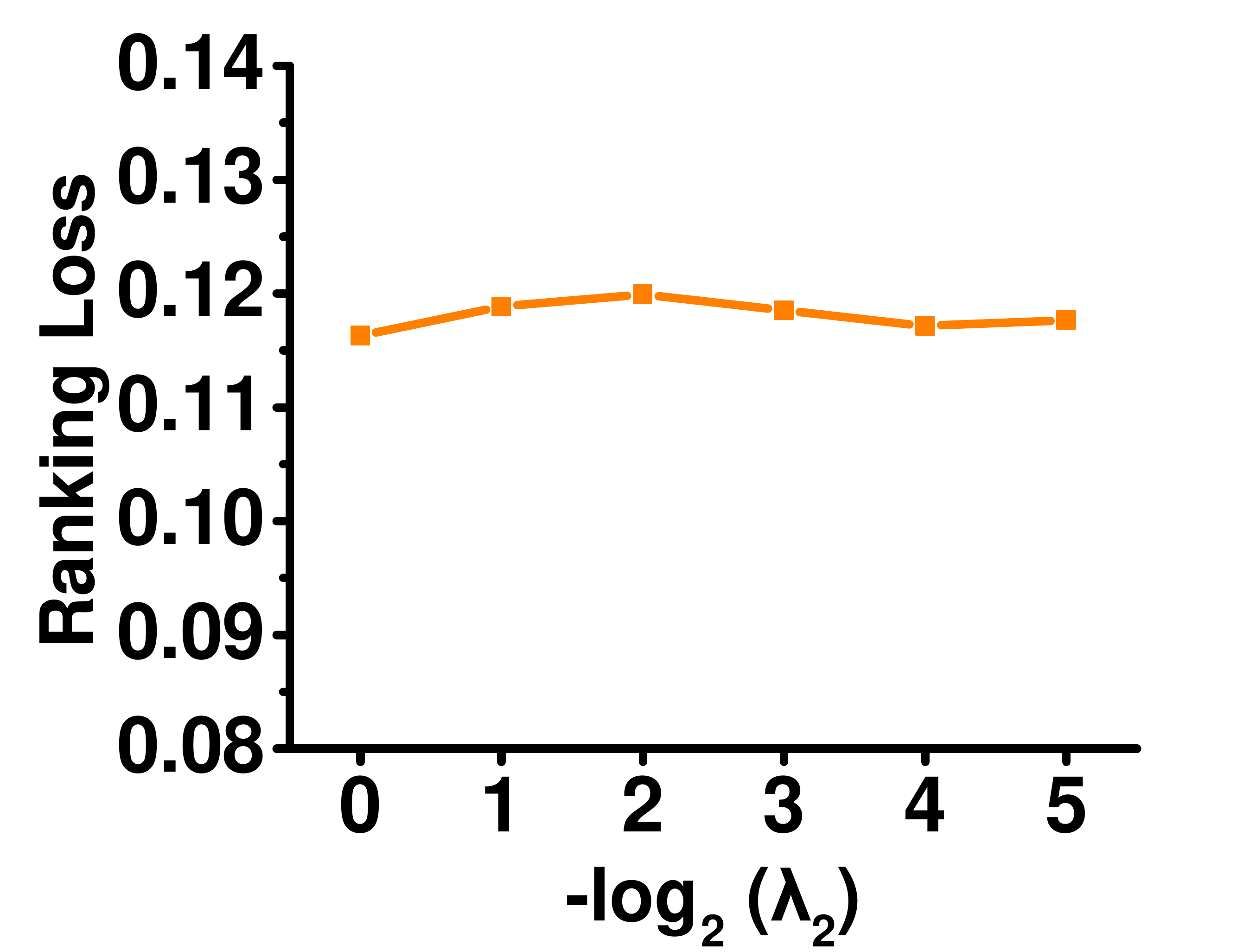}~
		\includegraphics[width=0.24\textwidth,height=0.2\textwidth]{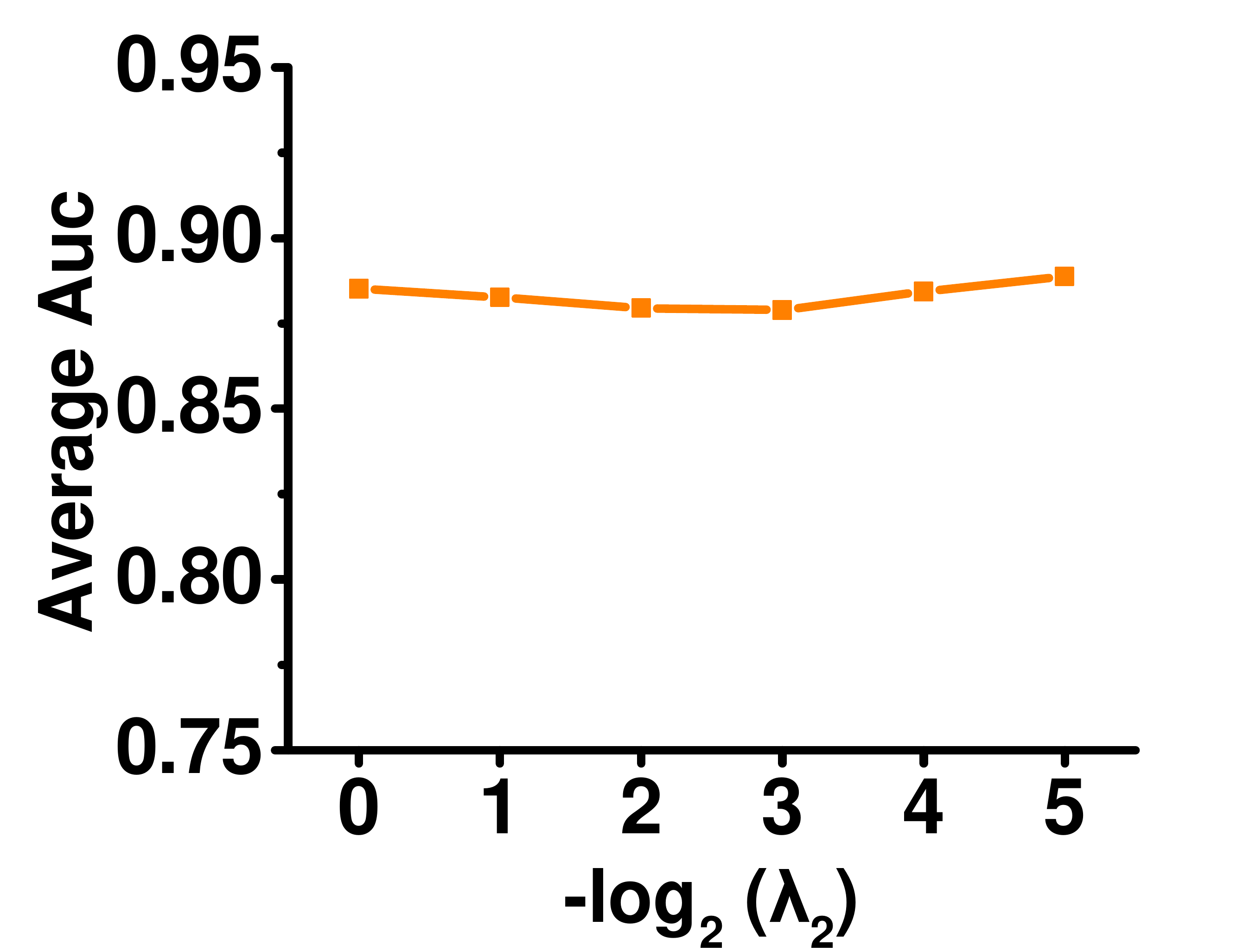}~
		\includegraphics[width=0.24\textwidth,height=0.2\textwidth]{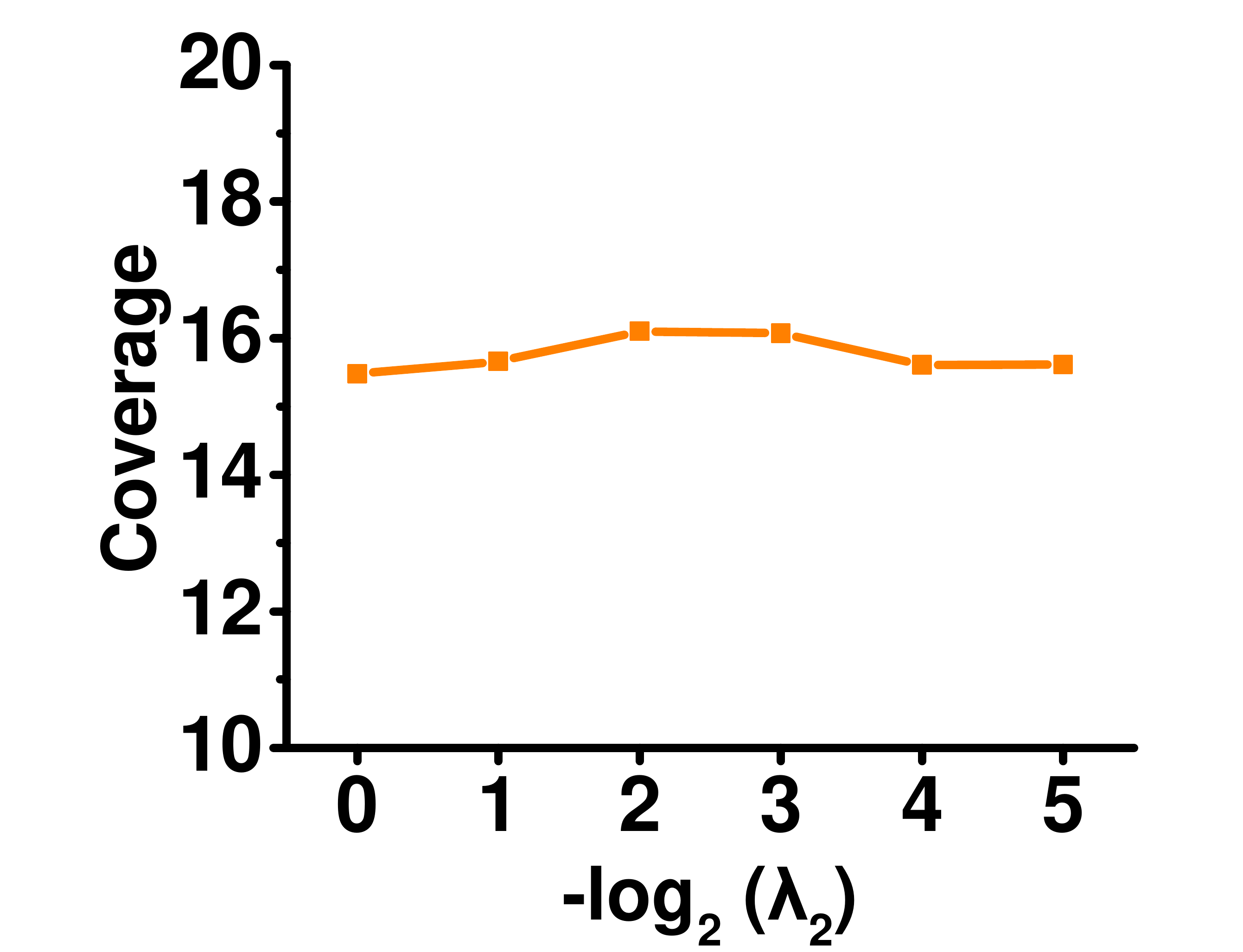}~
		\includegraphics[width=0.24\textwidth,height=0.2\textwidth]{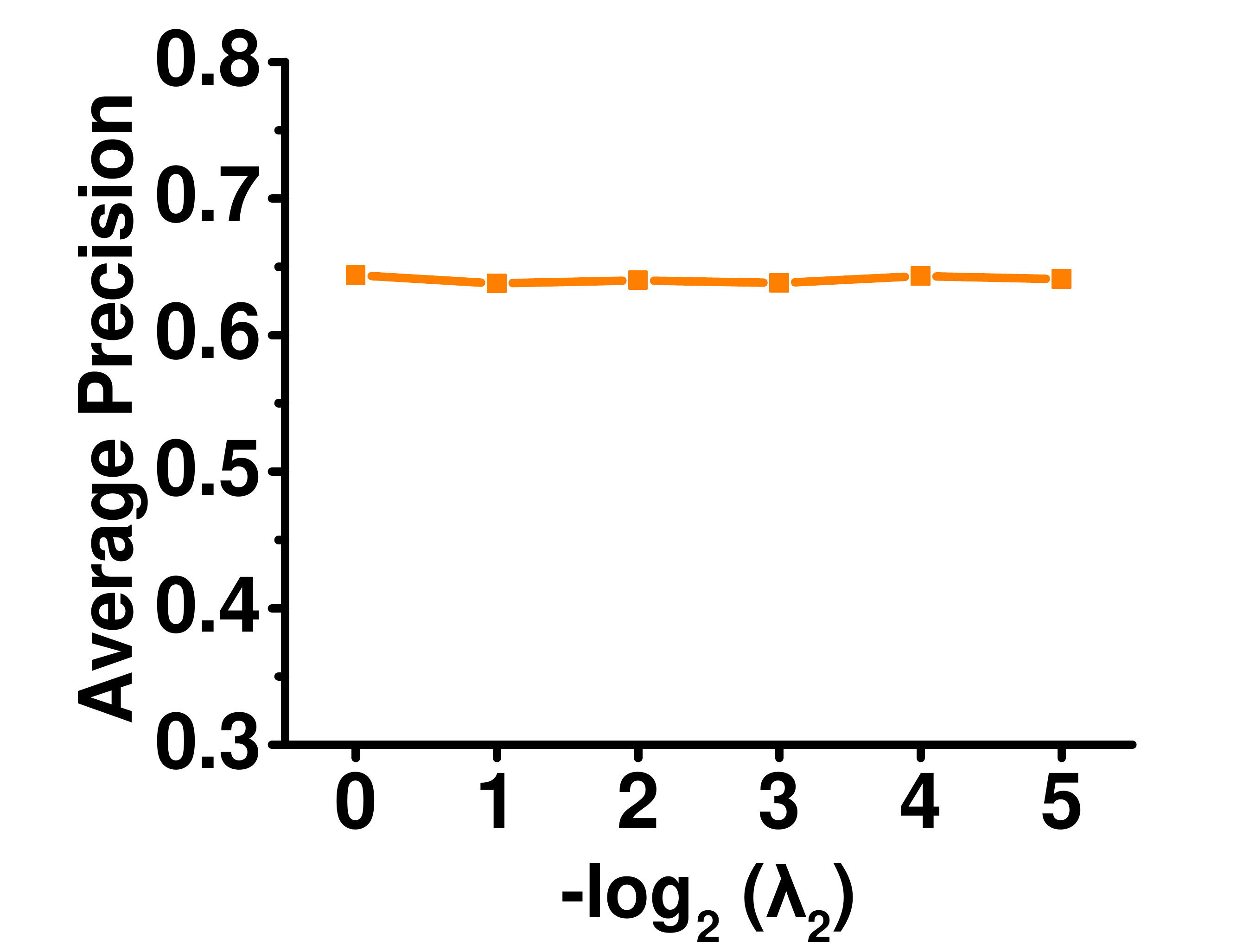}~
	\end{minipage}
	\caption{Varying $\lambda_2$ 
		on the Enron dataset.
		\label{fig:lambda2} 
	}
\end{figure}

Figure \ref{fig:lambda2} shows the effect of varying $\lambda_2$ on the Enron dataset.
As can be seen, \texttt{GLOCAL} is not sensitive to this parameter.

\section{Conclusion}\label{sec:conclusion}
%In multilabel learning, some label correlations are shared by all instances, while some are
%shared by a subset only.  It is also a usual case in the multi-label applications, only part
%of labels are observed, which makes it harder to exploit label correlations to improve the
%learning performance. Contrasting to existing approaches that either exploit label
%correlations globally or locally, working on full-label data, 

In this paper, we proposed a new multi-label correlation learning
approach \texttt{GLOCAL}, which simultaneously recovers the missing labels, trains the classifier and exploits both global and local label correlations, through learning a latent label representation and optimizing
the label manifolds. Compared with the previous work, it is the first to exploit both global
and local label correlations, which directly learns the Laplacian matrix without
requiring any other prior knowledge on label correlations. As  a result,  
%the  manifold regularizers will make 
the classifier outputs and label correlations best match each other, both globally
and locally. 
Moreover,   \texttt{GLOCAL}  provides a unified solution for both full-label and missing-label multi-label learning.  
Experimental results show that our approach outperforms the
state-of-the-art multi-label learning approaches on learning with both full labels and 
missing labels.
In our work, we handle the case that label correlations are symmetric. In many situations, correlations can be asymmetric. For example, ``mountain'' are highly correlated to ``tree'', since it is very common that a mountain has trees in it. However, ``tree'' may be less correlated to ``mountain'', because trees can be found not only in mountains, but often in the streets, parks, etc. So it is desirable to study the asymmetric label correlations in our future work.
\section*{Acknowledgment}
This research was supported by NSFC (61333014), 111 Project (B14020), and the Collaborative Innovation Center of Novel Software Technology and Industrialization.


\begin{thebibliography}{10}
	
	\bibitem{belkin2006manifold}
	M.~Belkin, P.~Niyogi, and V.~Sindhwani.
	\newblock Manifold regularization: a geometric framework for learning from
	labeled and unlabeled examples.
	\newblock {\em The Journal of Machine Learning Research}, 7:2399--2434, 2006.
	
	\bibitem{manopt}
	N.~Boumal., B.~Mishra, P.-A. Absil., and R.~Sepulchre.
	\newblock {M}anopt, a {M}atlab toolbox for optimization on manifolds.
	\newblock {\em Journal of Machine Learning Research}, 15:1455--1459, 2014.
	
	\bibitem{boutell2004learning}
	M.~Boutell, J.~Luo, X.~Shen, and C.~Brown.
	\newblock Learning multi-label scene classification.
	\newblock {\em Pattern Recognition}, 37(9):1757--1771, 2004.
	
	\bibitem{chuang2007network}
	H.-Y. Chuang, E.~Lee, Y.-T. Liu, D.~Lee, and T.~Ideker.
	\newblock Network-based classification of breast cancer metastasis.
	\newblock {\em Molecular Systems Biology}, 3(1):140--149, 2007.
	
	\bibitem{chung1997spectral}
	F.~Chung.
	\newblock {\em Spectral graph theory}, volume~92.
	\newblock American Mathematical Soc., 1997.
	
	\bibitem{REF08a}
	R.-E. Fan, K.-W. Chang, C.-J. Hsieh, X.-R. Wang, and C.-J. Lin.
	\newblock {LIBLINEAR}: A library for large linear classification.
	\newblock {\em Journal of Machine Learning Research}, 9:1871--1874, 2008.
	
	\bibitem{furnkranz2008multilabel}
	J.~F{\"u}rnkranz, E.~H{\"u}llermeier, E.~Menc{\'\i}a, and K.~Brinker.
	\newblock Multilabel classification via calibrated label ranking.
	\newblock {\em Machine Learning}, 73(2):133--153, 2008.
	
	\bibitem{goldberg2010transduction}
	A.~Goldberg, B.~Recht, J.~Xu, R.~Nowak, and X.~Zhu.
	\newblock Transduction with matrix completion: Three birds with one stone.
	\newblock In {\em Advances in Neural Information Processing Systems 23}, pages
	757--765. 2010.
	
	\bibitem{huang2012}
	S.-J. Huang and Z.-H. Zhou.
	\newblock Multi-label learning by exploiting label correlations locally.
	\newblock In {\em Proceedings of the 26th AAAI Conference on Artificial
		Intelligence}, pages 949--955, 2012.
	
	\bibitem{ji2008extracting}
	S.~Ji, L.~Tang, S.~Yu, and J.~Ye.
	\newblock Extracting shared subspace for multi-label classification.
	\newblock In {\em Proceedings of the 14th International Conference on Knowledge
		Discovery and Data Mining}, pages 381--389, 2008.
	
	\bibitem{luo2009non}
	D.~Luo, C.~Ding, H.~Huang, and T.~Li.
	\newblock Non-negative laplacian embedding.
	\newblock In {\em Proceedings of the 9th IEEE International Conference on Data
		Mining}, pages 337--346, 2009.
	
	\bibitem{melacci2011primallapsvm}
	S.~Melacci and M.~Belkin.
	\newblock {Laplacian Support Vector Machines Trained in the Primal}.
	\newblock {\em Journal of Machine Learning Research}, 12:1149--1184, 2011.
	
	\bibitem{NIPS2011_4239}
	J.~Petterson and T.~Caetano.
	\newblock Submodular multi-label learning.
	\newblock In {\em Advances in Neural Information Processing Systems 24}, pages
	1512--1520. 2011.
	
	\bibitem{Punera2005Automatically}
	K.~Punera, S.~Rajan, and J.~Ghosh.
	\newblock Automatically learning document taxonomies for hierarchical
	classification.
	\newblock {\em Proceedings of the 14th International Conference on World Wide
		Web}, pages 1010--1011, 2005.
	
	\bibitem{read2011classifier}
	J.~Read, B.~Pfahringer, G.~Holmes, and E.~Frank.
	\newblock Classifier chains for multi-label classification.
	\newblock {\em Machine Learning}, 85(3):333--359, 2011.
	
	\bibitem{subramanian2005gene}
	A.~Subramanian, P.~Tamayo, V.~Mootha, S.~M.~B. Ebert, M.~Gillette,
	A.~Paulovich, S.~Pomeroy, T.~Golub, E.~Lander, et~al.
	\newblock Gene set enrichment analysis: a knowledge-based approach for
	interpreting genome-wide expression profiles.
	\newblock {\em Proceedings of the National Academy of Sciences of the United
		States of America}, 102(43):15545--15550, 2005.
	
	\bibitem{turnbull2008semantic}
	D.~Turnbull, L.~Barrington, D.~Torres, and C.~Lanckriet.
	\newblock Semantic annotation and retrieval of music and sound effects.
	\newblock {\em IEEE Transactions on Audio, Speech and Language Processing},
	16(2):467--476, 2008.
	
	\bibitem{ueda2002parametric}
	N.~Ueda and K.~Saito.
	\newblock Parametric mixture models for multi-labeled text.
	\newblock In {\em Advances in Neural Information Processing Systems 15}, pages
	721--728. 2002.
	
	\bibitem{wang2009image}
	H.~Wang, H.~Huang, and C.~Ding.
	\newblock Image annotation using multi-label correlated green's function.
	\newblock In {\em Proceedings of the 12th International Conference on Computer
		Vision}, pages 2029--2034, 2009.
	
	\bibitem{xu2014learning}
	L.~Xu, Z.~Wang, Z.~Shen, Y.~Wang, and E.~Chen.
	\newblock Learning low-rank label correlations for multi-label classification
	with missing labels.
	\newblock In {\em Proceedings of the 14th IEEE International Conference on Data
		Mining}, pages 1067--1072, 2014.
	
	\bibitem{xu2013speedup}
	M.~Xu, R.~Jin, and Z.-H. Zhou.
	\newblock Speedup matrix completion with side information: Application to
	multi-label learning.
	\newblock In {\em Advances in Neural Information Processing Systems 26}, pages
	2301--2309. 2013.
	
	\bibitem{Yu2014}
	H.-F. Yu, P.~Jain, P.~Kar, and I.~Dhillon.
	\newblock Large-scale multi-label learning with missing labels.
	\newblock In {\em Proceedings of the 31th International Conference on Machine
		Learning}, pages 593--601, 2014.
	
	\bibitem{zhang2010multi}
	M.-L. Zhang and K.~Zhang.
	\newblock Multi-label learning by exploiting label dependency.
	\newblock In {\em Proceedings of the 16th International Conference on Knowledge
		Discovery and Data Mining}, pages 999--1008, 2010.
	
	\bibitem{zhang2014review}
	M.-L. Zhang and Z.-H. Zhou.
	\newblock A review on multi-label learning algorithms.
	\newblock {\em IEEE Transactions on Knowledge and Data Engineering},
	26(8):1819--1837, 2014.
	
\end{thebibliography}
\end{document}